%% file: main.tex
\documentclass[final,5p,times,twocolumn,authoryear]{elsarticle}

\usepackage{amssymb}
\usepackage{amsmath}
\usepackage{graphicx}
\usepackage{subcaption}
\usepackage{float}
\usepackage{booktabs}
\usepackage{siunitx}
\usepackage[hidelinks]{hyperref}
\usepackage{xcolor}
\usepackage{transparent-nometadata}
\makeatletter
\@namedef{ver@transparent.sty}{9999/99/99}
\makeatother
\usepackage{tikz}
\usetikzlibrary{arrows.meta, positioning}
\usepackage{listings}
\usepackage{algorithm}
\usepackage{algorithmic}
\usepackage{tabularx}
\graphicspath{{./}}

\journal{Ocean Engineering}

\begin{document}

\begin{frontmatter}

\title{Foundation models on the bridge: Semantic hazard detection and safety maneuvers for maritime autonomy with vision-language models}

\author[1]{Kim Alexander Christensen\corref{cor1}}
\ead{kimac@stud.ntnu.no}
\author[2]{Andreas Gudahl Tufte}
\author[3]{Alexey Gusev}
\author[4]{Rohan Sinha}
\author[5]{Milan Ganai}
\author[3]{Ole Andreas Alsos}
\author[4,6]{Marco Pavone}
\author[1]{Martin Steinert}

\affiliation[1]{organization={Dept. of Mechanical and Industrial Engineering, NTNU}, Department and Organization 
            country={Norway}}
\affiliation[2]{organization={Dept. of Engineering Cybernetics, NTNU}, country={Norway}}
\affiliation[3]{organization={Dept. of Design, NTNU}, country={Norway}}
\affiliation[4]{organization={Dept. of Aeronautics and Astronautics, Stanford University}, country={USA}}
\affiliation[5]{organization={Dept. of Computer Science, Stanford University}, country={USA}}
\affiliation[6]{organization={NVIDIA Research}, country={USA}}

\cortext[cor1]{Corresponding author}

\begin{abstract}
The draft IMO MASS Code requires autonomous and remotely supervised maritime vessels to detect departures from their operational design domain, enter a predefined fallback that notifies the operator, permit immediate human override, and avoid changing the voyage plan without approval. Meeting these obligations in the alert-to-takeover gap calls for a short-horizon, human-overridable fallback maneuver. Classical maritime autonomy stacks struggle when the correct action depends on meaning (e.g., diver-down flag means people in the water, fire close by means hazard). We argue (i) that vision–language models (VLMs) provide semantic awareness for such out-of-distribution situations, and (ii) that a fast–slow anomaly pipeline with a short-horizon, human-overridable fallback maneuver makes this practical in the handover window. We introduce \emph{Semantic Lookout}, a camera-only, candidate-constrained vision–language model (VLM) fallback maneuver selector that selects one cautious action (or station-keeping) from water-valid, world-anchored trajectories under continuous human authority. On 40 harbor scenes we measure per-call scene understanding and latency, alignment with human consensus (model majority-of-three voting), short-horizon risk-relief on fire hazard scenes, and an on-water alert$\rightarrow$fallback maneuver$\rightarrow$operator handover. Sub-10\,s models retain most of the awareness of slower state-of-the-art models. The fallback maneuver selector outperforms geometry-only baselines and increases standoff distance on fire scenes. A field run verifies end-to-end operation. These results support VLMs as semantic fallback maneuver selectors compatible with the draft IMO MASS Code, within practical latency budgets, and motivate future work on domain-adapted, hybrid autonomy that pairs foundation-model semantics with multi-sensor bird’s-eye-view perception and short-horizon replanning. Website: \url{https://kimachristensen.github.io/bridge_policy}
\end{abstract}


\end{frontmatter}


\input{01_introduction}
\input{02_related_work}
\input{03_problem_formulation}

\input{04_method}
\input{05_experiments}
\input{07_formative_handover}

\input{08_discussion}
\input{09_conclusion}

\section*{Acknowledgments}
We would like to acknowledge the financial support provided by Fosenregionen and the Norwegian Research Council through project number 321780. We would also like to thank Karl Philip Abrahamsen and Erik Alsos for their roles in design mockup/visualization and GUI programming respectively. 

\bibliographystyle{elsarticle-harv}
\bibliography{references}

\appendix

\section{Fast anomaly detection}
\label{app:e1}
\input{appendix_fast_anomaly}

\section{Platform, control, and ROC details}
\label{app:system}
\input{appendix_platform}

\section{Formative handover study protocol}
\label{app:e3}
\input{appendix_formative_handover}

\section{Formative handover study report}
\label{sec:e3_results}
\input{appendix_formative_results}

\section{Miscellaneous}
\label{app:misc}
\input{appendix_misc}

\end{document}

%% file: 01_introduction.tex
\section{Introduction}
\label{sec:intro}

The maritime industry is moving towards autonomous and remote operations using smaller autonomous surface vehicles (ASVs) and larger maritime autonomous surface ships (MASS), with potential benefits including cost savings, more efficient operations, and increased crew safety \citep{Brekke2022, wang2019roboat, blanke2024greenhopper, akbar2021economic}. Operational systems already exist in both harbor environments and at sea. However, despite the high performance of autonomy modules under nominal conditions, they might still fail in a large number of possible cases. Some of these are purely technical and foreseeable, like loss of a sensor or GNSS jamming/spoofing \citep{volden2022vision, bhatti2017hostile, liu2018impact}, while others arise from situations that traditionally rely on human judgment because they are ambiguous, diverse, or rare in data~\citep{SinhaSharmaEtAl2022}. These problems lead to reliance on land-based remote operation centers (ROCs), where a human operator continuously monitors the vessel and intervenes when needed~\citep{Alsos2022}. Current ROCs are largely set up so that one operator is responsible for one vessel, but over time, one operator (or a small team) is expected to supervise many vessels simultaneously~\citep{veitch2022systematic}. Given a large enough fleet, this will require robust monitors that can alert the ROC if a vessel function is exiting the Operational Design Domain (ODD)~\citep{IMO2024_MSC109_5} so that an operator can intervene, creating a handover window between the alert and human intervention during which the vessel must remain safe and legible. In this paper we argue (i) that vision–language models (VLMs) provide semantic awareness that is specifically valuable in such out‑of‑distribution (OOD) situations, and (ii) that a fast–slow anomaly pipeline with a short‑horizon, human‑overridable fallback maneuver makes this practically feasible in the handover window.

The regulatory IMO MASS Code currently being drafted speaks directly to this. Systems should \textit{``be able to detect whether [their] current state of operation meets the ODD''}, and if the ship deviates from its operational envelope it should \textit{``enter a predefined fallback state''} and \textit{``notify [its] crew and the operator''}. Navigation automation must be \textit{``capable of being overridden at all times''} and \textit{``allow for control to be taken immediately''}. The \textit{``use of the voyage plan, and any modification of the voyage plan''} by the navigation system is \textit{``not … possible without … approval … by the [operator]''}~\citep{IMO2024_MSC109_5}. We take this as a design requirement for an \emph{alert} $\rightarrow$ \emph{fallback maneuver} $\rightarrow$ \emph{human override} loop in which the fallback maneuver consists only of short‑horizon actions chosen from pre-approved primitives (or Station‑keeping), is immediately overridable, and never edits the overall voyage plan. In this paper, the IMO \emph{fallback state} is the vessel’s predefined degraded mode after an alert, and the \emph{fallback maneuver} is the single short-horizon, pre-approved motion action executed within that fallback state during the alert-to-override interval until the operator overrides (takes manual control) or the alert clears.

Classic maritime stacks typically cover obstacle detection (e.g., radar/LiDAR), tracking (e.g., AIS, multi‑target trackers), and short‑horizon collision avoidance~\citep{johansen2016ship,hem2024autonomous}. They don't, however, interpret \emph{semantic} cues such as ``diver‑down'' flags, “keep‑out” lines, or a vessel on fire; cases where the correct action depends on meaning, not geometry alone. To complicate things further, many hazards appear as out-of-distribution (OOD) scenes for vessel/object detectors: rare, open-ended ``unknown unknowns'' beyond prior experience~\citep{SinhaSharmaEtAl2022}. Foundation models like large language models (LLMs) and vision-language models (VLMs) trained on internet‑scale data provide strong semantic priors that support zero‑/few‑shot generalization and improved OOD behavior \citep{brown2020language, radford2021learning, wortsman2022robust}. Following \cite{sinha2024real}, we define anomalies as \emph{semantic deviations from prior operational experience} and leverage a two‑stage pattern in which a \emph{fast} embedding‑space monitor can trigger \emph{slower} generative reasoning. Their work shows this can enable real‑time OOD anomaly detection and reactive planning; what remains is to adapt it to the specific needs of maritime autonomy and the IMO MASS code requirement for immediate, human‑overridable control. We argue that being able to detect and react to such anomalies is an essential step in moving beyond constant, laborious human supervision and enabling one-to-many supervised maritime autonomy. 

We operationalize the IMO MASS constraints as an \emph{alert} $\rightarrow$ \emph{fallback maneuver} $\rightarrow$ \emph{human override} loop. Stage~(1) is a small, camera‑first \emph{fast anomaly alert} adapted from \citet{sinha2024real}. In~\ref{app:e1}, we provide details and small $n$ evidence that this monitor also functions in the maritime domain. Stage~(2) is a short‑horizon fallback maneuver that keeps the vessel safe and interpretable until the operator takes charge. Stage~(3) is a hard‑priority joystick override at the ROC. This paper focuses mainly on Stage~(2) while validating the full chain.

Specifically, we introduce \emph{Semantic Lookout}, a proof‑of‑concept, camera‑only \emph{fallback maneuver selector} that constrains a vision–language model (VLM) to make one cautious, short‑term choice among pre‑vetted, water‑safe trajectory candidates overlaid on the camera view (or to stop when uncertain). The system is designed to handle previously unseen semantic hazards and is aligned with the draft IMO MASS Code.

\begin{figure*}[!t]
  \centering
  \includegraphics[width=0.92\textwidth]{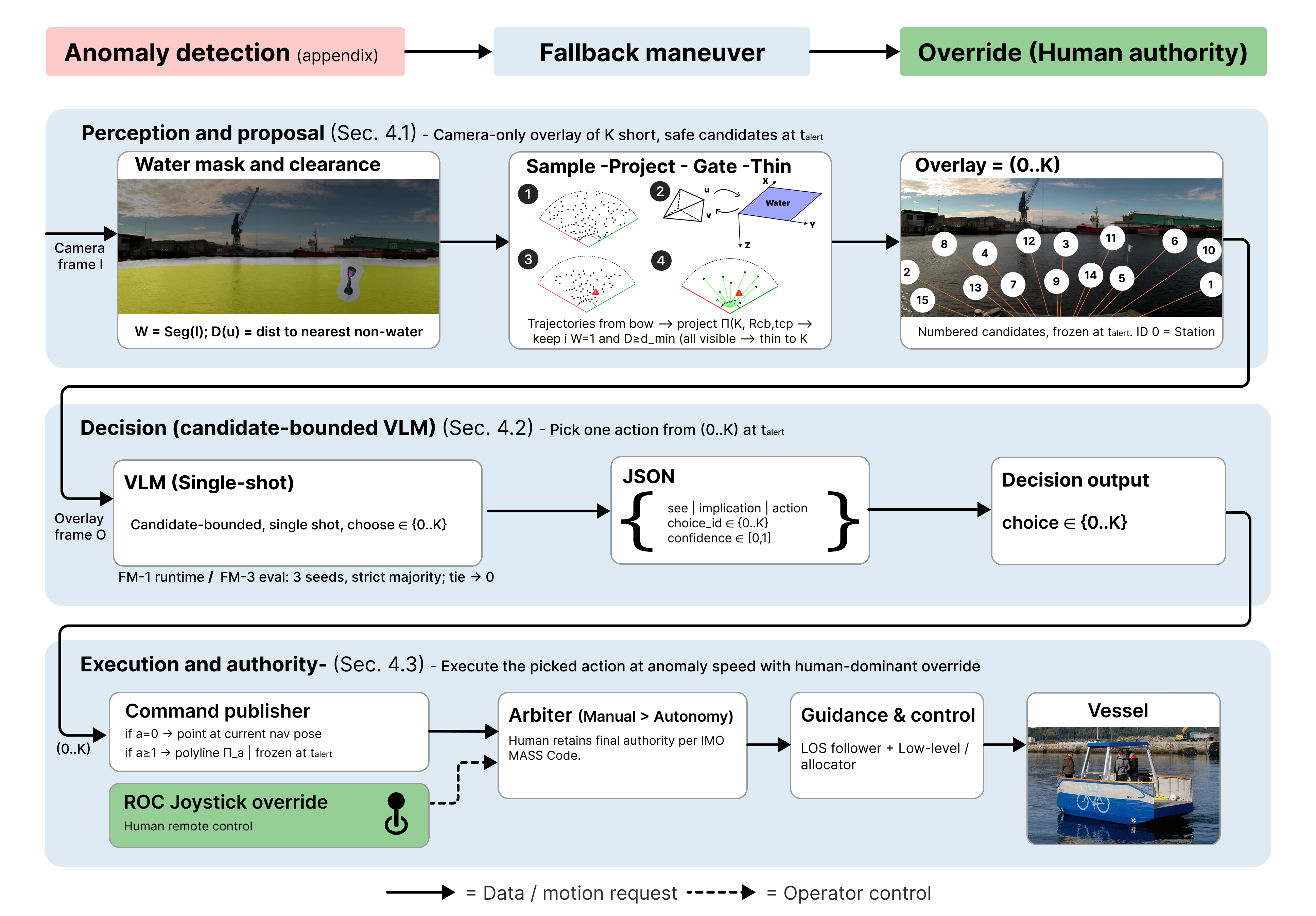}
  \caption{System overview showing the three main modules: \emph{perception and proposal}, \emph{decision} and \emph{execution and authority}.}
  \label{fig1}
\end{figure*}

The objective of this study is to evaluate whether a candidate-constrained VLM can serve as an IMO MASS Code--aligned fallback maneuver selector in the alert-to-override interval. More broadly, we aim to evaluate foundation models as a promising technology to build reliability in OOD edge-cases that would otherwise require human judgment. We evaluate four questions on the same overlay stack used in live field tests: (i) scene understanding and latency in semantic maritime anomalies; (ii) alignment of selected fallback maneuvers with aggregated human Accept/Best judgments relative to geometry-only baselines; (iii) short-horizon ``risk-relief'' on unambiguously dangerous fire hazards (standoff distance); and (iv) end-to-end operation of the alert$\rightarrow$fallback maneuver$\rightarrow$human override chain in a live harbor run with immediate joystick override and ROC-legible presentation, complemented by a formative handover human-machine interface (HMI) study. Detailed hypotheses, experiment setups and results are shown in Sec.~\ref{sec:experiments}. In summary, our contributions are as follows:

\begin{enumerate}
  \item \textbf{IMO MASS‑aligned fallback maneuver architecture.} We formalize the alert$\rightarrow$fallback maneuver$\rightarrow$override loop as the main design constraint: short‑horizon, pre‑approved actions (or Station‑keeping), immediate override, and no overall voyage‑plan edits. This leads to the following main modules (Fig.~\ref{fig1}).
  \item \textbf{Camera‑only candidate set and gating.} From a single image frame, we compute a water mask and pixel‑clearance, sample/project short motion primitives, gate them in pixel‑space, then reduce to a world‑anchored, numbered candidate set used consistently online and offline.
  \item \textbf{VLM fallback maneuver selector with low-level execution and human override.} A strict‑schema VLM decision over the above candidates and an execution/authority path that publishes world‑fixed waypoints to a line-of-sight waypoint follower with a direct joystick override blend ensuring immediate human authority.
  \item \textbf{Evidence of feasibility.} Small‑$n$ fast anomaly monitor experiment (\ref{app:e1}); offline evaluation on overlays with human labeled ground truth; short‑horizon fire “risk‑relief” vs. geometry‑only baselines; and a live alert$\rightarrow$fallback maneuver$\rightarrow$operator handover experiment with immediate joystick override.
\end{enumerate}

%% file: 02_related_work.tex
\section{Related work}
\label{sec:related}

\subsection{Remote operation centers, human-machine interface and shared control for MASS/ASV}
\label{related:roc}

\paragraph{Regulatory frame}
The IMO MASS Code---which is currently being drafted and might still change---positions autonomy as bounded and supervised. We adopt the Code’s terminology: \emph{ODD} (Operational Design Domain; conditions under which a specific autonomous/remote function is designed to operate) and \emph{OE} (Operational Envelope; ship‑level capabilities and limitations across modules). A deviation outside a function’s ODD is treated as a \emph{degraded state}; exceeding the ship’s OE should trigger a predefined \emph{fallback state}. In all cases, the navigation system must be \emph{capable of override at all times} so human control can be taken immediately. Any use of, or changes to, the voyage plan require Master (person in charge) approval \citep{IMO2024_MSC109_5} These provisions support our focus on the alert$\rightarrow$override interval.

\paragraph{ROC supervision and HMI}

The NTNU Shore Control Lab provides a flexible, instrumented ROC for studying remote supervision and handover under realistic constraints~\citep{Alsos2022}. Building on this infrastructure, a human-centered ``situation awareness by design''
ROC prototype emphasizes a camera-first view, explicit mode/authority cues, and predictable human control~\citep{gusev2025situation}. Evidence for one-to-many supervision remains limited: a controlled study reports degraded performance when operators supervise three vessels vs. one, and shows that shorter available response times (e.g., 20\,s vs.\ 60\,s) materially affect takeover outcomes, with lacking decision support systems compounding the effect~\citep{Veitch2024}.

Control takeover is not instantaneous: first experimental results on highly automated inland vessels suggest that takeover times of more than 20\,s can be expected even in simplified scenarios~\citep{shyshova2024takeover}. Survey-based results for conventional merchant ships further suggest that response and situation-awareness recovery can be on the order of minutes (including physical movement to the control position), with reported time-budgets for hazardous situations also on the order of several minutes and with high variability~\citep{wrobel2025time}. Together, these results motivate systems that both \emph{buy time} via safety-preserving degraded-mode actions and support rapid situation awareness during the alert-to-override window.

To our knowledge, there is no published demonstration of robust one-operator to multi-vessel supervision with acceptable performance in realistic conditions. In this work, we adapt the ROC prototype from~\cite{gusev2025situation} for our experiments and use Endsley's SA lens (perception, comprehension, projection) as a framing when designing the graphical user interface used in the formative HMI experiment in Sec.~\ref{sec:formative_handover}~\citep{Endsley2023}. We do not explicitly test one-to-many supervision, but see this system as one of the steps toward it.

\paragraph{FM/LLM UIs in maritime}
Prior work explored \emph{LLM}‑based conversational mission planning with the operator explicitly in the Master role \citep{Christensen2025}. Separately, a simulator study of a VHF conversational interface found lower trust than a human officer and argued for tighter coupling to autonomy and retained ROC oversight \citep{hodne2024conversational}. Our focus differs from these: we target the ODD/OE alert window with \emph{VLM}‑driven, world‑anchored actions.

\paragraph{Our approach}
We contribute an IMO MASS Code‑aligned anomaly‑handling method for the interval between alert and override that proposes a single cautious fallback maneuver (or abstains) with brief rationale, preserves \emph{immediate} override, and is intended to help buy time and reduce operator time pressure in the alert$\rightarrow$override window. This is aimed at making supervision of multiple vessels by one operator feasible, even though we evaluate a single vessel here. All closed‑loop experiments run on the Shore Control Lab workstation from~\citet{gusev2025situation}. This is to our knowledge the first use of foundation models in the real maritime domain used for anomaly handling specifically and for vision-to-action in general (one single exception comes close to the latter, which runs a rudimentary and simulated VLM vision to action ASV stack (and no anomaly detection or handling)~\citep{kim2025visionllm}).

\subsection{Classical maritime autonomy stack}
\label{related:classical_stack}

\paragraph{Perception}
Some common camera-based maritime perception methods include (i) \emph{semantic water/obstacle segmentation} to obtain a navigable-water mask, and (ii) \emph{object detection} to localize and classify vessels and other targets. Representative segmentation approaches include \emph{Water Segmentation and Refinement} (WaSR) and its embedded variant (eWaSR), along with inland/canal variants that can handle reflections, wakes, and narrow channels. Joint detect-and-segment models have also been demonstrated on edge hardware \citep{Bovcon2022wasr,Terek2023,zhou2022collision,yang2024joint}. On the detection side, YOLO-family models trained on domain-specific datasets (e.g., SMD/SMD-Plus, SeaShips) offer real-time performance but remain constrained by narrow taxonomies and out-of-distribution brittleness \citep{kim2022object}, with many rare or ad hoc hazards (e.g., diver-down flags, fire) lying outside the training distribution.

\paragraph{Collision avoidance}
Representative COLREGs-compliant collision avoidance methods formulate the problem as a scenario-based MPC (SB-MPC): at each step, a finite set of course/speed behaviors is simulated and scored by a cost that blends collision risk, COLREGs penalties, and maneuver effort \citep{johansen2016ship,hagen2022scenario}. In deployed stacks, SB-MPC is typically fed by multi-target tracks from AIS/radar fusion before optimization \citep{hem2024autonomous}. Surveys place SB-MPC alongside velocity-obstacle methods, potential-field/gradient methods, and graph/sampling-based planners; these approaches are fundamentally geometry- and rule-centric, encoding COLREGs \citep{COLREGs} over kinematics rather than interpreting semantic cues and scene meaning \citep{zhang2021collision,ozturk2022review}.

\paragraph{Our approach}
Our system is a camera-only proof-of-concept: eWaSR water masks gate a world-anchored candidate set, and our VLM fallback maneuver selector selects among admissible candidates based on scene meaning instead of only geometry or predefined categories. For baseline comparison, we evaluate simple geometry-only, semantics-agnostic heuristics on the same gated candidate set (Keep-station / Keep-course / Keep-starboard / Forward / Clearance). This is a camera-only simplified proxy to SB-MPC and similar systems for our proof-of-concept scope. Selected behaviors are executed by a standard dynamic positioning (DP) system with line-of-sight (LOS) guidance and azimuth-thruster control allocation on our vessel (see~\ref{app:system} and \cite{Tufte2025IntegratedMotion,Brekke2022} for details).

\subsection{Foundation models in robotics}
Foundation models (FMs)~\citep{bommasani2021opportunities}, such as LLMs and VLMs, have enabled semantic reasoning across various autonomous systems, including manipulation \citep{kim24openvla, bjorck2025gr00t, octo_2023}, navigation \citep{shah2023lm, shah23vinit}, aerial systems \citep{saviolo2024unifying}, and long-horizon planning \citep{driess2023palme}. These models have been used to bridge natural language instructions and visual input with physical plans and actions \citep{stone2023MOO}, generate code that can act as control policies \citep{liang2023code}, and construct reward functions \citep{yu2023language}. Various prompting strategies have been developed to elicit actionable knowledge from off-the-shelf VLMs, including iterative visual goal prompting \citep{nasiriany2024pivot}, keypoint prompting \citep{huang2025rekep, deitke2025molmo}, selecting from predefined skill libraries \citep{ichtersaycan2023}, and anticipating failure modes \citep{GanaiSinhaEtAl2025}. Vision-language-action (VLA) models trained end-to-end for robotics have been developed \citep{zitkovichRT2023, black25pi05}, but they often suffer from limited training data and lower generalization compared to large-scale VLM counterparts trained on internet-scale data \citep{radford2021learning}. These generalist policies face deployment challenges including latency, safety verification, lack of interpretability, and domain adaptation \citep{ren23knowno, sinha2024real}. 

\paragraph{Our approach}
In safety-critical domains such as maritime autonomy, regulatory frameworks like the IMO MASS Code mandate strict requirements on human oversight and explainability. We propose that pairing zero-shot semantic reasoning from off-the-shelf VLMs with the reliability of classical autonomy stacks offers a pragmatic path forward, leveraging the complementary strengths of foundation model reasoning and domain-specific perception-planning architectures.

\subsection{Robotic out-of-distribution detection and safety response}
    

\paragraph{OOD Detection} While deep learning systems have powered great advances in robot autonomy, such autonomous systems still struggle on rare and unexpected corner cases \citep{SinhaSharmaEtAl2022, geirhos}. Simply enumerating and controlling the ODD of ML-based robots is challenging, as a model's region of competence is implicit in its training data. Therefore, to alleviate the need for constant human supervision, many robotics works have proposed so-called out-of-distribution detectors in recent years  (e.g., see \cite{SharmaEtAl2021, LakshmiEtAl2017, RuffKauffmanEtAl2021}). These algorithms aim to detect anomalous conditions wherein model performance degrades so that a safety response can be triggered. While many works aim to detect component-level degradation, like when a perception model degrades in bad weather \citep{SinhaEtAl2023, GuptaEtAl2024, FilosTigas2020, RichterRoy2017}, some recent works introduce runtime monitors that detect contextual, semantic anomalies and reason about the appropriate safety intervention using LLMs and VLMs (e.g., see \cite{ElhafsiSinhaEtAl2023}). These works demonstrate promising results leveraging the general common-sense reasoning capabilities of foundation models (FMs) to handle scenarios that would otherwise require human judgment. Our goal is to validate these capabilities in the maritime domain and thereby establish anomaly handling as a valuable use case of FMs in maritime autonomy.

\paragraph{Maritime anomaly detection}
Maritime anomaly detection has mainly been studied for traffic surveillance and Vessel Traffic Service (VTS) decision support, where anomalies are deviations or inconsistencies in AIS-based trajectories and behavior \citep{riveiro2018maritime, wolsing2022anomaly, stach2023maritime}. Related work for MASS considers detecting abnormal surrounding vessels from AIS/GIS features \citep{tyasayumranani2022anomaly}, while onboard monitoring targets propulsion/engine-room faults from vibration and other machinery signals \citep{jeong2023study, oster2024human}. To our knowledge, no current work addresses onboard camera-based perception anomalies or FM-based contextual anomaly handling for maritime autonomy, which is our focus.

\paragraph{FM safety interventions} Two challenges must be addressed to practically integrate FMs into real-time decision making. First, querying state-of-the-art FMs like GPT-5 incurs significant latency and constantly querying such models is expensive. Therefore, we follow \citep{sinha2024real}, which proposes a ``thinking fast and slow'' approach wherein a fast, cheap anomaly detector runs in real-time and triggers the slower reasoning of a FM only in the event of an unusual situation. Second is to bridge the gap between the high-level, text-based reasoning of a VLM and the physical actions that the autonomy stack should execute to maintain safety. In \citep{knowno2023}, the authors assume the system can freeze in place to await a human takeover when the robot is uncertain. In contrast, other works execute fully autonomous recovery behaviors by making the FM choose from a predefined set of safety interventions \citep{sinha2024real} or by using additional VLMs to identify safe alternate goals \citep{GanaiSinhaEtAl2025}. 

\paragraph{Our approach} 
The existing methods cannot however be directly used due to the unique considerations in the maritime domain. In particular, a) the MASS Code  \citep{IMO2024_MSC109_5} necessitate a handover to a human, excluding fully autonomous recovery behaviors, and b) stopping a vessel to await a human takeover is not always possible or can even increase safety risks. Instead, we propose a novel framework wherein a VLM bridges the gap between the detection of an anomaly and operator takeover by reasoning over and executing short-term keep-safe behaviors.

%% file: 03_problem_formulation.tex
\section{Problem formulation}
\label{sec:problem}

\begin{table}[t]
    \centering
    \caption{Symbols and their meaning in the text. Different usages are explicitly stated when they appear.}
    {
    \footnotesize
    \begin{tabularx}{\columnwidth}{lXc}
        \hline
        \textbf{Symbol} & \textbf{Description} & \textbf{Unit} \\
        \hline
        $I$ & Calibrated camera image &  \\ 
        $\Omega$ & Image plane (pixel domain) & px \\
        $\mathbf{u}=(u,v)$ & Pixel coordinates in image plane $\Omega$ & px \\
        $\mathbf{q}$ & Auxiliary pixel coordinate used in clearance definition & px \\
        $W(\mathbf{u})$ & Binary water mask from segmentation ($1$ for water, $0$ for non-water), $W = \mathrm{Seg}(I)$ &  \\ 
        $D(\mathbf{u})$ & Pixel-space clearance distance to nearest non-water pixel; cf. Eq.\,\eqref{eq:clearance} & px \\ 
        $d_{\min}$ & Minimum clearance margin for feasible motion candidates & px \\ 
        $t_{\mathrm{alert}}$ & Time when anomaly alert is raised & s \\ 
        $\mathbf{p}^b=\begin{bmatrix}x&y&0\end{bmatrix}^\top$ & Point in body-fixed horizontal frame $B$ ($x$ forward, $y$ lateral) & m \\ 
        $\mathbf{p}^c$ & Point in camera frame (depth coordinate is $p^c_3$) & m \\ 
        $\mathbf{K}$ & Camera intrinsic matrix &  \\ 
        $\mathbf{R}^{n}_{b},\, \mathbf{r}^{b}_{nb}$ & Rotation and translation from body frame $\{b\}$ to world (NED) frame $\{n\}$ &  \\
        $\mathbf{R}^{c}_{b},\, \mathbf{r}^{b}_{cb}$ & Rotation and translation from body frame $\{b\}$ to camera frame $\{c\}$ &  \\
        $\Pi(\mathbf{K},\mathbf{R}^{c}_{b},\mathbf{r}^{b}_{cb},\mathbf{p}^b)$ & Projection from $\{b\}$ to image plane $\Omega$ (via camera extrinsics) &  \\
        $T_{N\leftarrow B}(t)$ & Navigation pose (body$\rightarrow$world) used for world anchoring at time $t_{\text{alert}}$ &  \\ 
        $\mathcal{C} = \{1,\dots,K\}$ & Index set of feasible motion candidates &  \\ 
        $a \in \{0, 1, \dots, K\}$ & Selected fallback action (0 = Station-keeping) &  \\ 
        $U_{\mathrm{anom}}$ & Vessel speed during anomaly mode & kn \\ 
        $R,\,\Delta$ & Acceptance radius and look-ahead distance for LOS path guidance & m \\ 
        $\boldsymbol{\tau}_\text{m}, \boldsymbol{\tau}_\text{h}, \boldsymbol{\tau}_\text{d}$ & Machine (autonomous), human, and desired (combined, shared) force input &  \\ 
        $\alpha$ & Override blending factor ($0$ = autonomy with immediate manual intervention, $1$ = full manual) &  \\
        \hline
    \end{tabularx}
    }
    \label{tab:symbols-used}
\end{table}

We consider an autonomous surface vessel (ASV) with a forward-facing monocular camera and a nominal waypoint-following system under human authority. An exogenous alert at time $t_{\text{alert}}$ triggers a one-shot fallback maneuver decision (this alert is based on the existing fast anomaly detection method adapted from~\cite{sinha2024real} and validated for the maritime domain in~\ref{app:e1}). Throughout the paper, a \emph{fallback maneuver} denotes the short-horizon, operator-overridable action executed in the alert-to-override interval, and the \emph{fallback maneuver selector} denotes the module that chooses it from a pre-vetted candidate set. During this alert, we assume that the system has the following information (see also Table\>\ref{tab:symbols-used} for description of symbols):
\begin{enumerate}
    \item A calibrated camera image $I$,
    \item an identified water mask $W=\mathrm{Seg}(I)$, where $\mathrm{Seg}()$ defines the segmentation function returning 1 for water and 0 for anything else, 
    \item a pixel-space clearance map on $\Omega$, i.e., the minimal distance (in the image plane) to shore or other identified objects in the water, given by
    \begin{equation}
    D(\mathbf{u}) \;=\; \min_{\mathbf{q}\in\Omega:\, W(\mathbf{q})=0}\, \|\mathbf{u}-\mathbf{q}\|,
    \label{eq:clearance}
    \end{equation}
\end{enumerate}
From this, we generate a finite set of $K$ short, straight, and feasible motion primitives proposed ahead of the vessel. The feasible motion candidates are paths in the water and contain a minimum safe margin. Their projected samples (via the known camera model) satisfy for all visible samples $\mathbf{u}$,
\begin{equation}
W(\mathbf{u})=1
\quad\text{and}\quad
D(\mathbf{u})\ge d_{\min},
\label{eq:gating}
\end{equation}
for a margin $d_{\min}$ (pixels). The retained set of motion primitives is indexed $\{1,\ldots,K\}$, with $0$ reserved to Station-keeping. 

\paragraph{Problem formulation} The fallback maneuver selector selects a single action $a\in\{0,\ldots,K\}$ once at $t_{\text{alert}}$ from the overlaid view; no replanning occurs within the episode. The selected action is executed at a cautious anomaly-mode speed until termination by human override or alert clearance; operator authority strictly dominates autonomy. 

We assume camera-only perception and start-from-rest. We evaluate (i) \emph{human alignment} of the chosen action (and scene understanding) and (ii) \emph{short-horizon directional risk relief} on the same overlays.

%% file: 04_method.tex
\section{Methods: Fallback maneuver selection}
\label{sec:methods}

Here we show how our implementation instantiates the one-shot fallback maneuver selection problem posed in Section~\ref{sec:problem} as a concrete, camera-only method executed at the alert rising edge $t_{\text{alert}}$. From a single calibrated camera frame, we (i) compute a conservative water mask and a pixel-space clearance map, (ii) generate and gate short straight motion trajectories in the boat frame, (iii) render a numbered overlay ($\{1,\ldots,K\}$ with $0$ denoting Station-keeping), (iv) query a vision–language model (VLM) once to select a fallback action, (v) and publish either a Station-keeping point or a world-fixed path to a line-of-sight waypoint follower. Operator input always has priority (MANUAL $>$ AUTONOMY). The same overlay/gating stack is used both for offline evaluation and closed-loop field testing, with only vehicle-specific execution and override being different (see details in section~\ref{sec:experiments}). The remainder of this section is divided into the following three components: \emph{Trajectory candidate generation and gating} (Section~\ref{sec:methods-cands}), \emph{VLM fallback maneuver selection} (Section~\ref{sec:methods-bridge}), and \emph{Execution and arbitration (Section~\ref{sec:methods-control})}, as illustrated in Fig.~\ref{fig1}. 

\subsection{Trajectory candidate generation and gating}
\label{sec:methods-cands}

At the alert rising edge $t_{\text{alert}}$, we operate on a single calibrated forward image $I$. We use four coordinate frames, see Fig.\>\ref{fig:boat-frames}: the image plane $\Omega\subset\mathbb{Z}^2$ with pixel coordinates $\mathbf{u}=(u,v)$, a body-fixed horizontal frame $\{b\}$ with $\mathbf{p}^b=\begin{bmatrix}x&y&0\end{bmatrix}^\top$ ($x$ forward, $y$ lateral starboard), a camera frame $\{c\}$, and a local North–East–Down (NED) world frame $\{n\}$. The calibrated projection from frame $\{b\}$ to $\Omega$ uses an intrinsic matrix $\mathbf{K}$, and the transformation from body frame $\{b\}$ to world frame $\{n\}$ uses an extrinsic transformation by translation $\mathbf{r}_{nb}^b$ (position of the vessel relative to the NED-frame given in the body frame) and rotation matrix $\mathbf{R}_{b}^n \in \text{SO}(3)$:

\begin{equation}
\begin{aligned}
\mathbf{p}^c &= \mathbf{R}^{c}_{b}\bigl(\mathbf{p}^b - \mathbf{r}^{b}_{cb}\bigr),\\
\mathbf{u}   &= \Pi(\mathbf{K},\mathbf{p}^c)
               := \frac{(\mathbf{K}\,\mathbf{p}^c)_{1:2}}{(\mathbf{K}\,\mathbf{p}^c)_3},\\
\mathbf{p}^n &= \mathbf{R}^{n}_{b}\bigl(\mathbf{p}^b - \mathbf{r}^{b}_{nb}\bigr).
\end{aligned}
\label{eq:projection}
\end{equation}
The above is valid when $p^c_3>0$ and $\mathbf{u}\in\Omega$ (in front of the camera, in-frame). All selection and gating in this subsection happens in $\Omega$, putting candidates into $\{n\}$ happens later in Sec.~\ref{sec:methods-bridge}.

{\graphicspath{{./}}
\begin{figure}[t!]
    \centering
    {
    \scriptsize
    \def\svgwidth{0.76\columnwidth}
    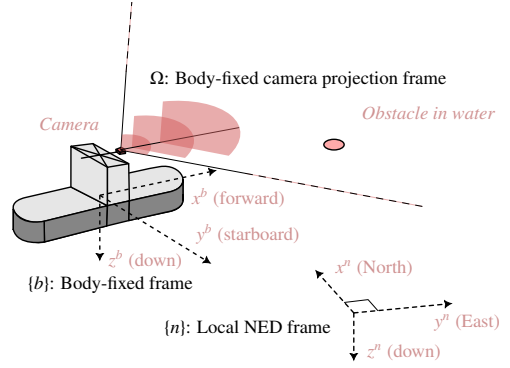
    }
    \caption{Coordinate frames and their relation. The water segmentation filters out any non-water obstacle in the field-of-view, in the projection $\Omega$.}
    \label{fig:boat-frames}
\end{figure}}

\paragraph{Water and clearance}
From the image $I$ we compute a conservative binary water mask
\begin{equation}
W(\mathbf{u})\in\{0,1\}, \qquad W=\mathrm{Seg}(I),
\label{eq:water}
\end{equation}
using the \emph{embedded Water Segmentation and Refinement} (eWaSR) maritime segmentation model~\citep{Terek2023}, and a per-pixel clearance map (Euclidean distance to the nearest non-water pixel) using Eq.\>\eqref{eq:clearance}. A narrow bottom band is treated as trivially water to ensure connectivity at the hull.

\paragraph{Sampling and projection}
We sample $N_{\text{raw}}$ straight motion primitives in $\{b\}$ from a fixed local anchor at the bow $(4.0,0.0)$\,m to endpoints drawn in an annulus $r\in[R_{\min},R_{\max}]$ within a forward half-angle $|\phi|\le\phi_{\max}$. Each primitive is discretized into $N_\ell$ points in $B$ and projected via \eqref{eq:projection}. A sample is visible if $p_3^c>0$ and $\mathbf{u}\in\Omega$. Let $k_0$ be the first visible index; the endpoint pixel must also be visible.

\paragraph{Pixel-space gating}
A candidate is retained if every visible projected sample from $k_0$ to the endpoint satisfies
\begin{equation}
W(\mathbf{u}_k)=1
\quad\text{and}\quad
D(\mathbf{u}_k)\ge d_{\min},
\qquad \forall\, k\in\{k_0,\ldots,N_\ell\}.
\label{eq:gating_cand}
\end{equation}
This enforces that the commanded segment lies entirely on water with a pixel margin, without requiring world geometry or multi-sensor fusion.

\paragraph{Thinning and indexing}
From the surviving set, we select up to $K$ primitives by farthest-point thinning on endpoint pixels to promote spatial spread (Appendix, Alg.~\ref{alg:farthest}). Candidates are indexed as $\{1,\ldots,K\}$ (ID~0 is reserved for Station-keeping). We use $K=15$ and $d_{\min}=40$\,px in both offline evaluation and live field tests. Here $d_{\min}=40$ px is an image-space clearance margin used as a conservative camera-only proxy to reject candidates whose projected samples pass too close to non-water regions. A fixed pixel margin corresponds to a larger physical separation for more distant boundaries, so it is most critical in the near field. In a production system this gating module would naturally be replaced or augmented by range-aware perception (e.g., stereo, LiDAR, radar) without changing the remainder of the Semantic Lookout fallback maneuver selector.

The whole generation and gating algorithm is summarized in Alg.~\ref{alg:candidates}. 

\begin{algorithm}[t]
\small
\caption{Candidate generation, projection, gating, and thinning}
\label{alg:candidates}
\begin{algorithmic}[1]
\REQUIRE Image $I$ at $t_{\text{alert}}$; camera $(\mathbf{K},\mathbf{R}^{c}_{b},\mathbf{r}^{b}_{cb})$; $R_{\min},R_{\max},\phi_{\max}$; $N_{\text{raw}},N_\ell$; $d_{\min}$; target $K$
\STATE $W \leftarrow \mathrm{Seg}(I)$
\STATE $D \leftarrow$ Euclidean distance transform on $(1-W)$, sampled on $W{=}1$
\STATE $\mathcal{S}\leftarrow\emptyset$
\FOR{each sampled endpoint $e$ in annulus $(R_{\min},R_{\max},\phi_{\max})$}
    \STATE $\text{line}_b \leftarrow$ $N_\ell$ samples from anchor $(4.0,0.0)$ to $e$ in $B$
    \STATE $\text{pix\_seq} \leftarrow$ project all samples via \eqref{eq:projection}
    \STATE $k_0 \leftarrow$ first index with a valid pixel; require endpoint pixel valid
    \IF{$k_0$ exists \AND $\forall k\in\{k_0,\ldots,N_\ell\}:\ W(\mathbf{u}_k){=}1 \land D(\mathbf{u}_k)\ge d_{\min}$}
        \STATE $\mathcal{S}\leftarrow \mathcal{S}\cup\{(e,\mathbf{u}_{\text{end}})\}$ \hfill (store endpoint and its pixel)
    \ENDIF
\ENDFOR
\STATE $\mathcal{C} \leftarrow \mathrm{FarthestPointThinning}(\mathcal{S},K,\delta_{\mathrm{px}})$ \hfill (Appendix Alg.~\ref{alg:farthest})
\STATE Index $\mathcal{C}$ as $\{1,\ldots,|\mathcal{C}|\}$ and define ID~0 $\equiv$ Station-keeping
\RETURN $\mathcal{C}$ (for overlay rendering)
\end{algorithmic}
\end{algorithm}

\subsection{VLM fallback maneuver selector}
\label{sec:methods-bridge}

\paragraph{Input and output schema}
The model receives only the overlayed image $O$ with labeled candidates $\{1,\ldots,K\}$ (ID~0 denotes Station-keeping) and an instruction prompt (see~\ref{app:misc}). It returns a single strict JSON object
\begin{verbatim}
{"see":"<= 15 words",
 "implications":"<= 15 words",
 "action":"<= 15 words (no ids)",
 "choice_id": 0..K,
 "confidence": 0..1}
\end{verbatim}
We parse and connect \texttt{choice\_id} to $\{0,\ldots,K\}$ (invalid JSON maps to Station-keeping (\texttt{choice\_id} = 0). The free-text fields (\texttt{see}, \texttt{implications}, \texttt{action}) are logged for analysis and are also used with the operator UI to support situational awareness in the formative handover study (Sec.~\ref{sec:formative_handover}).

\paragraph{Failure handling and safe defaults}
The fallback maneuver selector is invoked only after an external anomaly alert at $t_{\mathrm{alert}}$. It is therefore a \emph{post-alert} safety module: it does not guarantee anomaly detection, and missed or late alerts are handled by the nominal autonomy stack and the upstream alerting monitor (\ref{app:e1}). Within the fallback maneuver selector itself, we implement conservative safe defaults for three failure modes. (i) If the VLM call times out, returns an API error, or produces an invalid or non-conforming JSON object, we set $\texttt{choice\_id}=0$ (Station-keeping) and notify the ROC. (ii) If the candidate generation/gating yields no feasible motion candidates (i.e., $K=0$ after gating), we likewise default to Station-keeping and notify the ROC. (iii) If voting in FB-$n$ produces no strict majority (ties, see below), we default to Station-keeping by design. In all cases, joystick override remains available at all times and dominates autonomy.

\paragraph{Runtime single-shot selection (FB-1)}
We issue one call per alert. Let $f_\theta$ denote the VLM and let $y=f_\theta(O)$; the executed action is
\begin{equation}
a^{(1)} \;=\; \mathrm{clip}_{\{0,\ldots,K\}}\!\bigl(f_\theta(O).\texttt{choice\_id}\bigr).
\label{eq:br1}
\end{equation}
and is applied once within the episode (no replanning).

\paragraph{Evaluation ensemble (FB-$n$)}
To probe robustness, we evaluate $n$ independent calls on the same overlay $O$ with distinct seeds $\{\eta_s\}_{s=1}^n$:
\begin{equation}
a_s \;=\; \mathrm{clip}_{\{0,\ldots,K\}}\!\bigl(f_\theta(O;\eta_s).\texttt{choice\_id}\bigr),\quad s=1,\ldots,n.
\label{eq:br3-seeds}
\end{equation}
We aggregate by strict majority of $n$ (ties $\Rightarrow$ Station-keeping):
\begin{equation}
a^{(n)} \;=\;
\begin{cases}
\arg\max_k c_k, & \max_k c_k \;>\; \lfloor n/2 \rfloor,\\[2pt]
0, & \text{otherwise.}
\end{cases}
\label{eq:br3}
\end{equation}

\noindent\textit{Where} $a_s\in\{0,\ldots,K\}$ is the ID from the $s$-th stochastic call, 
$c_k=\lvert\{\,s:\,a_s=k\,\}\rvert$ counts votes for candidate $k$, 
$\arg\max_k c_k$ returns any maximizer, and ID $0$ denotes Station‑keeping.
We refer to the runtime selector as FB‑1 and the evaluation ensemble as FB‑$n$. In all experiments, we use $n{=}3$ (FB‑3, see Sec.~\ref{sec:experiments}). 
Sampling multiple test‑time rollouts and aggregating by majority follows standard self‑consistency / test‑time compute‑scaling practice in LLM reasoning~\citep{wang2022self}.

\paragraph{Temporal anchoring}
Candidates are frozen in the world frame at $t_{\text{alert}}$ using the navigation pose $T_{N\leftarrow B}(t_{\text{alert}})$. At command issue, we re-project those world-fixed polylines for visualization and publish the world-fixed geometry to the controller. This decouples perception/decision latency from actuation while preserving the action space from Sec.~\ref{sec:methods-cands}.

The whole fallback maneuver selector algorithm is summarized in Alg.~\ref{alg:bridge}. 

\begin{algorithm}[t]
\small
\caption{fallback maneuver selection and command issue (FB-1 runtime; FB-$n$ evaluation)}
\label{alg:bridge}
\begin{algorithmic}[1]
\REQUIRE Overlay $O$ with candidate count $K$; frozen world polylines $\{\Pi_k\}_{k=1}^K$; current navigation pose; ensemble size $n \in \mathbb{N}, n\ge 1$
\STATE \textbf{FB-1 (runtime):} $y \leftarrow f_\theta(O)$; \quad $a \leftarrow \mathrm{clip}_{\{0,\ldots,K\}}(y.\texttt{choice\_id})$
\STATE \textbf{FB-$n$ (evaluation):} \textbf{for} $s \in \{1,\ldots,n\}$ \textbf{do}
\STATE \hspace{1.6em} $a_s \leftarrow \mathrm{clip}_{\{0,\ldots,K\}}\!\bigl(f_\theta(O;\eta_s).\texttt{choice\_id}\bigr)$
\STATE \textbf{end for}
\STATE \textbf{(vote counts)} $c_k \leftarrow \bigl|\{\,s: a_s = k\,\}\bigr|$ for all $k \in \{0,\ldots,K\}$
\STATE \textbf{(strict majority)} \textbf{if} $\max_k c_k > \lfloor n/2 \rfloor$ \textbf{then} $a \leftarrow \arg\max_k c_k$ \textbf{else} $a \leftarrow 0$ \textbf{end if} \hfill (ties $\Rightarrow$ Station-keeping)
\IF{$a = 0$}
  \STATE \textbf{Station-keeping:} publish a single world point at the current nav pose
\ELSE
  \STATE \textbf{Track:} publish world-fixed polyline $\Pi_a$ (to LOS follower)
\ENDIF
\end{algorithmic}
\end{algorithm}

\subsection{Execution, arbitration, and control}
\label{sec:methods-control}

\begin{figure*}[!t]
  \centering
  \footnotesize
  \resizebox{0.75\textwidth}{!}{\input{TikZ_control-allocation}}
  \caption{Block diagram for automatic waypoint follow with motion guidance on the ASV. A combined motion is the result of an override logic in which a human-in-the-loop may override the automatic motion guidance.}
  \label{fig:MA-waypoint-architecture}
\end{figure*}
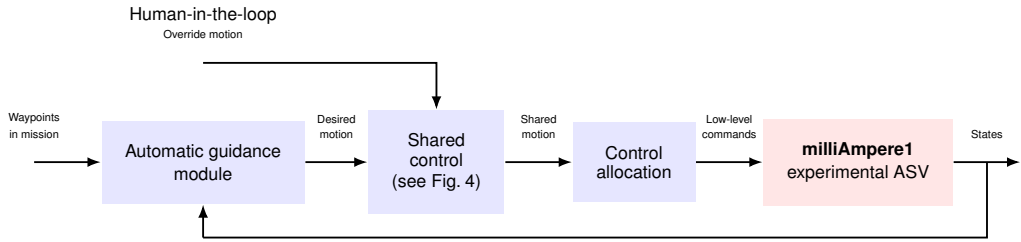

\begin{figure}[!htbp]
    \centering
    \input{TikZ_override-switch}
    \vspace{-1em}
    \caption{Override switch logic implemented from autonomous operation to manual intervention and back. The percentage is the weight given to each actor, showing that during autonomy operations, the operator may intervene in the motion with both immediate action and with a gradual complete override by a defined phase-in time period.}
    \label{fig:override-illustration}
\end{figure}
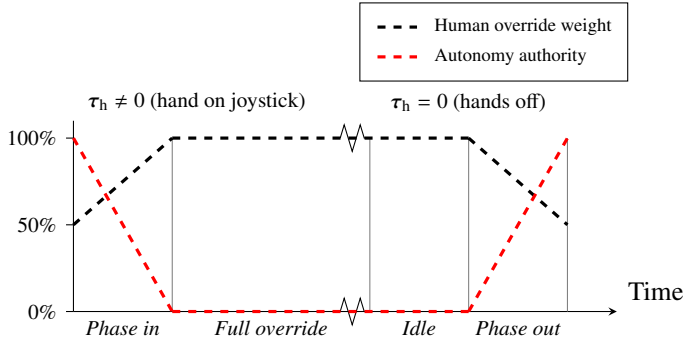

This section summarizes the closed-loop elements used only in the field experiments and formative handover study. Given the fallback maneuver $a\in\{0,\ldots,K\}$ (Sec.~\ref{sec:methods-bridge}), the system either tracks a world-fixed path ($a\ge 1$) or holds position ($a=0$, Station-keeping). Figures~\ref{fig:MA-waypoint-architecture} and \ref{fig:override-illustration} illustrate the overall architecture and the override blend. Allocator and platform details are provided in~\ref{app:system}.

\paragraph{Waypoint following}
For $a\ge 1$ the pre-frozen world polyline $\Pi_a$ is tracked by a line-of-sight (LOS) follower with acceptance radius $R=7.5$\,m and lookahead distance $\Delta=10$\,m at the anomaly-mode speed $U_{\mathrm{anom}}=1.0$\,kn. Full controller and allocation details follow~\cite{Tufte2025IntegratedMotion} and~\ref{app:platform}. For $a=0$ a single world point at the current navigation pose is published.

\paragraph{Operator override}
Let $\boldsymbol{\tau}_d$ denote the desired actuation after arbitration, $\boldsymbol{\tau}_m$ the autonomy (motion) command, $\boldsymbol{\tau}_h$ the human (joystick) input, and $\alpha\in[0,1]$ a blending factor (see Fig.~\ref{fig:override-illustration}). We use a direct blend that guarantees at least 50\% human authority and ramps to full override:
\begin{equation}
    \boldsymbol{\tau}_\text{d} = (1-\alpha) \boldsymbol{\tau}_\text{m} + (0.5 + 0.5 \alpha)\boldsymbol{\tau}_\text{h}
\end{equation}
Here, the blending factor $\alpha$ determines
\begin{equation*}
\alpha = \quad
\begin{cases}
0, & \text{Automatic steering} \\
0 < \alpha < 1, & \text{Shared steering} \\
1, & \text{Manual override}
\end{cases}
\end{equation*}
We do this to enable continuous control and a seamless transition between vessel and operator in Sec.~\ref{sec:formative_handover}, and to revert back to fully autonomous control in case of loss of signal mid-transition. Details of the $\alpha$ schedule transitions are specified in\>\ref{app:platform}.


%% file: boat-frames.pdf_tex
\begingroup%
  \makeatletter%
  \providecommand\color[2][]{%
    \errmessage{(Inkscape) Color is used for the text in Inkscape, but the package 'color.sty' is not loaded}%
    \renewcommand\color[2][]{}%
  }%
  \providecommand\transparent[1]{%
    \errmessage{(Inkscape) Transparency is used (non-zero) for the text in Inkscape, but the package 'transparent.sty' is not loaded}%
    \renewcommand\transparent[1]{}%
  }%
  \providecommand\rotatebox[2]{#2}%
  \newcommand*\fsize{\dimexpr\f@size pt\relax}%
  \newcommand*\lineheight[1]{\fontsize{\fsize}{#1\fsize}\selectfont}%
  \ifx\svgwidth\undefined%
    \setlength{\unitlength}{170.41023314bp}%
    \ifx\svgscale\undefined%
      \relax%
    \else%
      \setlength{\unitlength}{\unitlength * \real{\svgscale}}%
    \fi%
  \else%
    \setlength{\unitlength}{\svgwidth}%
  \fi%
  \global\let\svgwidth\undefined%
  \global\let\svgscale\undefined%
  \makeatother%
  \begin{picture}(1,0.70842898)%
    \lineheight{1}%
    \setlength\tabcolsep{0pt}%
    \put(0,0){\includegraphics[width=\unitlength,page=1]{boat-frames.pdf}}%
    \put(0.26925501,0.54719993){\color[rgb]{0,0,0}\makebox(0,0)[lt]{\lineheight{1.25}\smash{\begin{tabular}[t]{l}$\Omega$: Body-fixed camera projection frame\end{tabular}}}}%
    \put(0.68294767,0.47902151){\color[rgb]{0.63921569,0.17254902,0.17254902}\transparent{0.51741308}\makebox(0,0)[lt]{\lineheight{1.25}\smash{\begin{tabular}[t]{l}\textit{Obstacle in water}\end{tabular}}}}%
    \put(0.05246202,0.45646568){\color[rgb]{0.63921569,0.17254902,0.17254902}\transparent{0.51741308}\makebox(0,0)[lt]{\lineheight{1.25}\smash{\begin{tabular}[t]{l}\textit{Camera}\end{tabular}}}}%
    \put(0.0303281,0.14355087){\color[rgb]{0,0,0}\makebox(0,0)[lt]{\lineheight{0}\smash{\begin{tabular}[t]{l}$\{b\}$: Body-fixed frame\end{tabular}}}}%
    \put(0.29671125,0.05746048){\color[rgb]{0,0,0}\makebox(0,0)[lt]{\lineheight{0}\smash{\begin{tabular}[t]{l}$\{n\}$: Local NED frame\end{tabular}}}}%
    \put(0.34956317,0.31097853){\color[rgb]{0.63921569,0.17254902,0.17254902}\transparent{0.51741308}\makebox(0,0)[lt]{\lineheight{1.25}\smash{\begin{tabular}[t]{l}$x^b$ (forward)\end{tabular}}}}%
    \put(0.62996332,0.17482638){\color[rgb]{0.63921569,0.17254902,0.17254902}\transparent{0.51741308}\makebox(0,0)[lt]{\lineheight{1.25}\smash{\begin{tabular}[t]{l}$x^n$ (North)\end{tabular}}}}%
    \put(0.8247408,0.06896759){\color[rgb]{0.63921569,0.17254902,0.17254902}\transparent{0.51741308}\makebox(0,0)[lt]{\lineheight{1.25}\smash{\begin{tabular}[t]{l}$y^n$ (East)\end{tabular}}}}%
    \put(0.69337021,0.0100392){\color[rgb]{0.63921569,0.17254902,0.17254902}\transparent{0.51741308}\makebox(0,0)[lt]{\lineheight{1.25}\smash{\begin{tabular}[t]{l}$z^n$ (down)\end{tabular}}}}%
    \put(0.3596277,0.23936937){\color[rgb]{0.63921569,0.17254902,0.17254902}\transparent{0.51741308}\makebox(0,0)[lt]{\lineheight{1.25}\smash{\begin{tabular}[t]{l}$y^b$ (starboard)\end{tabular}}}}%
    \put(0.1880149,0.18894986){\color[rgb]{0.63921569,0.17254902,0.17254902}\transparent{0.51741308}\makebox(0,0)[lt]{\lineheight{1.25}\smash{\begin{tabular}[t]{l}$z^b$ (down)\end{tabular}}}}%
    \put(0,0){\includegraphics[width=\unitlength,page=2]{boat-frames.pdf}}%
  \end{picture}%
\endgroup%

%% file: TikZ_control-allocation.tex
\begin{tikzpicture}[
    block/.style = {rectangle, draw, thick, minimum height=2em, minimum width=1cm, align=center},
    arrow/.style = {thick, -{Latex[length=2mm]}},
    node distance=1cm,
    font=\sffamily\footnotesize, 
    inner sep=10pt,      
]

\node[coordinate] (input) {};
\node[block, right=of input, fill=blue!10, draw=none] (auto) {\shortstack{Automatic guidance \\ module}};
\node[block, draw=none,
      right of=auto, node distance=3.5cm, fill=blue!10] (sum) {\shortstack{Shared \\ control \\ (see Fig.\>\ref{fig:override-illustration})}};
\node[coordinate, above of=auto, node distance=1.5cm] (manual_input) {};
\node[block, right=of sum, draw=none, fill=blue!10] (assist) {\shortstack{Control \\ allocation}};
\node[block, right=of assist, draw=none, fill=red!10] (craft) {\shortstack{\textbf{milliAmpere1} \\ experimental ASV}};
\node[coordinate, right=of craft] (output) {};
\node[coordinate, right=0.5cm of craft] (feedback) {};
\node[coordinate, below=0.5cm of auto] (below_law) {};

\draw[arrow] (input) -- node[pos=0, above] {\tiny \shortstack{Waypoints \\in  mission}} (auto);
\draw[arrow] (manual_input) -| node[pos=0, above] {\shortstack{Human-in-the-loop \\ \tiny Override motion}} (sum);
\draw[arrow] (auto) -- node[above] {\tiny \shortstack{Desired \\ motion}} (sum);
\draw[arrow] (sum) -- node[above] {\tiny \shortstack{Shared \\ motion}} (assist);
\draw[arrow] (assist) -- node[above] {\tiny \shortstack{Low-level \\ commands}
} (craft);
\draw[arrow] (craft) -- node[above] {\tiny States} (output);
\draw[-, thick] (feedback) |- (below_law);
\draw[arrow] (below_law) -> (auto);



\end{tikzpicture}

%% file: TikZ_override-switch.tex
\usetikzlibrary{decorations.pathmorphing, matrix}
\usetikzlibrary{matrix,positioning}

\begin{tikzpicture}[x=0.65cm,y=0.023cm,>=stealth]
\def\xa{0}   
\def\xb{2}   
\def\xc{6}   
\def\xd{8}   
\def\xe{10}  
\def\ymax{110}

\draw[->] (\xa,0) -- (\xe+1,0) node[above right=1pt] {Time};
\draw (\xa,0) -- (\xa,\ymax);
\foreach \y/\lab in {0/0\%,50/50\%,100/100\%}
  {\draw (\xa,\y) -- ++(-0.15,0) node[left] {\footnotesize \lab};}

\foreach \x in {\xb,\xc,\xd,\xe}
  \draw[gray] (\x,0) -- (\x,100);

\node[below] at ({(\xa+\xb)/2},0) {\footnotesize \textit{Phase in}};
\node[below] at ({(\xb+\xc)/2},0) {\footnotesize \textit{Full override}};
\node[below] at ({(\xc+\xd)/2},0) {\footnotesize \textit{Idle}};
\node[below] at ({(\xd+\xe)/2},0) {\footnotesize \textit{Phase out}};

\node[above] at ({(\xa+\xc)/2-0.5},110) {\footnotesize $\boldsymbol{\tau}_{\rm h}\neq 0$ (hand on joystick)};
\node[above] at ({(\xd)},110) {\footnotesize $\boldsymbol{\tau}_{\rm h}=0$ (hands off)};

\draw[decorate,decoration={zigzag,segment length=6,amplitude=5}]
    (\xc,100) ++(-0.6,0) -- ++(0.6,0);
\draw[decorate,decoration={zigzag,segment length=6,amplitude=5}]
    (\xc,0)   ++(-0.6,0) -- ++(0.6,0);

\draw[very thick,dashed]
  plot[smooth] coordinates {(\xa,50) (\xb,100)};
\draw[very thick,dashed]
  plot[smooth] coordinates {(\xb,100) (\xd,100)};
\draw[very thick,dashed]
  plot[smooth] coordinates {(\xd,100) (\xe,50)};

\draw[very thick,red,dashed]
  plot[smooth] coordinates {(\xa,100) (\xb,0)};
\draw[very thick,red,dashed]
  plot[smooth] coordinates {(\xb,0) (\xd,0)};
\draw[very thick,red,dashed]
  plot[smooth] coordinates {(\xd,0) (\xe,100)};

\matrix[matrix of nodes,
        draw, fill=white,
        nodes={anchor=west,font=\footnotesize},
        column sep=0pt, row sep=-1pt,
        below left, inner sep=3pt, xshift=-0.5cm, yshift=-0.5cm]
      at (\xe+2,\ymax+90)
{
  \tikz{\draw[very thick,dashed] (0,0) -- (0.8,0);} & \scriptsize Human override weight \\
  \tikz{\draw[very thick,red,dashed] (0,0) -- (0.8,0);} & \scriptsize Autonomy authority \\
};

\end{tikzpicture}

%% file: 05_experiments.tex
\section{Experiments and results}
\label{sec:experiments}

In this section we evaluate the method on camera overlays at alert time, using the candidate generation and gating stack (and control/arbitration where applicable) described in Section~\ref{sec:methods}. We structure the evaluation around a series of experiments that test the following four hypotheses:

\begin{itemize}
    \item \textbf{H1 (Scene understanding).} 
    Modern VLMs can correctly recognize semantic, marine-specific, hazards and their implications, and propose appropriate high–level action descriptions, at latencies that are usable in the alert$\rightarrow$override handover window.

    \item \textbf{H2 (Action alignment with human choices).} 
    The corresponding fallback maneuvers selected from these action descriptions align with aggregated human “Acceptable/Best’’ judgments on the same overlays, outperforming simple geometry-only baselines.

    \item \textbf{H3 (Risk–relief).} 
    On unambiguously dangerous fire scenes, the fallback maneuver selector selects short–horizon actions that increase separation from the hazard relative to keep-course/starboard geometry-only baselines.

    \item \textbf{H4 (Integrated field test and handover).} 
    The full alert$\rightarrow$fallback maneuver$\rightarrow$override chain can be executed on a real vessel with immediate joystick override and a ROC-legible user interface (UI), consistent with IMO MASS Code constraints.

\end{itemize}

\paragraph{Experiment rationale}
We run four main experiments, corresponding to each respective hypothesis. Experiment~1 (Sec.~\ref{sec:exp-reasoning}) evaluates scene understanding: given a single overlaid camera frame and a strict output schema, can models describe the hazard, its implication, and a safe high-level action, and how does this trade off with latency? Experiment~2 (Sec.~\ref{sec:exp-action}) evaluates to what extent the  selector’s chosen trajectory IDs under FB-3 agree with human judgments on the same overlays, using aggregated Accept/Best sets to approximate “reasonable” actions in semantic anomalies. Experiment~3 (Sec.~\ref{sec:exp-riskrelief}) focuses on the unambiguously dangerous fire subset, measuring short-horizon directional risk relief versus geometry-only baselines. Experiment~4 (Sec.~\ref{sec:exp-field}) closes the loop on water, testing the full alert$\rightarrow$fallback maneuver$\rightarrow$override chain on a real ASV with ROC supervision.

Each (model, scene) is run $n$ times with distinct seeds. Unless stated otherwise, we use $n{=}3$ (FB-3) and aggregate by strict majority (ties $\Rightarrow$ Station-keeping). Single–shot results (FB-1) are reported where relevant. Models receive only the numbered overlay image and an instruction prompt, no additional geometry or labels are provided at inference time.

In~\ref{app:e1}, we also evaluate the fast alert of our alert$\rightarrow$fallback maneuver$\rightarrow$operator system (an established fast anomaly detection alert method adapted from \cite{sinha2024real}), to validate that it also works, given the distribution shift for the maritime domain.

\subsection{Shared dataset, platform, and setup}
\label{sec:exp-setup}

\begin{figure}[!htbp]
  \centering
  \includegraphics[width=\linewidth]{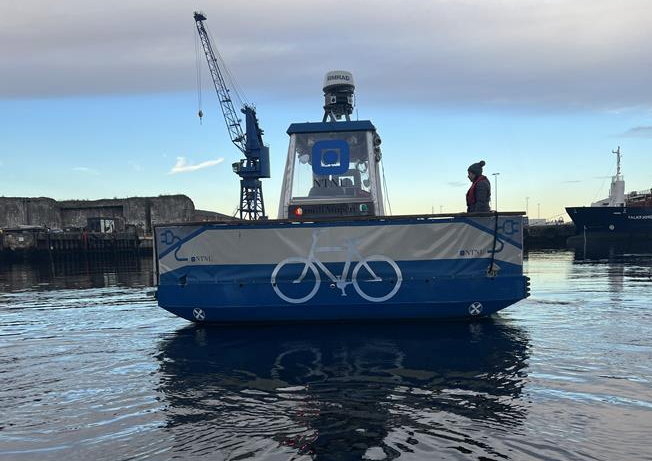}
  \caption{The Milliampere research ASV was used for data collection and in closed-loop experiments. Details in~\ref{app:platform}.}
  \label{fig:ma1}
\end{figure}

\begin{figure}[!htbp]
  \centering
  \includegraphics[width=\linewidth]{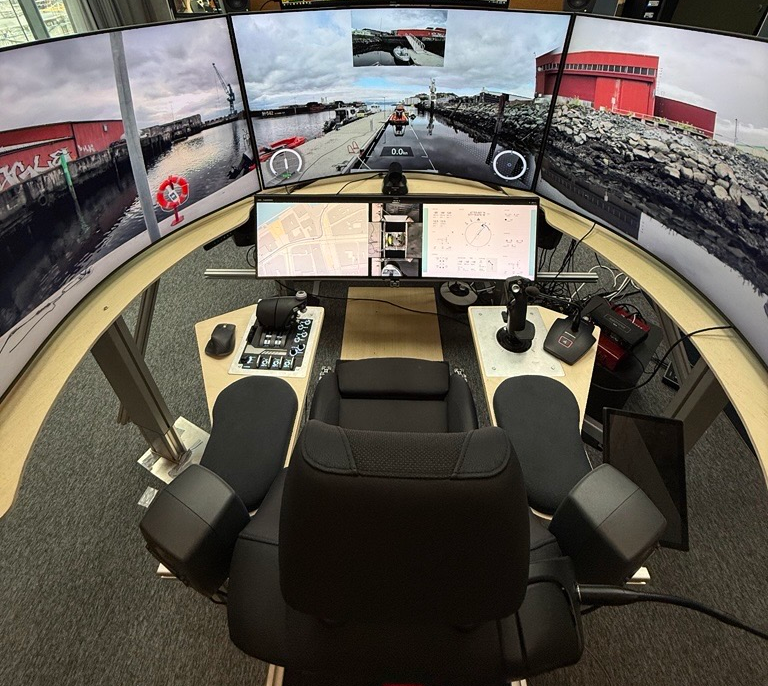}
  \caption{The remote operation center used in closed-loop experiments. Details in~\ref{app:platform}.}
  \label{fig:roc}
\end{figure}

\paragraph{Platform and environment}
All scenes are captured from the milliAmpère research ASV operating in a sheltered harbor, shown in Fig.~\ref{fig:ma1}. The closed-loop live experiment is performed with this vessel and the ROC shown in Fig.~\ref{fig:roc}. Platform, control, and ROC details are summarized in~\ref{app:system}. The offline overlays used in this section are produced by the same camera/overlay/gating stack that is used in field tests, only the execution/override differs. Specifically, Experiments~1 and~2 are performed on the offline dataset detailed below. Experiment~3 is performed on the fire subset of that data, where movement is simulated along the trajectory in the vessel frame. Experiment~4 is performed closed-loop on water with real human manual override. 

\begin{figure*}[!t]
  \centering

  \begin{subfigure}{0.49\textwidth}
    \centering
    \includegraphics[width=\linewidth]{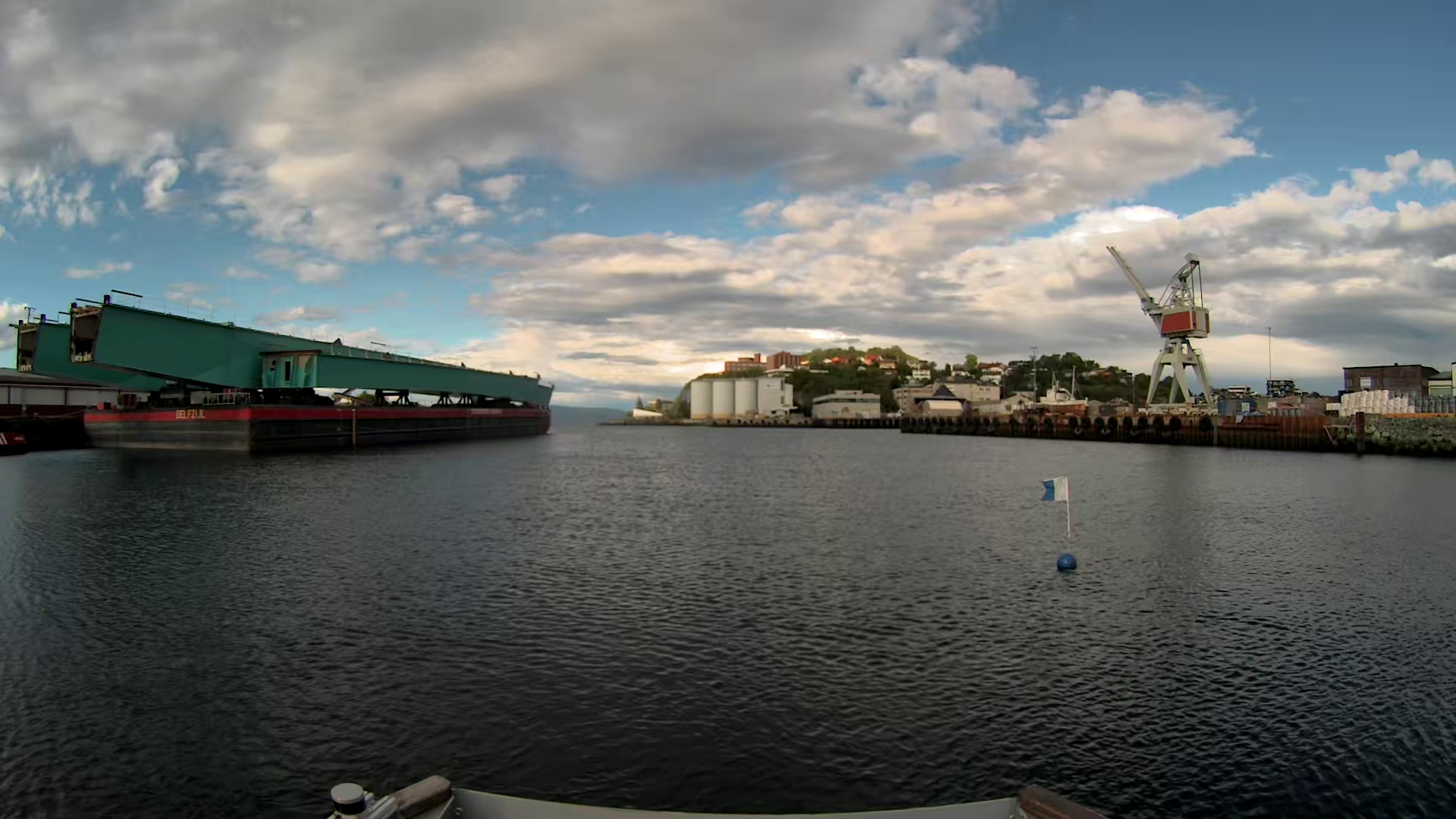}
    \caption{Diver flag.}
  \end{subfigure}\hfill
  \begin{subfigure}{0.49\textwidth}
    \centering
    \includegraphics[width=\linewidth]{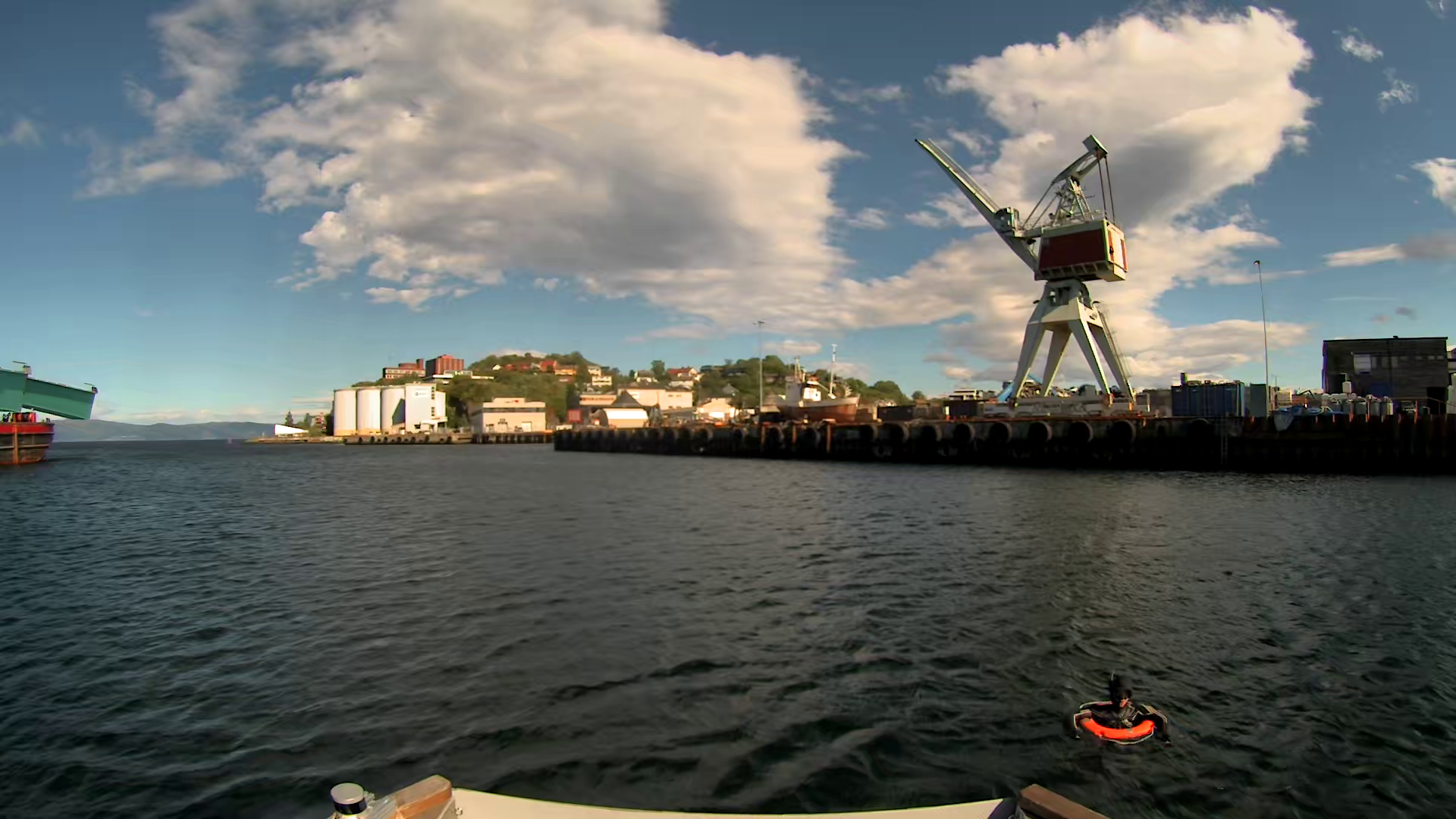}
    \caption{Man overboard (MOB).}
  \end{subfigure}

  \vspace{0.25em}

  \begin{subfigure}{0.49\textwidth}
    \centering
    \includegraphics[width=\linewidth]{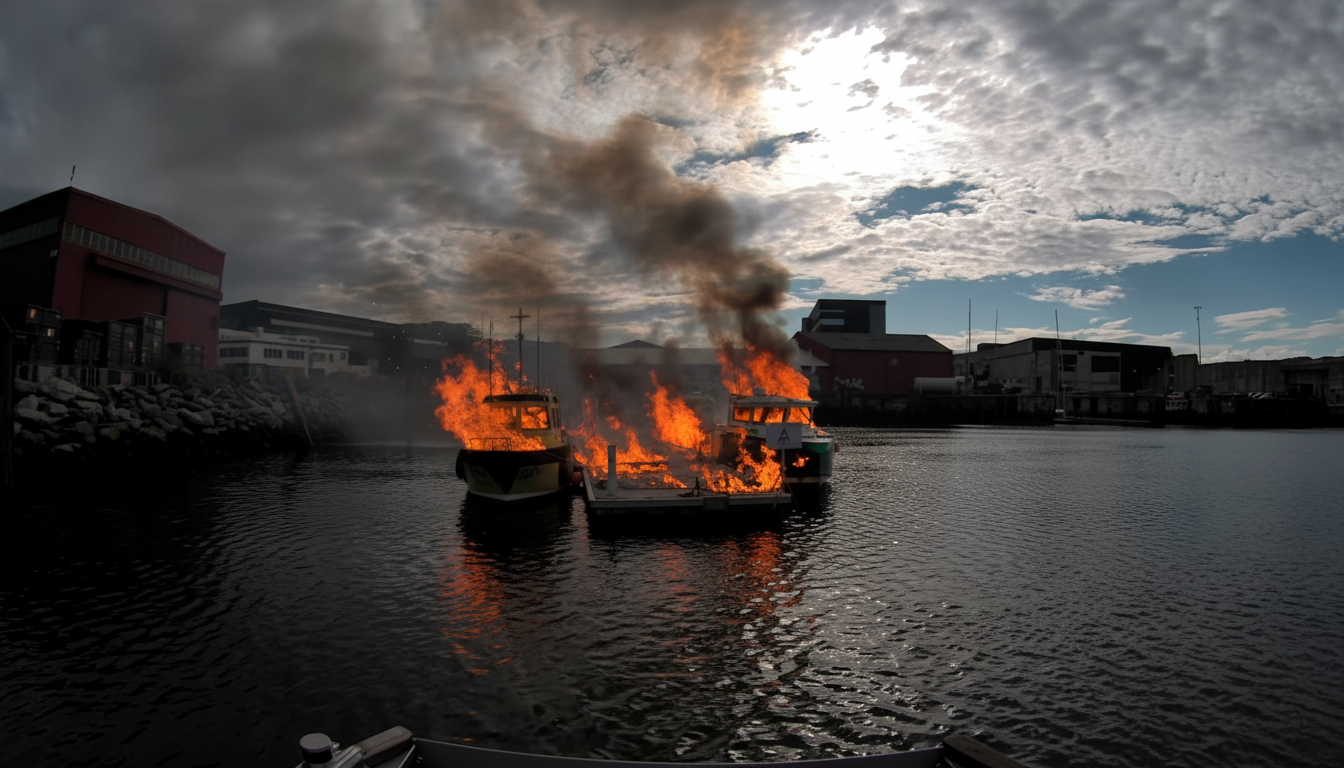}
    \caption{Fire scene.}
  \end{subfigure}\hfill
  \begin{subfigure}{0.49\textwidth}
    \centering
    \includegraphics[width=\linewidth]{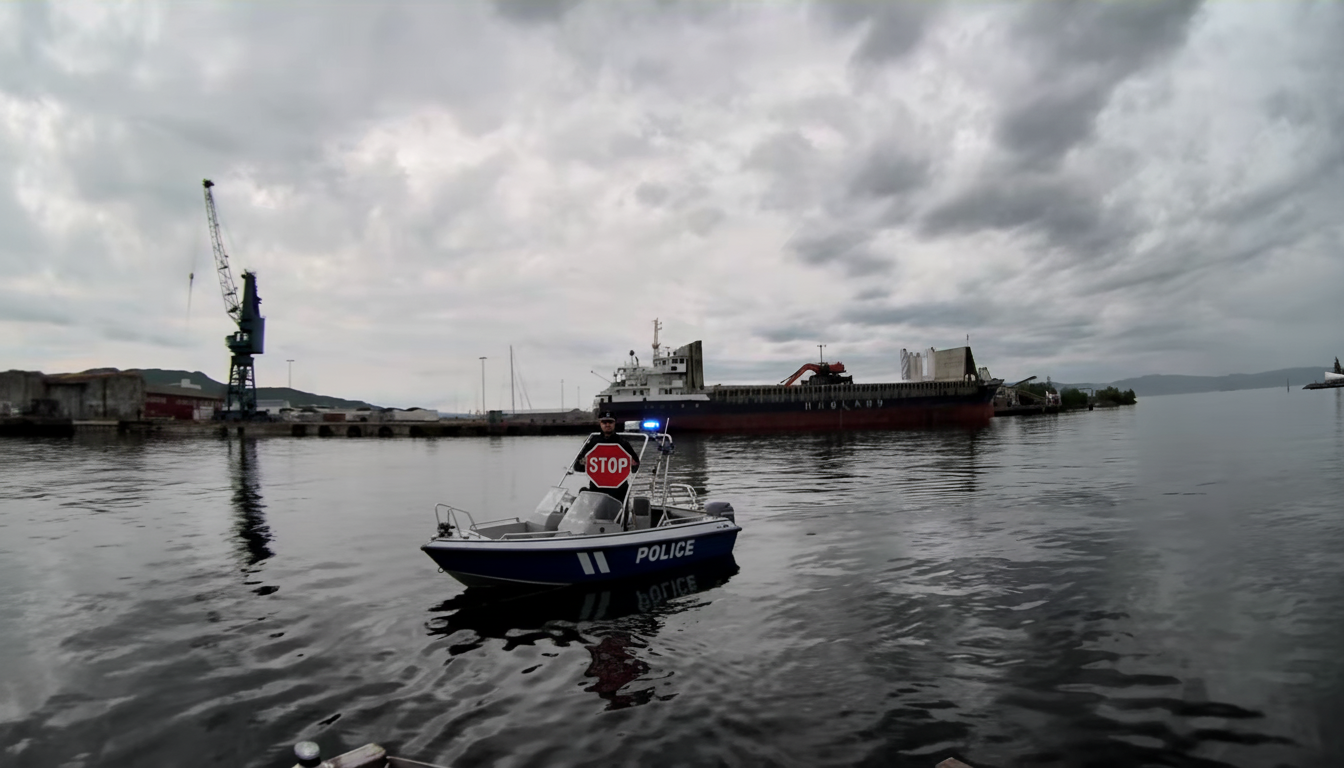}
    \caption{Custom sign.}
  \end{subfigure}

  \caption{Representative scene examples of (a) Diver flag, (b) MOB, (c) fire, (d) custom sign.}
  \label{fig:scene_examples}
\end{figure*}

\paragraph{Offline evaluation dataset}
In Experiments~1–2 (and partially 3), we target four visual/semantic anomaly types that require understanding of scene meaning and are therefore challenging for classical ASV stacks: 
\begin{enumerate}
    \item \textbf{Alpha diver-down flag} (real): The official international diver-down flag requiring reasoning about the implications of unseen divers in the area around it. 
    \item \textbf{Man overboard (MOB)} (real): a person in the water requiring context-dependent actions. 
    \item \textbf{Fire} (AI–enhanced): flames/smoke inserted into frames captured from the vessel to emulate fire hazard nearby.
    \item \textbf{Custom signs} (AI–enhanced): e.g., keep–out or restricted–area signage inserted into vessel–captured frames.
\end{enumerate}
We collect 10 scenes per category. For AI-enhancement, we use \emph{Gemini 2.5 Flash Image} \citep{fortin2025gemini25flashimage} on frames from the real vessel, which are treated identically to real scenes. This enables testing anomalous situations that would have been dangerous or impractical otherwise. Representative real and enhanced example scenes are shown in Figure~\ref{fig:scene_examples}.

\paragraph{Model grid and run protocol}
In Experiments~1 and~2 we evaluate a diverse set of vision–language models and inference settings shown in Table~\ref{tab:eval-models}. Specific models are chosen for Experiments~3 and~4 based on performance in the first two. 

\begin{table}[!htbp]
{
\setlength{\tabcolsep}{3.5pt}
\footnotesize
\centering
\caption{Evaluated VLM models and options.}
\label{tab:eval-models}
\begin{tabular}{l|l|l}
\hline
\textbf{Provider} & \textbf{Model} & \textbf{Variants / options} \\
\hline
Anthropic & \texttt{claude-opus-4-1} & -- \\
Anthropic & \texttt{claude-sonnet-4} & -- \\
Google    & \texttt{gemini-2.5-flash} & -- \\
Google    & \texttt{gemini-2.5-pro}   & reasoning=\texttt{auto} \\
Google    & \texttt{gemini-robotics-er-1.5} & think=\{\texttt{off}, \texttt{medium}, \texttt{full}\} \\
OpenAI    & \texttt{gpt-4.1}          & -- \\
OpenAI    & \texttt{gpt-4o}           & -- \\
OpenAI    & \texttt{gpt-5}            & \{\texttt{high}, \texttt{medium}, \texttt{low}, \texttt{minimal}\} \\
OpenAI    & \texttt{gpt-5-mini}       & \{\texttt{high}, \texttt{medium}, \texttt{low}, \texttt{minimal}\} \\
OpenAI    & \texttt{gpt-5-nano}       & \{\texttt{high}, \texttt{medium}, \texttt{low}, \texttt{minimal}\} \\
\hline
\end{tabular}

}
\end{table}

For each (model, scene) we issue three calls with distinct seeds to assess stability (strict–majority voting as above). We use a conservative prompt that advises Station-keeping when uncertain (alternative prompts, neutral/proactive, are ablated in~\ref{app:prompt_sensitivity}). For every call we record end–to–end latency.

\paragraph{Reporting conventions}
We report 95\% confidence intervals where relevant. For proportions (e.g., Accept@1, Best-set@1) we use the Wilson interval and write $p\,[L,U]$. For continuous outcomes (e.g., awareness score, latency) we report the mean $\pm$ 95\% CI. Pareto scatter plots omit intervals to reduce clutter; the corresponding leaderboards/tables include them.

\subsection{Experiment 1: Scene understanding (H1)}
\label{sec:exp-reasoning}

\paragraph{Overview and method}
Experiment~1 evaluates whether, given the overlaid image and strict output schema, models accurately describe the hazard, its safety implications, and a high-level safe action, and how this trades off with latency. This addresses H1 at the level of textual scene understanding, independent of the specific candidate ID (which is evaluated separately in~\ref{sec:exp-action}).

Each scene is associated with a short human-labeled ground truth that describes the hazard, implications and high-level safe behavior. 
We compare model output to ground truth with an LLM-as-judge configured as GPT-5-low (see~\cite{gu2024survey} for an overview of LLM-as-judge as a method).
The judge is provider–blind and sees only the textual ground truth and model output (not image, candidates etc.) It also tolerates synonyms and penalizes off-topic or unsafe recommendations (see~\ref{app:prompt_llm_judge} for the specific LLM-as-judge prompt).

The judge emits three component scores with fractional credit in $[0,1]$:
\begin{enumerate}
  \item \textbf{Hazard recognition} (1.0 = explicitly correct cue, 0.75 = clearly implied, 0.5 = generic hazard, 0 = wrong/missing).
  \item \textbf{Implication} (does the text state why it matters for safety, e.g., people in water, restricted area, fire risk).
  \item \textbf{Action} (is the proposed high-level maneuver broadly consistent with ground truth, independent of choice ID).
\end{enumerate}

We aggregate to a single awareness score in $[0,1]$ with fixed weights:
\begin{equation}
{
\footnotesize
\text{Awareness} = 0.50\cdot\text{hazard} + 0.25\cdot\text{implication} + 0.25\cdot\text{action}.
}
\end{equation}

Reasoning is evaluated \emph{per call} (FB-1, all three seeds). For each (model, anomaly) and overall, we report (i) the mean awareness score and (ii) the mean end-to-end model latency.

\paragraph{Results and analysis}
\label{sec:result_reasoning}

\begin{figure}[htbp]
  \centering
  \includegraphics[width=\linewidth]{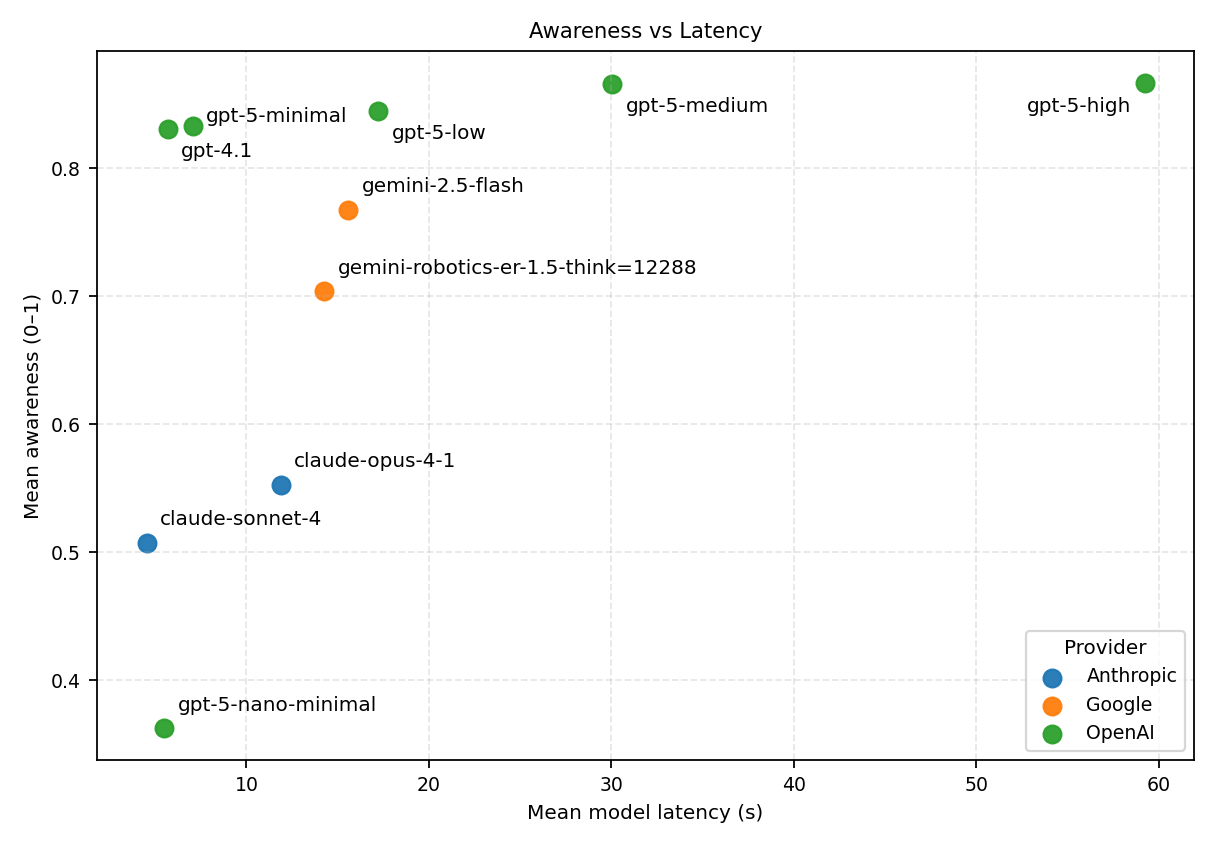}
  \caption{Model awareness vs latency, showing only the pareto frontier.}
  \label{fig:e2a1}
\end{figure}

\begin{figure}[htbp]
  \centering
  \includegraphics[width=\linewidth]{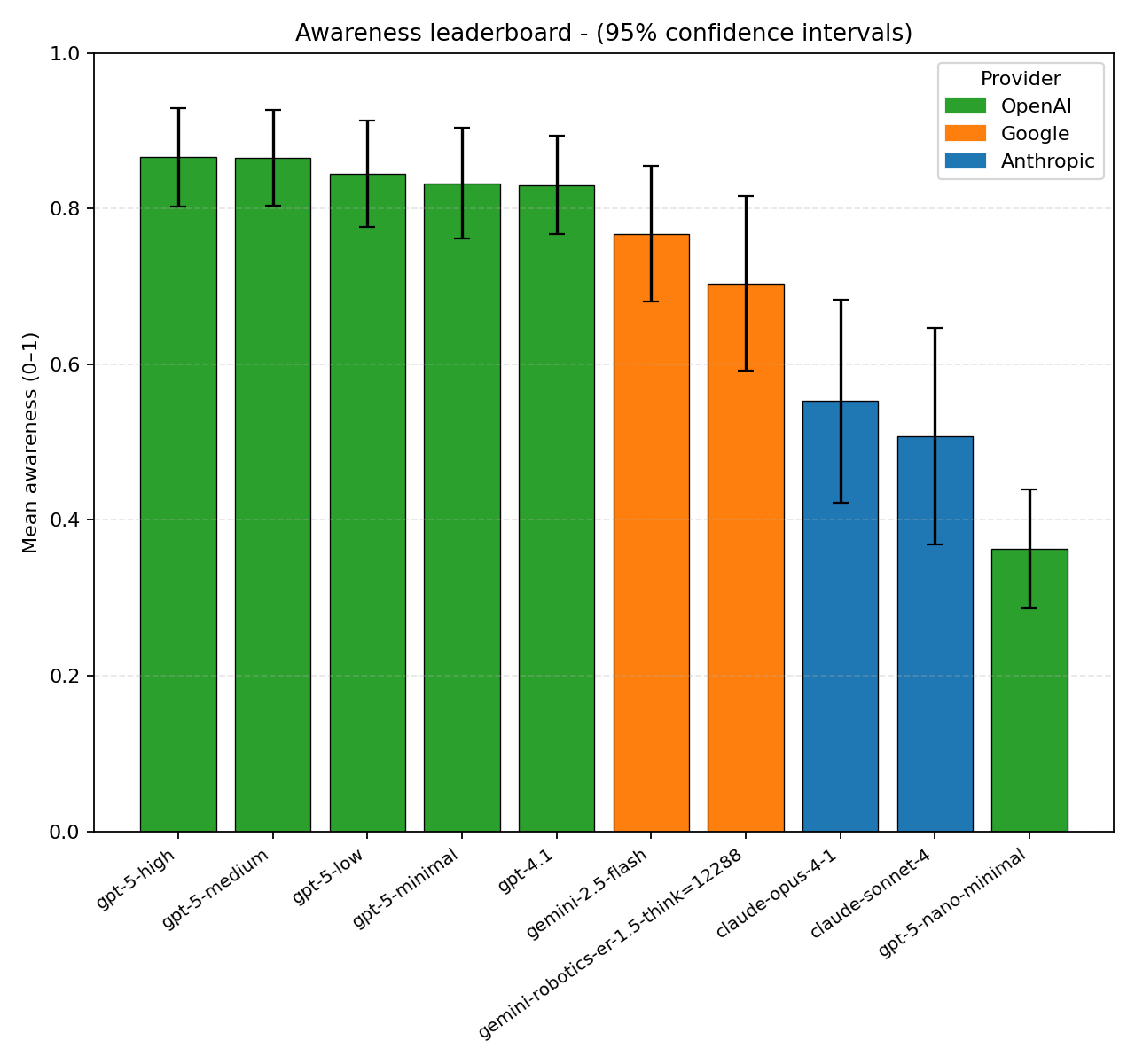}
  \caption{Awareness leaderboard, only showing the per-provider pareto frontier}
  \label{fig:e2a2}
\end{figure}

We evaluate per-call scene understanding (hazard, implication, action) using the LLM-as-judge protocol above and report mean awareness and end-to-end latency across the 40 scenes (10 per anomaly). Figures~\ref{fig:e2a1} and~\ref{fig:e2a2} summarize the awareness-latency Pareto frontier by provider. Error bars in Figure~\ref{fig:e2a2} show 95\% confidence interval (CI). 

OpenAI models show overall best performance: gpt-5-high shows the highest mean awareness ($0.866\pm0.063$) at $59.2\pm4.0$\,s, while strong sub-10\,s alternatives, gpt-4.1 ($0.830\pm0.063$ at $5.69\pm0.22$\,s) and gpt-5-minimal ($0.833\pm0.071$ at $7.07\pm0.47$\,s), provide most of the performance at much lower latency (Table~\ref{tab:scene-top5}). Incremental gains beyond $\sim$30\,s are within CI (like gpt-5-medium at $0.866\pm0.062$ at $30.0\pm1.9$\,s vs. gpt-5-high).

The best non-OpenAI frontier points are gemini-2.5-flash ($0.768\pm0.087$ at $15.6\pm1.1$\,s) and gemini-robotics-er-1.5 think=12288 ($0.704\pm0.112$ at $14.3\pm0.6$\,s), while Anthropic's fastest frontier point is claude-sonnet-4 ($0.507\pm0.139$ at $4.53\pm0.11$\,s). The best models mostly span from $\sim$5--7\,s at $\sim$0.83 awareness to $\sim$60\,s at $\sim$0.87 awareness score.

Across the Top-5 frontier entries, hazard, implication and action scores are generally high and balanced (Hazard: $\sim$0.83--0.87, Implication: $\sim$0.81--0.85, Action: $\sim$0.84--0.88) (Table~\ref{tab:scene-top5}). \ref{app:scene_understanding_all} shows the results for all models tested. 

\begin{table}[htbp]
\footnotesize
\centering
\caption{Top 5 Pareto-frontier models for awareness-latency trade-off (means $\pm$95\% CI over $n{=}40$ scenes). H (hazard)/ I (implication) / A (action) lists component means.}
\label{tab:scene-top5}
\begin{tabular}{l|r|r|l}
\hline
\textbf{Model} & \textbf{Latency [s]} & \textbf{Awareness} & \textbf{H/I/A} \\
\hline
gpt-4.1 & $5.69 \pm 0.22$ & $0.830 \pm 0.063$ & 0.83/0.83/0.84 \\
gpt-5-minimal & $7.07 \pm 0.47$ & $0.833 \pm 0.071$ & 0.82/0.82/0.86 \\
gpt-5-low & $17.21 \pm 1.11$ & $0.845 \pm 0.068$ & 0.85/0.81/0.86 \\
gpt-5-medium & $30.04 \pm 1.94$ & $0.866 \pm 0.062$ & 0.87/0.85/0.88 \\
gpt-5-high & $59.22 \pm 4.01$ & $0.866 \pm 0.063$ & 0.87/0.85/0.87 \\
\hline
\end{tabular}
\end{table}

\paragraph{Failure-case analysis and anomaly-specific observations}
Per-anomaly results (\ref{app:scene_understanding_fire}--\ref{app:scene_understanding_sign}) show the largest inter-model spread on \emph{Diver flag} and \emph{MOB}, with most models near a high plateau on \emph{Fire} and \emph{Custom sign}. When defining ``low awareness'' as the bottom quartile of per-call awareness scores, failures concentrate overwhelmingly in the small/low-salience categories: 63.7\% of \emph{Diver flag} calls and 30.4\% of \emph{MOB} calls fall into this low-awareness subset, vs. 11.7\% for \emph{Signs} and 1.4\% for \emph{Fire}. In the low-awareness subset, mean rubric subscores are low for scene understanding and implications (\texttt{see}/hazard: 0.16, implication: 0.18), while action is higher (0.33), indicating that the dominant root cause is missed or misinterpreted hazards rather than fine-grained action disagreement. Notably, even among low-awareness outputs, models sometimes propose plausible generic maritime actions (e.g., ``slow down'' / ``keep clear'') without correctly identifying the cue, which can make the text appear reasonable while being semantically ungrounded. Typical recurring modes are: missed anomaly (no grounding), misidentification as a benign buoy/marker/clutter, a generic COLREG-like caution template, and implication gaps (e.g., noticing a flag/shape but not inferring divers or rescue constraints).

\paragraph{Takeaways}
These results support hypothesis H1: modern VLMs can achieve high maritime scene awareness at practical latencies, with sub-10\,s models retaining most of the awareness of much slower systems. When the  selector correctly recognizes the hazard, the resulting rationale is specific enough to support explainable handovers in the ROC GUI, while remaining failure modes are dominated by subtle, low-salience cues such as the alpha diver flag and small, distant MOB targets.

\subsection{Experiment 2: Action alignment with human choices (H2)}
\label{sec:exp-action}

\paragraph{Overview and method}
Experiment~2 evaluates whether the picked trajectory ID under FB-3 matches what humans consider reasonable actions on the same overlays. Because these semantic anomalies involve value-laden tradeoffs and partial observability from only a single camera view, there is no unique ground-truth ``correct'' action label. Instead we treat aggregated human judgments as a proxy for reasonable behavior in cases where geometry alone is insufficient.

Human raters mark any number of \emph{Acceptable}
candidates and select a single \emph{Best} candidate. The Station-keeping option (ID 0) is available and
treated like any other candidate. Each scene is rated by $N\!\ge\!11$ raters. For rater $r$, let $A^{(r)}\subseteq\{0,\dots,K\}$ be the
Acceptable set and $b^{(r)}\in\{0,\dots,K\}$ the Best pick. “None acceptable’’ is treated as abstention and set to Station-keep.

To form the \emph{Acceptable} consensus we define the acceptance frequency
\begin{equation}
\hat p_k \;=\; \frac{1}{N}\sum_{r=1}^N \mathbf{1}\!\left[k \in A^{(r)}\right], 
\qquad k\in\{0,\dots,K\},
\label{eq:accept_freq}
\end{equation}
and set
\begin{equation}
\mathrm{ACCEPT} \;=\; \bigl\{k:\ \hat p_k \ge \tau_{\text{acc}}\bigr\},
\qquad \tau_{\text{acc}}=0.6.
\label{eq:accept_set}
\end{equation}

To allow for scenes with more than one defensible “best” action (e.g., stop near a MOB vs.\ steer away), we permit up to three \emph{Best} items when support is split across strong contenders.

We first count “best” votes per candidate:
\begin{equation}
v_k \;=\; \sum_{r=1}^N \mathbf{1}\!\bigl[b^{(r)} = k\bigr],
\qquad k\in\{0,\dots,K\}.
\label{eq:best_votes}
\end{equation}
Let the top vote count be
\begin{equation}
v_{\max} \;=\; \max_{k} v_k.
\label{eq:vmax}
\end{equation}
We require each \emph{Best} item to clear a threshold that encodes two simple safeguards:
(i) it must be reasonably competitive with the top option (at least half as many votes), and
(ii) it must have a non-trivial minimum of raters (at least one quarter of \(N\)):
\begin{equation}
\theta \;=\; \max \left(\bigl\lceil 0.5\,v_{\max}\bigr\rceil,\ \bigl\lceil 0.25\,N\bigr\rceil\right).
\label{eq:best_theta}
\end{equation}
Candidates meeting this support are retained,
\begin{equation}
\mathrm{BEST}^{\star} \;=\; \{\,k:\ v_k \ge \theta\,\},
\qquad
\mathrm{BEST} \;=\; \mathrm{BEST}^{\star} \cap \mathrm{ACCEPT}.
\label{eq:best_sets}
\end{equation}

If \(|\mathrm{BEST}|>3\), we keep the three items with largest \(v_k\) (ties broken by larger \(\hat p_k\), then smaller ID). This results in one to three “best” choices when raters split between a small number of strong alternatives, while still filtering outlier picks. Qualitative inspection of per-scene \(\mathrm{ACCEPT}\) and \(\mathrm{BEST}\) sets shows reasonable results (see~\ref{app:consensus_reasonable} for examples).

\paragraph{Metrics and baselines}
Each (model, scene) is run three times with distinct seeds; unless otherwise stated we report majority-of-three (FB-3). We measure:

\begin{itemize}
    \item \textbf{Accept@1:} fraction of scenes where the chosen ID lies in the human consensus \(\mathrm{ACCEPT}\) set.
    \item \textbf{Best@1:} fraction of scenes where the chosen ID lies in the human consensus \(\mathrm{BEST}\) set.
\end{itemize}

These metrics explicitly tie the model’s action to what humans regard as acceptable or preferred among the pre-vetted candidates. Purely geometric metrics like distance to shore or relative bearing would not capture semantics such as divers, signage, or people in the water. By aggregating raters, we approximate the notion of “reasonable behavior” in these ambiguous, meaning-dependent scenes.

We compare the  selector method to geometry-only baselines, which operate on the identical, gated set: \textbf{Station-keep} (always pick ID 0); \textbf{Keep-course} (pick the candidate with the smallest bearing relative to vessel heading, $|\phi_k|$); \textbf{Keep-starboard} (pick the most starboard gated trajectory); \textbf{Forward} (maximize forward displacement ($x$) of the endpoint); \textbf{Clearance} (maximize the minimum pixel-space clearance along the candidate’s projected samples). 

\paragraph{Results and analysis}
\label{sec:results_alignment}

We evaluate whether the picked candidate aligns with the human consensus \emph{Accept} and \emph{Best} sets on the same overlay, using FB-3. Geometry-only baselines operate on the same, gated set of candidates (\emph{Keep-station}, \emph{Keep-starboard}, \emph{Keep-course}, \emph{Forward}, \emph{Clearance}) as defined above.

\begin{figure}[htbp]
  \centering
  \includegraphics[width=\linewidth]{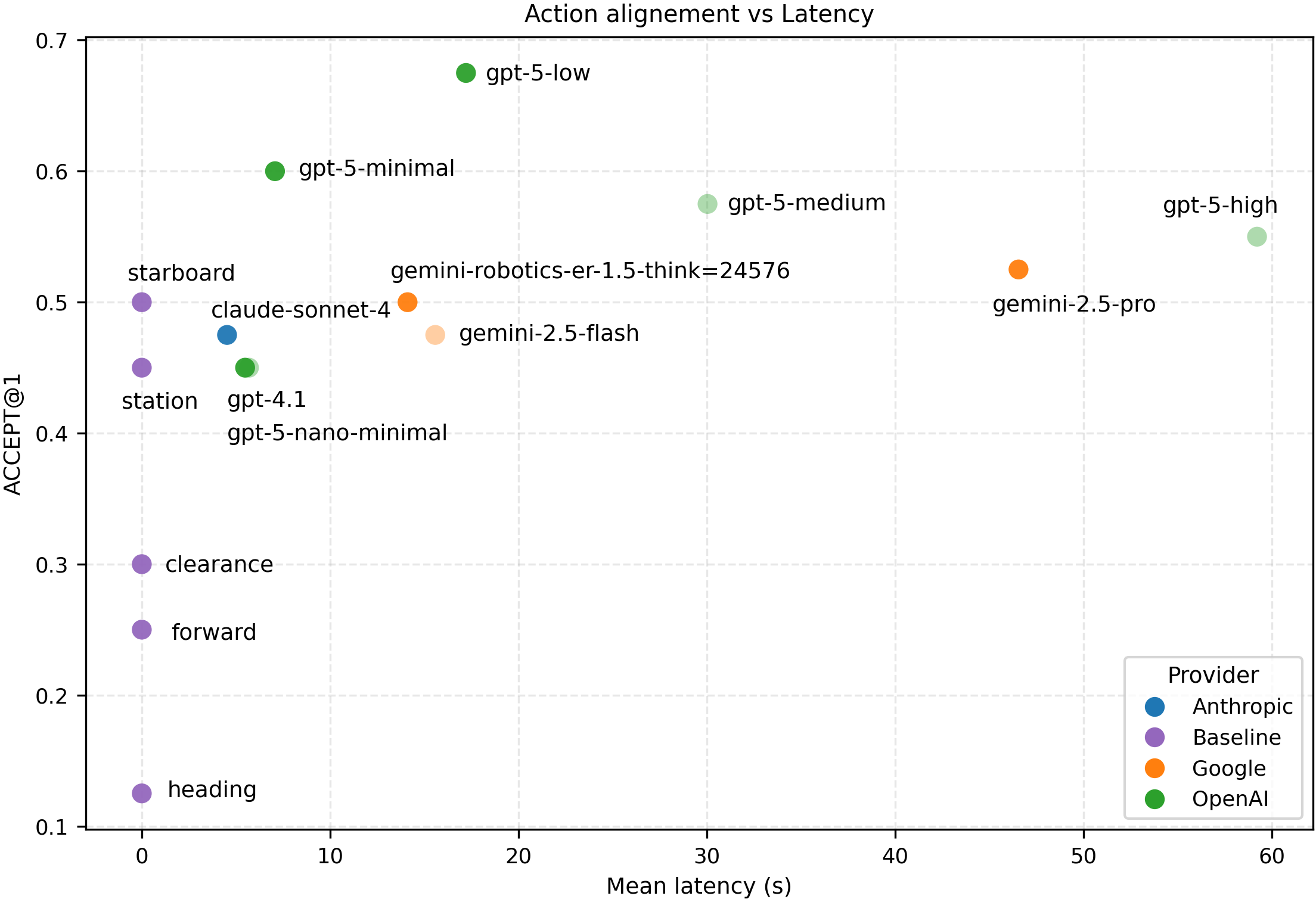}
  \caption{Accept rate vs latency, showing only the per provider Pareto frontier and shaded carryovers from Experiment~1.}
  \label{fig:e2_pareto_accept}
\end{figure}

\begin{figure}[htbp]
  \centering
  \includegraphics[width=\linewidth]{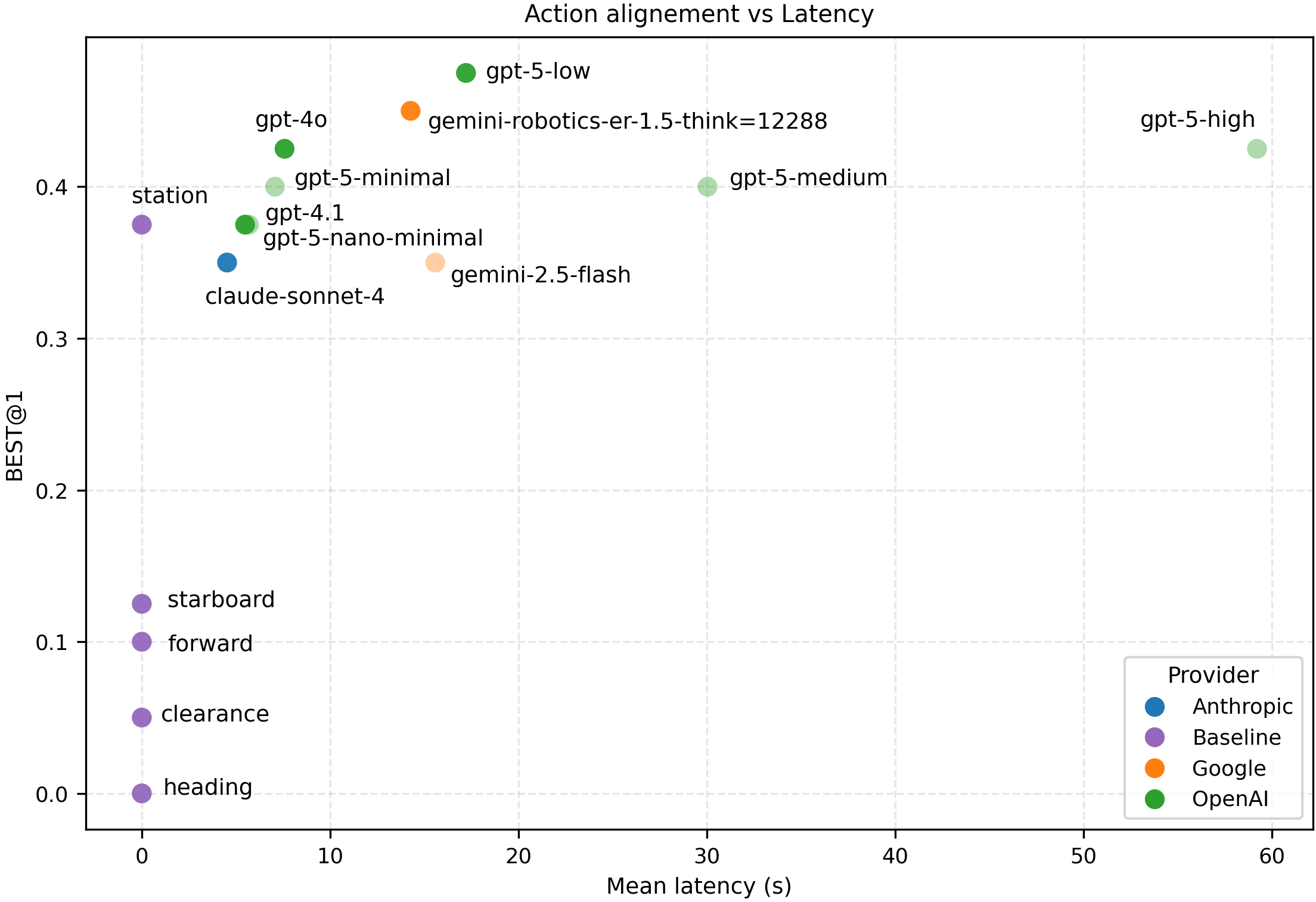}
  \caption{Best rate vs latency, showing only the per provider Pareto frontier and shaded carryovers from Experiment~1.}
  \label{fig:e2_pareto_best}
\end{figure}

\begin{figure}[htbp]
  \centering
  \includegraphics[width=\linewidth]{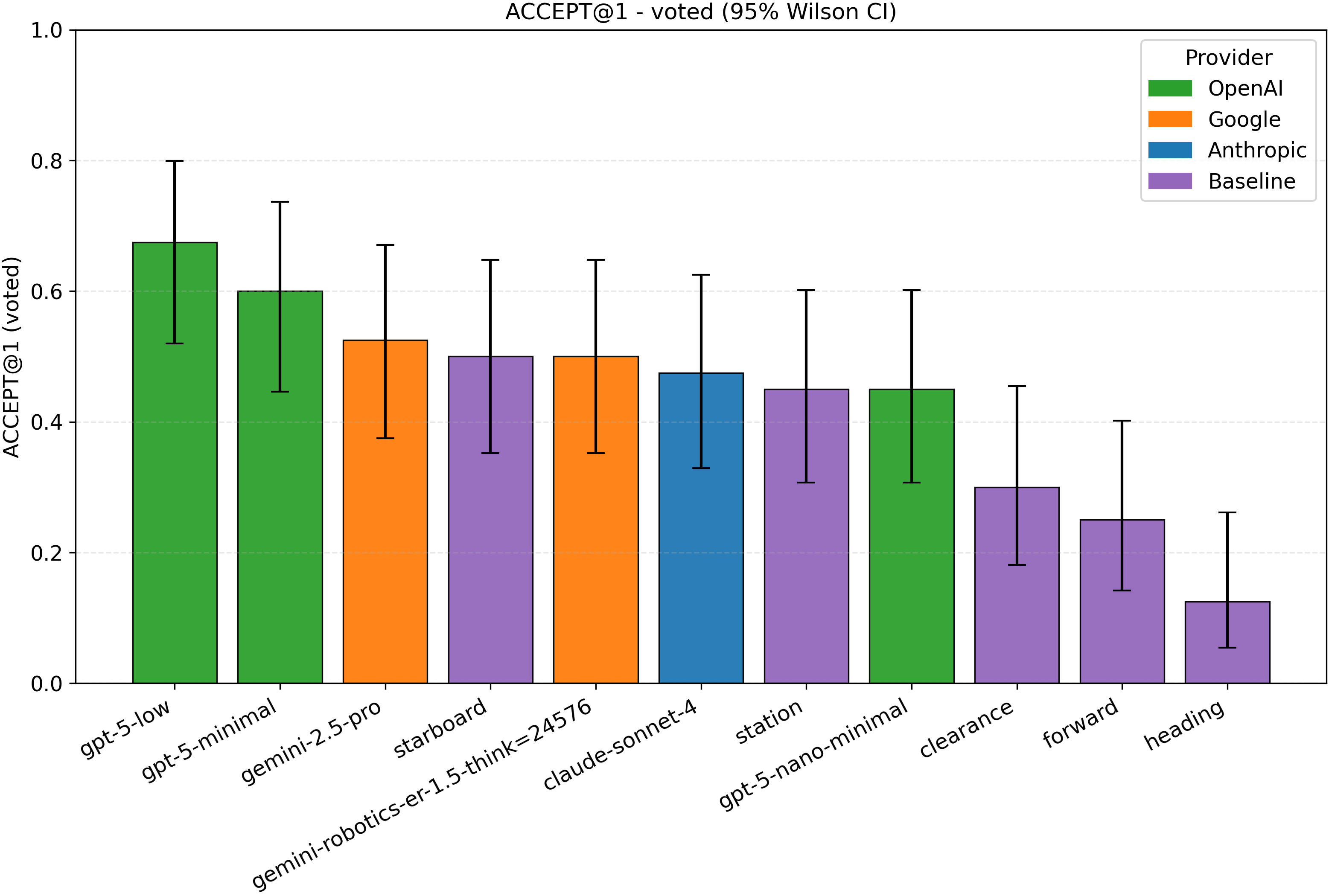}
  \caption{Accept@1 rate leader board with Wilson CI. Only the Pareto frontier models and all baseline models shown.}
  \label{fig:e2b1}
\end{figure}

\begin{figure}[htbp]
  \centering
  \includegraphics[width=\linewidth]{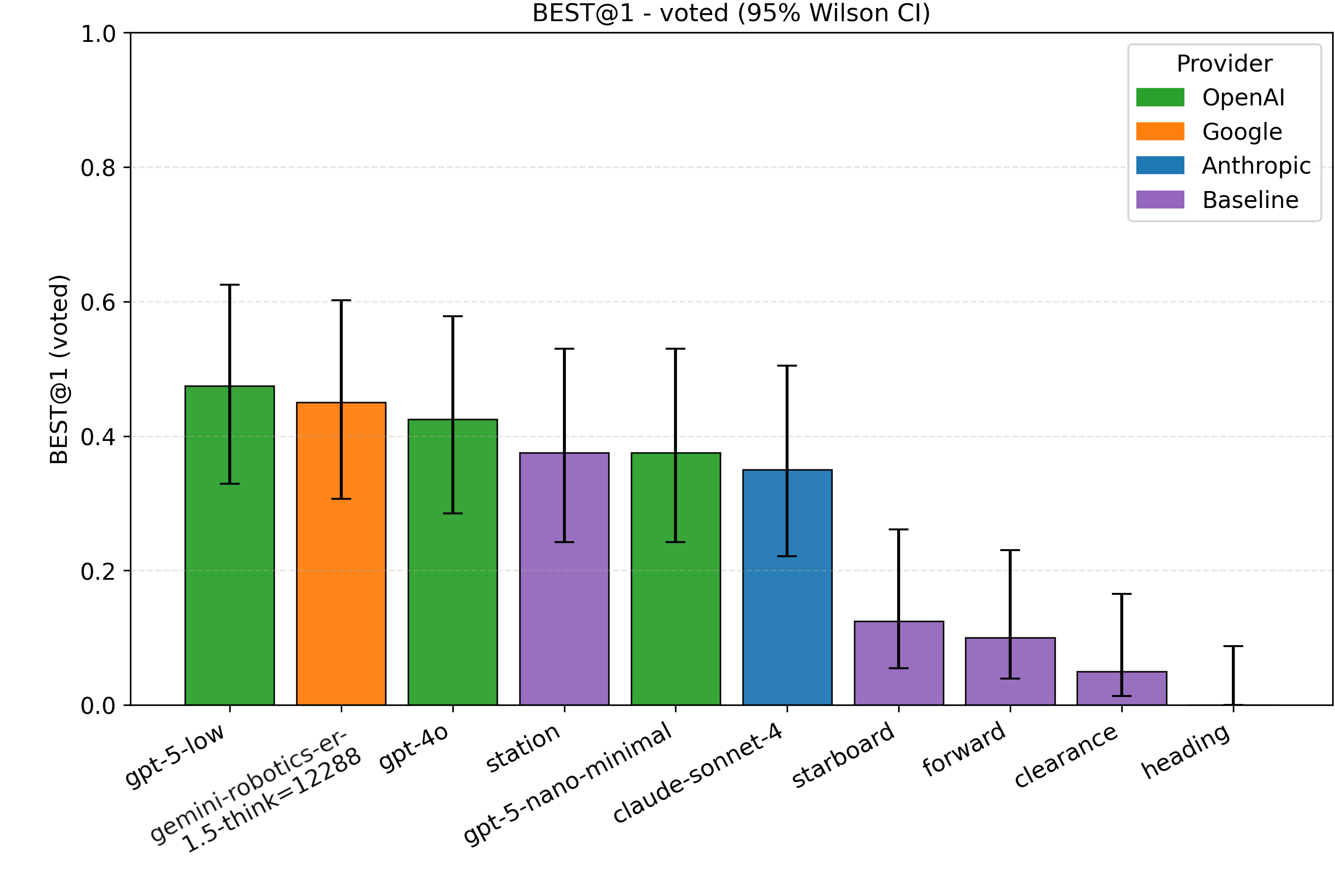}
  \caption{Best@1 rate leader board with Wilson CI. Only the Pareto frontier models and all baseline models shown.}
  \label{fig:e2b2}
\end{figure}

Figures~\ref{fig:e2_pareto_accept} and \ref{fig:e2_pareto_best} show per-provider Pareto frontiers for action alignment versus mean latency, all baselines are included and are plotted at zero latency. Models that were on the frontier in Fig.~\ref{fig:e2a1} (scene understanding) but are not on the current frontier appear as transparent carryovers for reference. The leader boards in Figures~\ref{fig:e2b1} and~\ref{fig:e2b2} similarly show frontier models (but not carryover) and baselines, with 95\% Wilson CIs.

Across the full set of scenes, the best overall model on both metrics is gpt-5-low, which we therefore use for FB-3 in the risk–relief analysis in Experiment~3. Its FB-3 results are Accept@1 \(=0.68\) \([0.52,\,0.80]\) and Best@1 \(=0.48\) \([0.33,\,0.63]\) at a latency of \(16.1\,\mathrm{s}\) (Table~\ref{tab:alignment_overall}). In the sub-10\,s region, gpt-5-minimal achieves Accept@1 \(=0.60\) \([0.45,\,0.74]\) and Best@1 \(=0.40\) \([0.26,\,0.55]\) at \(6.7\,\mathrm{s}\), while gpt-4o yields Accept@1 \(=0.55\) \([0.40,\,0.69]\) and Best@1 \(=0.43\) \([0.29,\,0.58]\) at \(5.8\,\mathrm{s}\). These provide strong and fast alternatives with a modest drop from the overall best. Like in Experiment~1, OpenAI models dominate the frontier. 

Because Station (ID~0) appears frequently in the human consensus, baseline \emph{Keep-station} is relatively competitive: it obtains Accept@1 \(=0.45\) \([0.31,\,0.60]\) and Best@1 \(=0.38\) \([0.24,\,0.53]\) (Table~\ref{tab:alignment_overall}). This also highlights the general dataset composition: Station is in the rater \emph{Accept} set for \(45\%\) of scenes and in the \emph{Best} set for \(37.5\%\) of scenes. In contrast, \emph{Keep-starboard} is often acceptable but rarely best (\textsc{Accept@1} \(=0.50\) \([0.35,\,0.65]\), \textsc{Best@1} \(=0.13\) \([0.05,\,0.26]\)), consistent with open-water frames and pre-vetted trajectories that already avoid obvious non-water regions. All model and baseline results are in~\ref{app:action_all}.

\paragraph{Failure-case analysis and anomaly-specific observations}
Per-anomaly tables (Fire, Diver flag, MOB, Custom sign) are provided in~\ref{app:action_fire} --~\ref{app:action_sign}. As in Experiment~1, action alignment spreads are largest on the low-salience categories (\emph{Diver flag}, \emph{MOB}) and smaller on \emph{Fire}/\emph{Sign}. Typical misalignment modes mirror the scene-understanding failures: missing or misinterpreting the cue (e.g., diver-down marker or distant MOB confused with buoy/clutter) can lead to selecting a candidate that passes too close to the hazard region. A second reason is on the conversion from understanding to concrete action: even when the hazard is correctly described at a high level, mapping that intent to a single numbered best candidate among many options is stricter and requires better grounding than producing a plausible textual recommendation. This is reflected in a consistent gap between high-level scene understanding and concrete action selection: the best mean awareness in Experiment~1 reaches $0.866$, whereas the best overall action alignment here peaks at Accept@1 $=0.68$ and Best@1 $=0.48$ (Table~\ref{tab:alignment_overall}), indicating that selecting an acceptable/best \emph{specific} short-horizon path is considerably harder for the models than describing the situation and a generic safe response.

\paragraph{Takeaways}
The experiment supports hypothesis H2: within our pre-vetted candidate set, the FB-3  selector aligns with human Accept/Best judgments more often than geometry-only baselines, even though Station-keeping and keep-starboard remain competitive in many scenes. This suggests that semantics—not only simple geometric rules—matter in the relatively small set of cases where those defaults would be unsafe, and that action alignment with human consensus is a useful proxy for ``reasonable'' behavior in such semantic anomalies. The observed gap between awareness and action alignment suggests that translating high scene understanding into concrete safe actions is a key remaining bottleneck.

\begin{table}[!htbp]
\centering
\footnotesize
\caption{Overall action alignment (FB-3 majority-of-three) with 95\% Wilson CIs over \(N{=}40\) scenes; median latency across calls.}
\label{tab:alignment_overall}
\begin{tabular}{l|c|c|r}
\hline
\textbf{Method} & \textbf{Accept@1} & \textbf{Best@1} & \textbf{Latency (s)} \\
\hline
FB-3 (gpt-5-low) & $0.68 \;[0.52,\,0.80]$ & $0.48 \;[0.33,\,0.63]$ & 16.11 \\
Keep-station     & $0.45 \;[0.31,\,0.60]$ & $0.38 \;[0.24,\,0.53]$ & $\sim$0.00 \\
Keep-starboard   & $0.50 \;[0.35,\,0.65]$ & $0.13 \;[0.05,\,0.26]$ & $\sim$0.00 \\
\hline
\end{tabular}
\end{table}

\subsection{Experiment 3: Fire-only risk–relief (H3)}
\label{sec:exp-riskrelief}

\paragraph{Overview and method}
In the event of an anomaly detection, the time between alert and human override could be considerable. Experiment~3 asks whether the  selector can reason about an unambiguously dangerous hazard and move the vessel toward safety, addressing H3 on directional risk relief. We focus on the fire subset, where the hazard location is well defined.

We evaluate the winning  selector, \textbf{FB-3} (gpt-5-low), on short-horizon directional risk relief and compare it to baselines (\textbf{Keep-station}, \textbf{Keep-course} and \textbf{Keep-starboard}) on the unambiguously dangerous \emph{fire} subset of the dataset. For each scene we annotate a single hazard point $h$ on the water plane, back-projected with the known camera intrinsics and extrinsics from \eqref{eq:projection}, Section~\ref{sec:methods-cands}. $h$ is only used for evaluation. We start from the bow anchor $x_0=(4.0,0.0)$\,m in the boat frame and execute the chosen trajectory at a constant anomaly speed $U_{\mathrm{anom}}=0.514$\,m/s (1\,kn) without replanning. If the endpoint is reached early, the vessel stops. This is simulated only and assumes the hazard is stationary. 

For a horizon $H(s)$, let $x_H(p)$ be the position after straight–line motion toward the chosen endpoint (or $x_H=x_0$ for Keep-station). We measure change in separation
\begin{equation}
\Delta d_H(p)\;=\;\bigl\|x_H(p)-h\bigr\|_2\;-\;\bigl\|x_0-h\bigr\|_2,
\end{equation}
so that positive values indicate increased distance from the hazard (risk relief), negative values decreases distance from the hazard, and Keep-station is always $0$. Because fire is unambiguously dangerous and localized on the water plane, this geometric standoff metric directly captures whether a choice moves away from or toward the hazard and is easier to interpret than in the more ambiguous anomaly types where multiple actions could be reasonable for humans.

\paragraph{Results and analysis}
\label{sec:rr_fire}

\begin{figure}[!htbp]
  \centering
  \includegraphics[width=\linewidth]{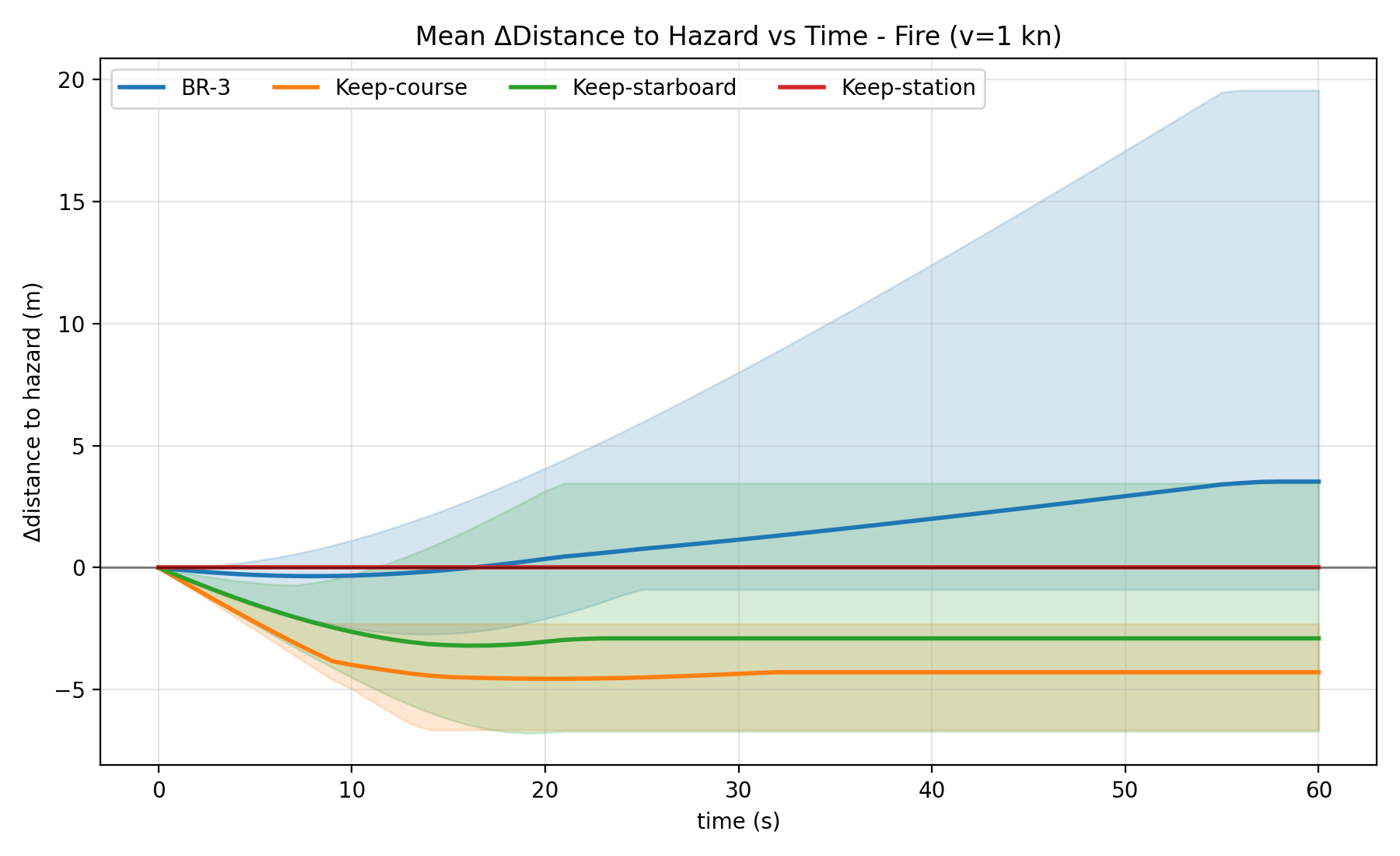}
  \caption{Risk-relief on the fire scenes: mean $\Delta d_H$ (m) over time with min--max shading across $n{=}10$ scenes. Positive values indicate increased separation from the annotated hazard.}
  \label{fig:rrvs}
\end{figure}

We report \emph{mean} $\Delta d_H$ across scenes as a function of time and at fixed
horizons $H\in\{10,30,60\}$\,s, and also show min-max values to visualize best and worst runs (Table~\ref{tab:delta-means} and Fig.~\ref{fig:rrvs}). FB-3 yields large positive relief on some scenes while remaining neutral on others, depending on the situation. It outcompetes all baselines, which show mean zero (station) or negative (the rest). 

\paragraph{Takeaways}
These findings support H3: on unambiguously dangerous fire scenes, the FB-3 method tends to increase standoff distance compared to keep-course, keep-starboard, or Station-keeping, providing directional evidence that the human-alignment result from Experiment~2 translates into short-horizon risk relief when the hazard is spatially localized and semantically unambiguous.

{
\setlength{\tabcolsep}{5pt}
\begin{table}[!htbp]
{\footnotesize
\centering
\caption{Fire scenes ($n=10$). Mean change in separation $\Delta d_H$ [m] at $H\in\{10,30,60\}$\,s. Brackets show per-scene range [min, max]. Positive indicates increased distance from the hazard.}
\label{tab:delta-means}
\begin{tabular}{l|r|r|r}
\hline
\textbf{Fallback} & \textbf{$\Delta d_{10\text{s}}$} & \textbf{$\Delta d_{30\text{s}}$} & \textbf{$\Delta d_{60\text{s}}$} \\
\hline
FB-3           & $-0.3$ [$-2.5$, $1.1$]  & $1.1$  [$-0.9$, $8.0$]   & $3.5$  [$-0.9$, $19.5$] \\
K-course    & $-4.0$ [$-5.0$, $-2.3$] & $-4.3$ [$-6.6$, $-2.3$]  & $-4.3$ [$-6.6$, $-2.3$] \\
K-starboard & $-2.6$ [$-4.5$, $-0.3$] & $-2.9$ [$-6.7$, $3.5$]   & $-2.9$ [$-6.7$, $3.5$]  \\
K-station   & $0.0$  [$0.0$, $0.0$]    & $0.0$  [$0.0$, $0.0$]     & $0.0$  [$0.0$, $0.0$]    \\
\hline
\end{tabular}
}
\end{table}
}

\subsection{Experiment 4: Integrated sea trial (H4)}
\label{sec:exp-field}

\paragraph{Overview and method}
To validate the  selector in a real, closed-loop system, we ran the end-to-end stack on a real ASV with control and override capabilities as described in Sec.~\ref{sec:methods-control}. This addresses H4 by testing whether the full alert$\rightarrow$fallback maneuver$\rightarrow$override chain functions under realistic constraints.

We executed \(n=5\) diving flag scenario runs from the ROC with real human operators who could accept the alert and take manual control. The  selector used FB-1 (gpt-5-medium), selected offline because of its high performance on diver-flag scene understanding cases. For each run:

\begin{enumerate}
    \item An anomaly alert was raised (manually for this proof-of-concept).
    \item The  selector produced a single action ID and textual rationale, which were shown to the operator together with a map view.
    \item The world-fixed path for the chosen candidate was sent to the LOS controller and executed at anomaly-mode speed, with joystick override always available.
    \item The operator took control from the ROC and performed a short manual approach and safe stop at the dock.
\end{enumerate}

The same $n=5$ sessions are later used for a separate formative handover HMI study (Sec.~\ref{sec:formative_handover}), which analyzes operator experience qualitatively.

\paragraph{Results and analysis}
\label{sec:seatrial_results}

\begin{figure}[!t]
  \centering

  \begin{subfigure}{\linewidth}
    \centering
    \includegraphics[width=\linewidth]{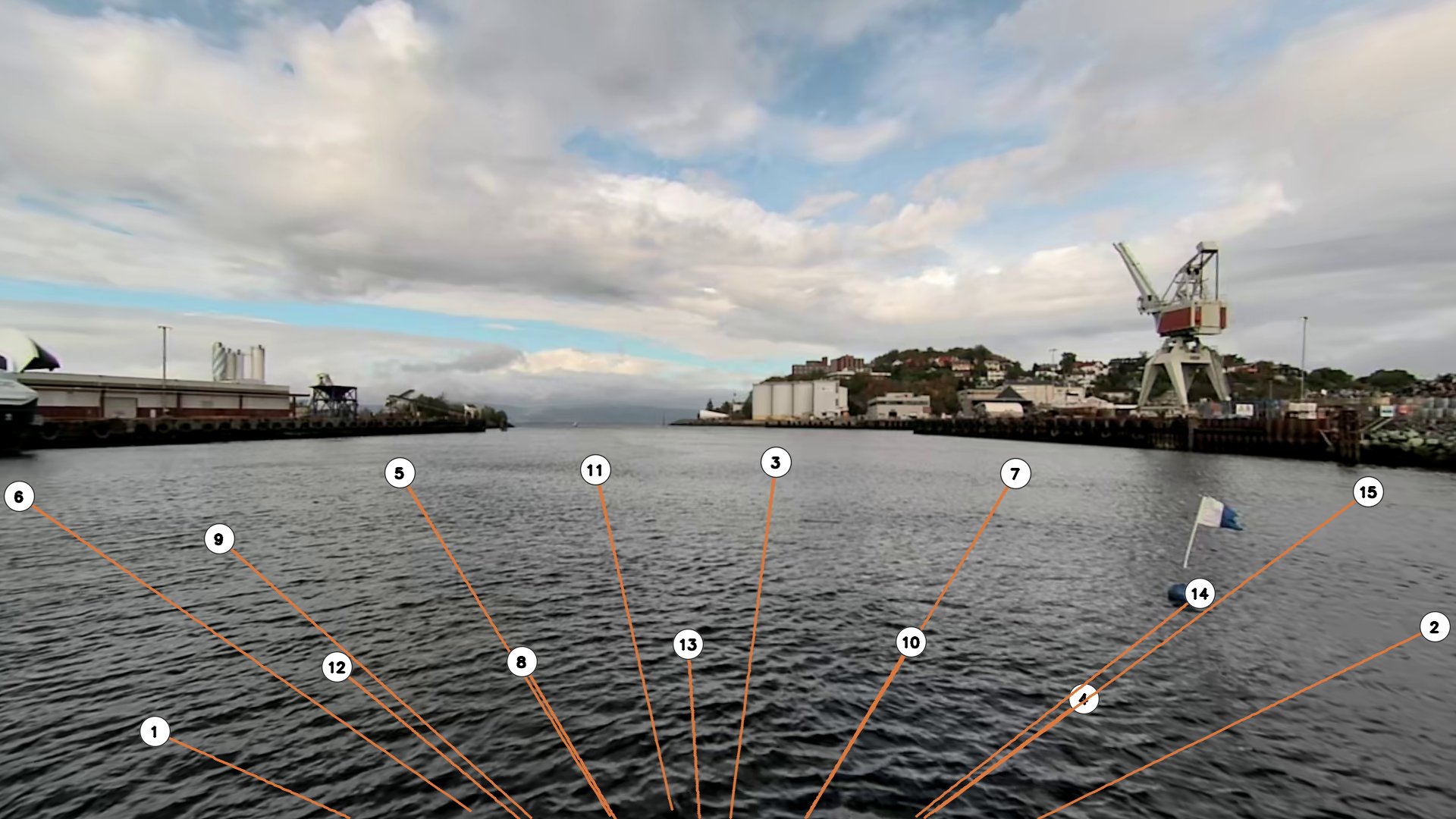}
    \caption{Trajectory options overlaid on the image frame used by FB-1.}
    \label{fig:live_example_a}
  \end{subfigure}

  \par\medskip

  \begin{subfigure}{\linewidth}
    \centering
    \includegraphics[width=\linewidth]{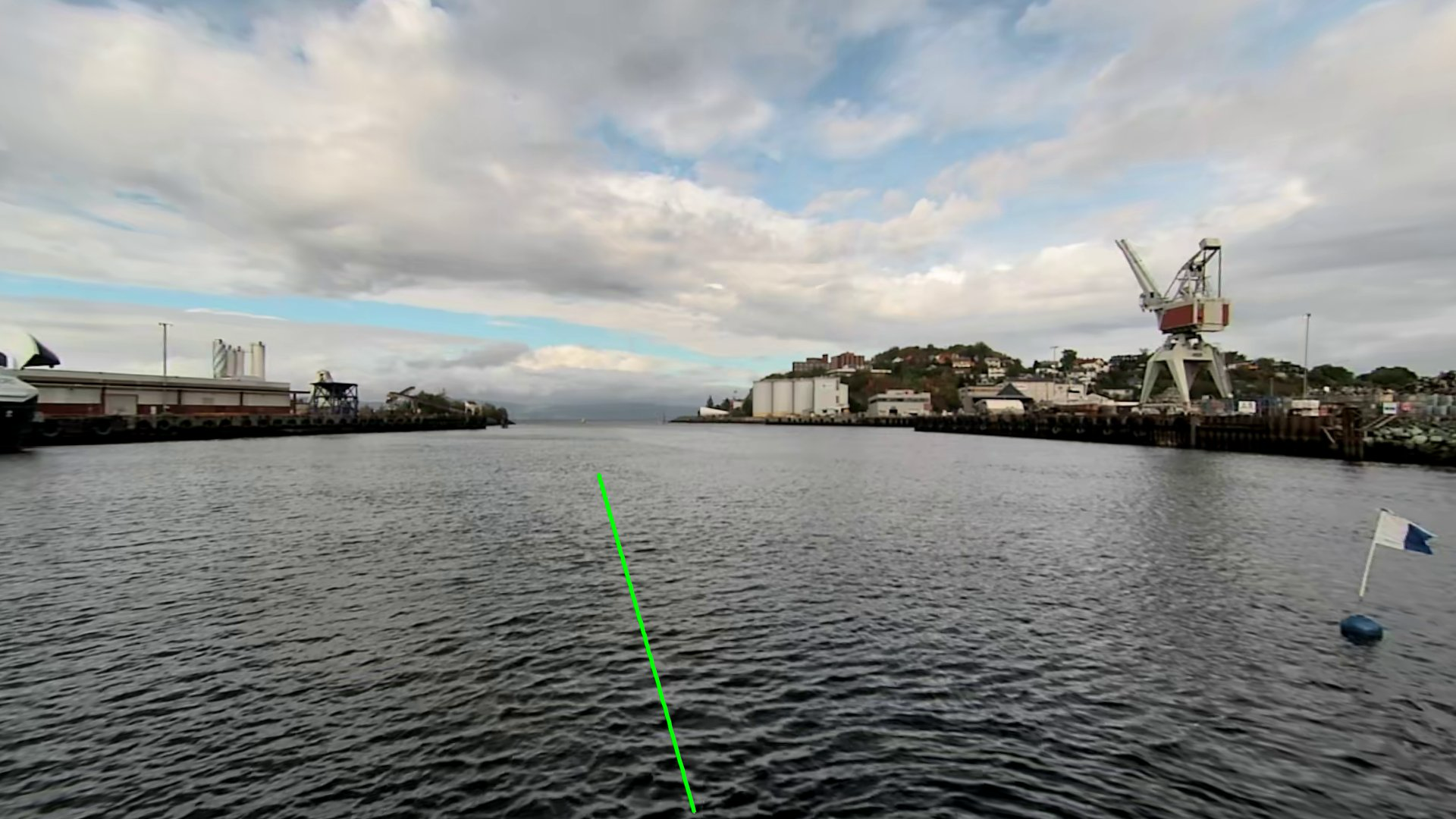}
    \caption{Chosen trajectory overlay (11).}
    \label{fig:live_example_b}
  \end{subfigure}

  \caption{Live example of trajectory options and the chosen trajectory (11).}
  \label{fig:live_example}
\end{figure}

\begin{figure}[!htbp]
  \centering
  \includegraphics[width=0.97\linewidth]{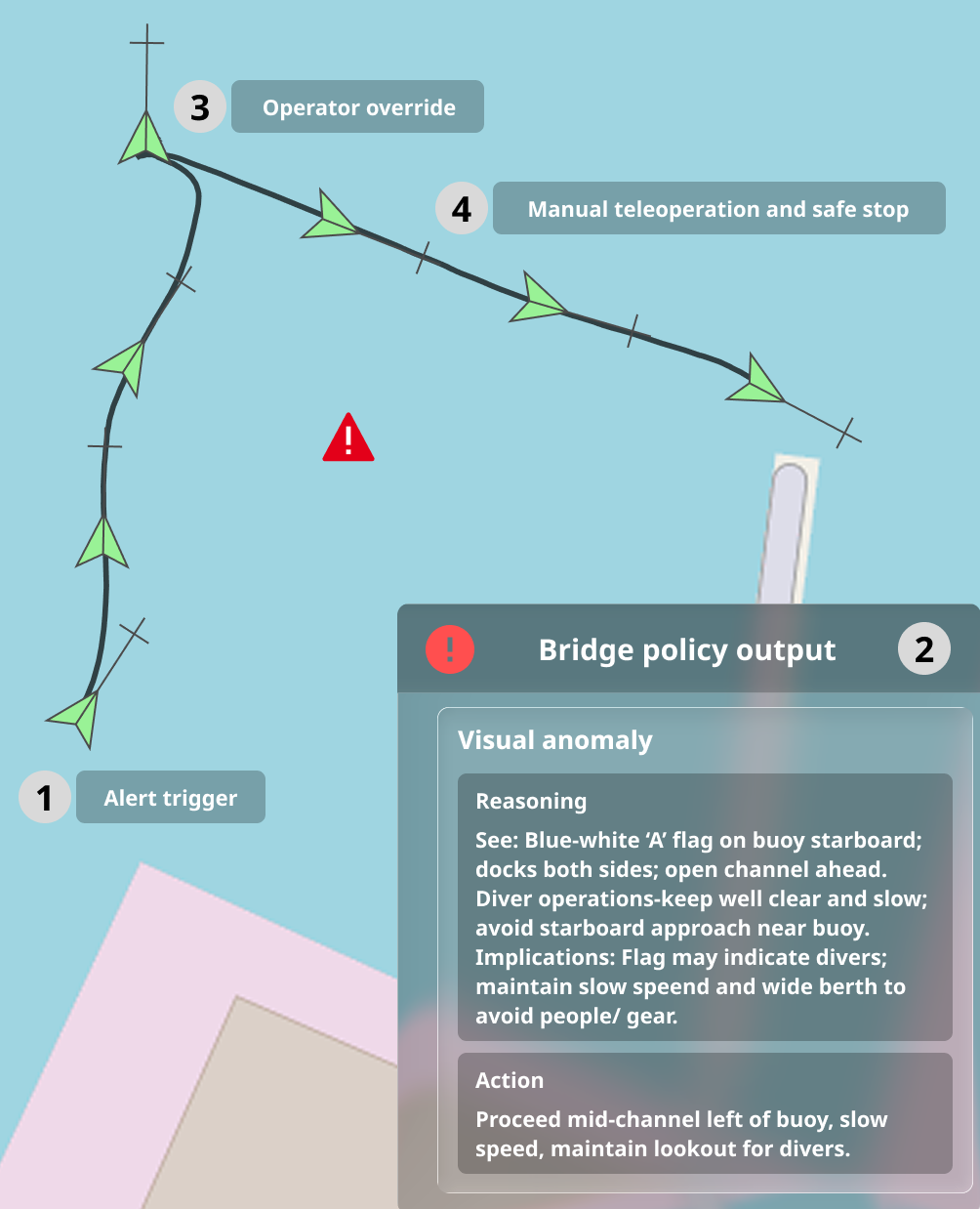}
  \caption{Data from the integrated sea trial showing the chain of events: (1) the alert is manually triggered; (2) the  selector makes a choice; (3) the operator takes manual control and (4) drives to safety.}
  \label{fig:live_example_map}
\end{figure}

The integrated stack functioned as designed on-vehicle, though with high latency (30\,s) because of model choice. Figures~\ref{fig:live_example} (overlay view) and~\ref{fig:live_example_map} (map view) illustrate one representative run: (1) An anomaly alert was raised (manually); (2) the  selector produced the action (number 11) and rationale shown to the operator together with a map like Fig.~\ref{fig:live_example_map}. The model output is reproduced below:

\begin{quote}\small
\textbf{Reasoning:} See: Blue-white ‘A’ flag on buoy starboard; docks both sides; open channel ahead. Diver operations—keep well clear and slow; avoid starboard approach near buoy. \emph{Implications:} Flag may indicate divers; maintain slow speed and wide berth to avoid people/gear. \\
\textbf{Action:} Proceed mid-channel left of buoy, slow speed, maintain lookout for divers.
\end{quote}

\noindent (3) the operator took control from the ROC; and (4) a short manual approach and safe stop at the dock followed. This is a qualitative demonstration that the alert to handover system executes on real hardware on water. Qualitative HMI findings drawn from the same sessions are reported separately in Sec.~\ref{sec:formative_handover}.

\paragraph{Takeaways}
This experiment supports H4 at a proof-of-concept level: the alert$\rightarrow$fallback maneuver$\rightarrow$override chain can run end-to-end on real ASV hardware with joystick override in a ROC user interface. In practice, the main limitation is VLM latency (around 30\,s with the chosen model), suggesting that deployments should pair faster semantic models with the same arbitration and HMI pattern observed in our formative study.

%% file: 07_formative_handover.tex
\section{Additional results: Formative handover HMI study}
\label{sec:formative_handover}

As a complementary evaluation to the closed-loop sea trial in Sec.~\ref{sec:exp-field}, we conducted a small formative handover human-machine interface (HMI) study on the same diver-flag sessions (\(n=5\)) to understand operator needs during the alert-to-override interval. Operators supervised from the ROC, followed a think‑aloud protocol during the run, and completed a brief semi‑structured interview afterward. Analysis used a qualitative, deductive lens mapped to Endsley’s SA levels (L1–L3), with cross‑cuts on workload/attention and trust/mode awareness~\citep{Endsley2023}. The operators had access to a ROC with a wide field of view camera feed with GUI overlay elements, shown in Fig.~\ref{fig:roc}, with additional modules added for our experiment, including the fallback maneuver output and map view like the one shown in Fig.~\ref{fig:live_example_map}. The full experimental protocol is included in~\ref{app:e3} and the complete report is in~\ref{sec:e3_results}.

\subsection{Key observations}
(i) \emph{Mode/authority salience} needed to be persistent and explicit under time pressure. (ii) \emph{Split attention} between map and camera pushed operators to request in‑view AR overlays (hazard label with distance–bearing, recommended corridor, own‑ship path, persistent tracking). (iii) \emph{Text brevity and placement} mattered: short phrasing was preferred over dense text. (iv) \emph{Temporal context} (detection/“last seen”, tracking status) would have improved confidence and reduced unnecessary scanning.

\subsection{Design implications}
The following design principles should be implemented on production systems: Mode/authority should always be visible with explicit transition banners; critical cues should be moved to the camera view with lightweight, continuously updated overlays; the map should be retained for broader context; prefer short, high‑value text with expand‑on‑demand functionality. 

\emph{Limitations:} These are single-scenario tests with small \(n\). The findings are therefore formative only.

%% file: 08_discussion.tex
\section{Discussion}
\label{sec:discussion}

Our results suggest that a candidate-constrained VLM fallback maneuver selector can achieve high maritime scene understanding at practical latencies and translate that into action alignment with human consensus. Sub-10\,s models retain most of the awareness of slow state-of-the-art models (Sec.~\ref{sec:result_reasoning}), and FB-3 alignment exceeds geometry-only baselines (Sec.~\ref{sec:results_alignment}). On fire scenes, where the hazard is unambiguous, the fallback maneuver selector increases standoff distance relative to keep-course/starboard baselines as well as the neutral keep-station (Sec.~\ref{sec:rr_fire}). A live on-water run confirms that the alert$\rightarrow$fallback maneuver$\rightarrow$operator handover executes end-to-end (Sec.~\ref{sec:seatrial_results}), and the formative HMI study highlights how these semantics can be surfaced in a ROC interface (Sec.~\ref{sec:formative_handover}).

High model awareness leads to correct rationales on the ROC GUI overlay, which supports explainable handover: when the fallback maneuver selector recognizes what it is looking at, operators can receive useful, targeted explanations, provided that the user interface makes mode/authority and key cues salient. The HMI study suggests that dense text is limiting under time pressure, and that short, high-value rationales combined with in-view overlays and explicit mode/authority indicators better connect the fallback maneuver selector's semantics to operator situation awareness.

The relative competitiveness of \emph{Keep-station} reflects both our specific dataset and maritime reality: in many scenes, stopping is acceptable or even best. Likewise, starboard-only is often acceptable in open water, especially after we have filtered out obviously unsafe candidates. However, the operational risk resides in the few cases where those defaults are \emph{not} appropriate. There, semantics (e.g., ``fire is dangerous; increase clearance'' or ``divers may be in the water; avoid passing close to the flag'') matter, and the fallback maneuver selector's alignment with human judgment is a proxy for good decisions in such semantic anomalies. The fire risk-relief analysis provides directional validation of that proxy despite low $n$ by showing that, on clearly dangerous hazards, the fallback maneuver selector tends to increase standoff distance while simple geometry-only defaults can move the vessel closer.

The anomaly types also illustrate where current VLMs struggle. Diver flag and MOB scenes present small, low-salience cues that are frequently confused with buoys or background clutter, whereas Fire and Custom sign scenes are salient and explicit, and most models perform well. AI-edited scenes may also be easier (e.g., increased salience or data distribution overlap), but the dominant factor seems to be subtlety and rarity. The internationally recognized alpha diver flag may be underrepresented in pretraining data, and our blue-buoy mounting is atypical. Consistent with this, longer-latency reasoning models tend to do better on Diver-flag recognition and implications. Ultimately, despite 16 alternatives per scene, the best model selects among the top 1--3 human-preferred actions nearly half the time, indicating good but not perfect alignment. Common failure modes include missing the diver-down semantics, confusing distant MOB targets with buoys/clutter, and selecting conservatively when the safest action is not present in the forward camera-constrained candidate set.

Current constraints of the action space and sensors likely limit overall performance. We project straight‑line candidates on a single camera view in front of the vessel. Here we treat $K{=}15$ as a pragmatic proof-of-concept choice rather than an optimized value. Selecting the optimal maneuver library and candidate count $K$ is an important direction for future work. Furthermore, the best action might be outside the current field of view, and camera‑only gating with a water mask is a proxy for true scene geometry. A multi‑sensor bird’s‑eye‑view (e.g., radar/thermal/lidar) for candidate filtering and prompting, plus short‑horizon replanning, should enlarge the reachable safe set and reduce failure modes attributable to missing context. Many of the same limitations apply to the human consensus ground truth; in both cases, the action is chosen from a single 2D camera view. Moreover, the baselines used are also camera‑only proxies for real, more advanced, marine collision avoidance systems that fuse multiple sensors.

From an operational and IMO MASS perspective, we treat the fallback maneuver selector as a pre-approved degraded-mode safety-maneuver method. On ODD exit, it selects within a pre-approved short-horizon envelope (station-keep or steer-away within a given radius) from pre-vetted, water-valid candidates, leaves the voyage plan unchanged, notifies the ROC, and remains immediately overridable. Operationally, we recommend a fast default model ($<$10 s) with repeated re-evaluation under a safety envelope, escalating to slower high-awareness models when time budget permits.

\paragraph{Limitations and scope}
This work is a proof of concept that targets the \emph{post-alert} interval: we evaluate the fallback maneuver selector that runs \emph{after} an anomaly alert has been raised. Comprehensive validation of the upstream anomaly alert (including miss/false-alarm behavior in diverse maritime conditions) is an important but separate problem and remains future work beyond the small-$n$ appendix check. In the live run the alert was triggered manually. Our perception and candidate gating are intentionally camera-only and rely on a water mask as a conservative proxy for range-aware obstacle clearance. Hazards outside the camera field of view, occlusions, and degraded visual conditions (e.g., glare, rain/spray, night) are not addressed in this study. The action space is limited to simplified straight-line, short-horizon candidates and a single-shot decision without replanning, which can exclude otherwise safer maneuvers. The offline dataset covers one harbor/ODD with 40 scenes and includes AI-enhanced fire/sign scenarios used for early-stage validation, while this enables controlled testing of rare hazards, it may not capture real fire dynamics (e.g., wind-driven smoke) or all sources of sensor noise. 

Given the modest number of scenes ($N{=}40$), the fire risk-relief subset ($n{=}10$), and the limited number of closed-loop runs ($n{=}5$), statistical power is limited and some estimates have non-trivial uncertainty. We therefore report 95\% confidence intervals throughout and interpret the results as directional proof-of-concept evidence rather than definitive performance guarantees. Repeated stochastic calls per scene (FB-$n$) probe decision stability, but do not substitute for additional independent scenarios spanning multiple ODDs and environmental conditions. 

Finally, our awareness metric uses an LLM-as-judge and our model evaluations use API-accessed foundation models, so absolute latency and availability may vary across deployments, we therefore emphasize relative comparisons and conservative safe defaults (Sec.~\ref{sec:methods-bridge}).

\paragraph{Future work}
The anomaly$\rightarrow$fallback maneuver$\rightarrow$override loop provides a path to collect labeled scenes and operator outcomes for fine-tuning and continuous improvement, potentially leading to more domain-adapted VLMs and fallback maneuver selectors. To improve reproducibility and reduce API- and connectivity-related failure modes, future work should include evaluations with open-weight, vision-centric models (e.g., Qwen-VL~\citep{bai2023qwenvlversatilevisionlanguagemodel}) and smaller locally deployable VLMs (e.g., 7B--30B) that can run on the ROC workstation or onboard. This would enable more predictable latency and availability, and provides a practical path toward domain adaptation (fine-tuning etc.) while retaining the same candidate-constrained interface and human-override guarantees. Next steps include expanding the dataset to include more real-world operational data, performing sensitivity and ablation experiments (e.g., measuring awareness--alignment correlation), integrating a multi-sensor BEV with a type of receding-horizon method, and evaluating non-motion responses (e.g., VHF communication) where appropriate.

%% file: 09_conclusion.tex
\section{Conclusion}
\label{sec:conclusion}

We presented a proof-of-concept, camera-first, candidate-constrained VLM fallback maneuver selector that turns semantic scene understanding into short-horizon actions while keeping explicit human authority. Across 40 harbor scenes, fast models retained most of the awareness of slower systems, and the FB-3 selector aligned better with human preference than geometry-only baselines. On the unambiguously dangerous fire hazards, the fallback maneuver selector increased standoff distance, and a live on-water run verified the alert $\rightarrow$ fallback maneuver $\rightarrow$ override chain. Together, this supports VLMs as a semantic fallback maneuver selector that fits the IMO MASS Code-style degraded mode fallback maneuver within practical latency budgets. We believe this work can support the transition to one-to-many supervision of maritime autonomy. 

Looking ahead, the main challenges are: handling rare, low-salience cues;  broader context than the single camera view; and decisions that evolve over time rather than a single shot. We see the path forward in domain-adapted models and pairing foundation-model semantics with a multi-sensor BEV and receding-horizon control, while using the general anomaly alert $\rightarrow$ fallback maneuver $\rightarrow$ override loop to turn override situations into training data for continual improvement.

%% file: appendix_fast_anomaly.tex
We adapt the fast embedding-based anomaly monitor of~\cite{sinha2024real}\ to maritime camera scenes and show it works out of the box on small-\textit{n} datasets. We use it only as a detector in a detect$\rightarrow$fallback maneuver$\rightarrow$override chain, the fallback maneuver stages are described elsewhere in our paper.

\subsection{Setup and notation}
Let $o_t \in \mathcal O$ denote the current camera observation.
Following Sinha et~al., we construct a nominal experience set $\mathcal D_{\mathrm{nom}}=\{o_i\}_{i=1}^N$ and a text-embedding function $\phi$ that maps a scene description to $e \in \mathbb R^d$.
We create a nominal embedding cache $\mathcal D_e=\{e_i\}_{i=1}^N$ offline, where $e_i=\phi(o_i)$. At runtime we obtain an embedding $e_t=\phi(o_t)$ and score its deviation from prior experience.

\subsection{Image to embedding}
Each image is first summarized into a single, concise, navigation-centric sentence by a VLM (GPT-4o).
We explicitly instruct: one sentence, focus on navigationally relevant objects/hazards, exclude own-ship/people on board.
This text is then embedded with an off-the-shelf text embedding FM (OpenAI \texttt{text-embedding-3-small}), yielding $e_t$.

\subsection{Score and decision rule}
We use the max-cosine similarity heuristic from Sinha et~al. and define the anomaly score
\begin{equation}
s(e_t;\mathcal D_e)\;=\;1-\max_{e_i\in\mathcal D_e}\;\frac{e_i^\top e_t}{\|e_i\|\,\|e_t\|}\,.
\label{eq:score}
\end{equation}
Equivalently, this is the negative of the maximum cosine similarity up to a constant shift.
To classify, we calibrate a threshold $\tau$ by leave-one-out on the nominal cache and take the empirical $\alpha$-quantile,
\begin{equation}
\tau \;=\; \inf\left\{q\in\mathbb R\;:\;\frac{1}{N}\,\big|\{\,e_i\in\mathcal D_e: s(e_i;\mathcal D_e\!\setminus\!\{e_i\}) \le q\,\}\big|\;\ge\;\alpha\right\},
\label{eq:quantile}
\end{equation}
where $\alpha=0.95$. We declare $o_t$ anomalous if $s(e_t;\mathcal D_e)>\tau$.

\subsection{Data}
\begin{figure}[!tbp]
  \centering

  \begin{subfigure}{\linewidth}
    \centering
    \includegraphics[width=\linewidth]{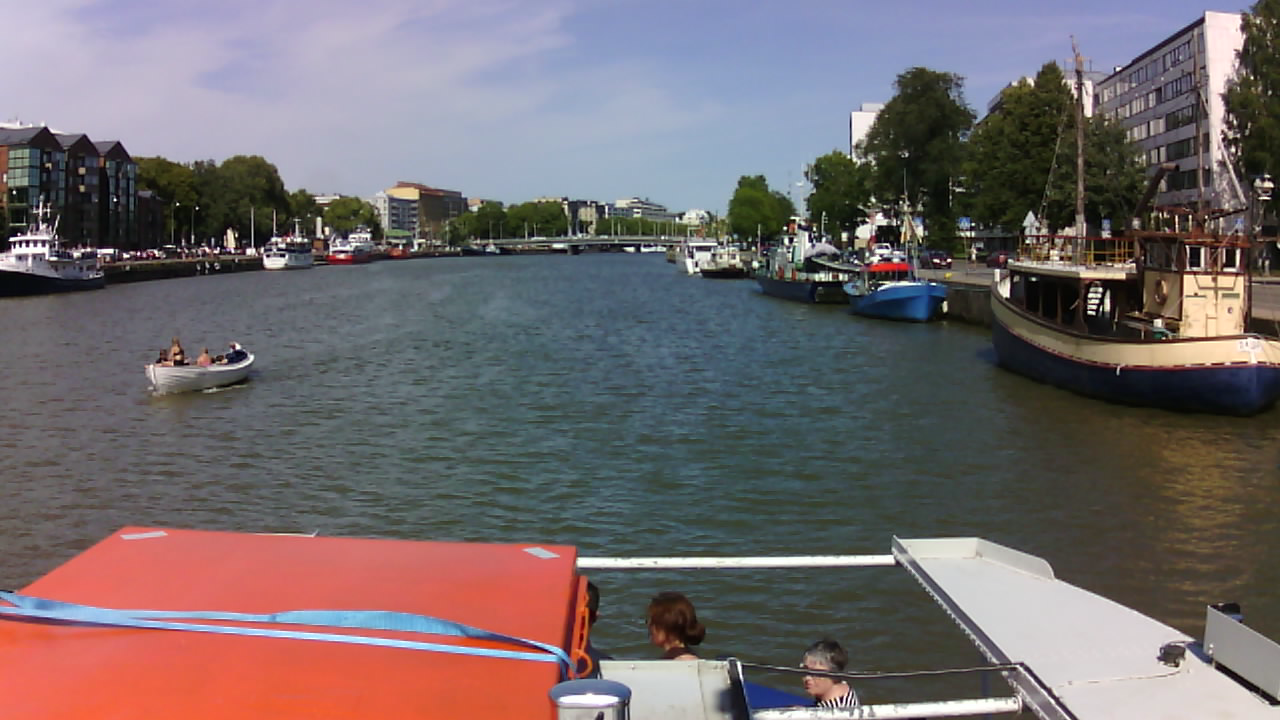}
    \caption{Nominal frame from ABOSHIP PLUS.}
  \end{subfigure}

  \vspace{0.4em}

  \begin{subfigure}{\linewidth}
    \centering
    \includegraphics[width=\linewidth]{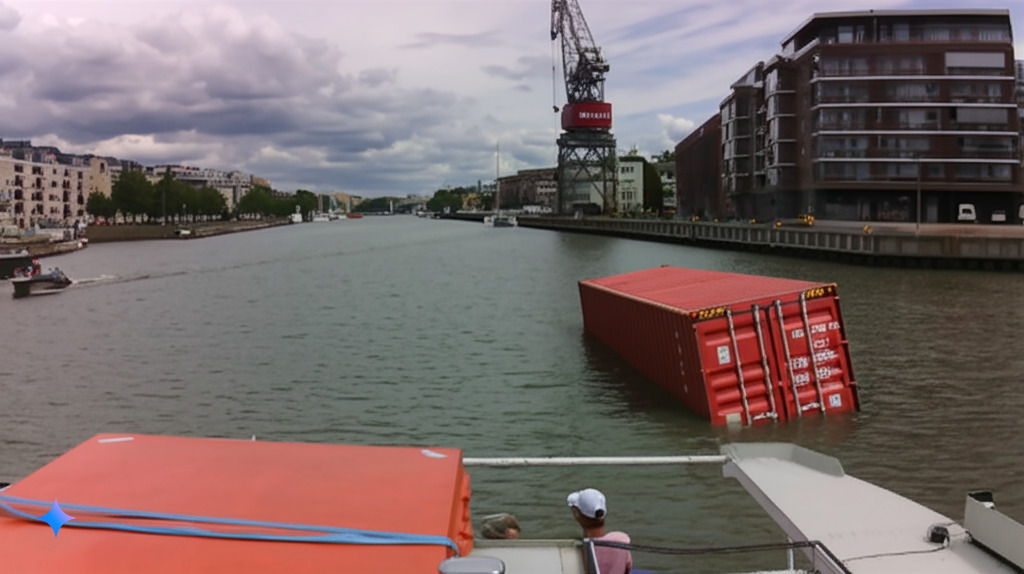}
    \caption{Synthetically edited anomalous frame (same dataset).}
  \end{subfigure}

  \caption{Examples from the dataset used for fast anomaly detection: (a) nominal, (b) synthetic anomaly.}
  \label{fig:maritime-nom-anom}
\end{figure}

Nominal frames come from the \emph{ABOSHIP PLUS} dataset.
We reserve 1000 nominal images for the cache $\mathcal D_e$ and 100 nominal images for testing.
We synthesize 10 anomaly scenes (e.g., dock on fire, ice sheets, floating containers) to test anomalies.
See Fig.~\ref{fig:maritime-nom-anom} for one nominal and one synthetic examples.

\subsection{Calibration and runtime}
With $\alpha=0.95$ we obtain $\tau=0.2375$, i.e.\ flag anomaly if $\max\cos<0.7625$.
End-to-end rate, including cloud VLM inference, is in the order of $0.5$-$1$~Hz.

\subsection{Results}
\begin{figure}[!htbp]
  \centering
  \includegraphics[width=\linewidth]{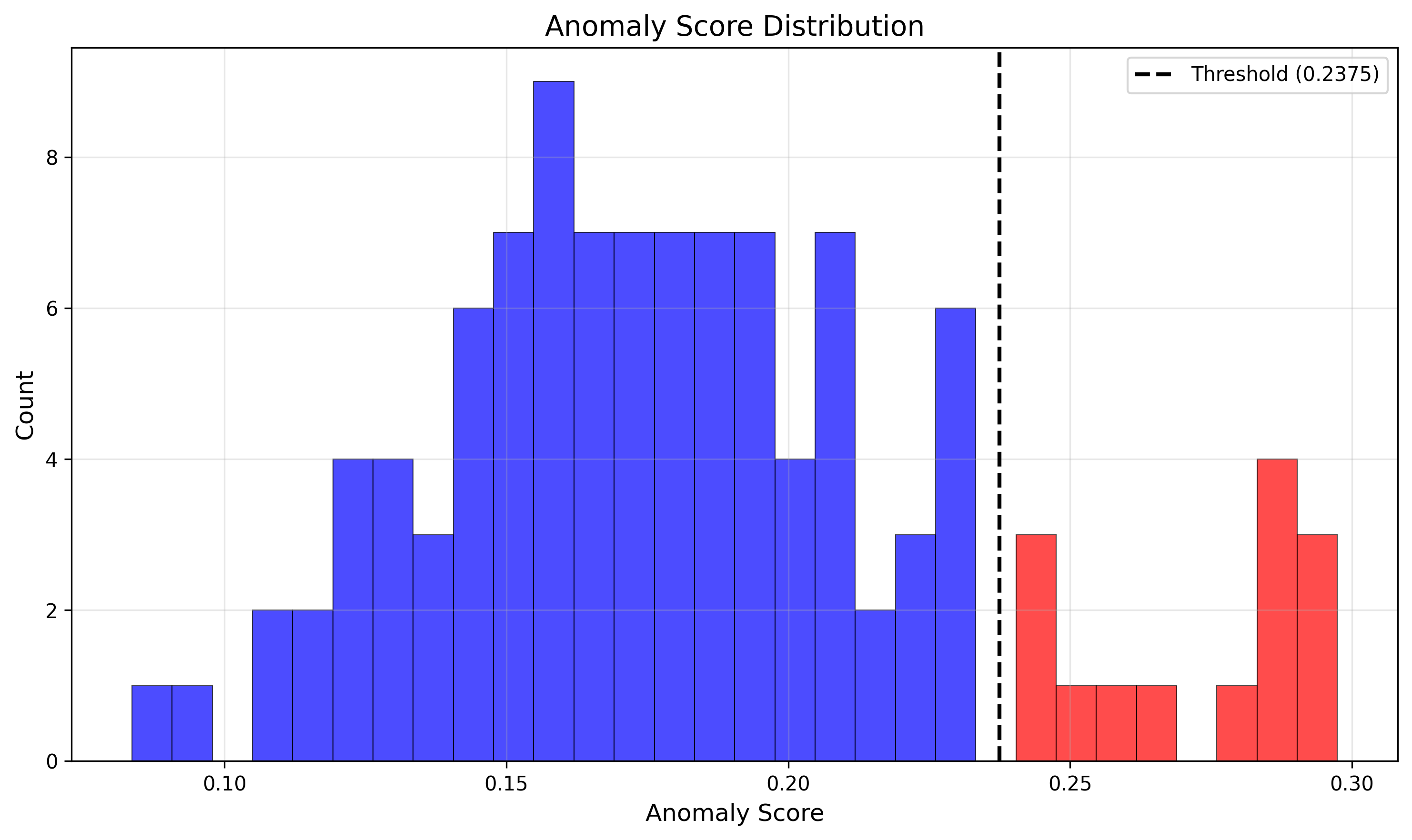}
  \caption{Histogram showing the theshold and nominal/anomalous frames from the fast anomaly detection experiment.}
  \label{fig:fast_histogram}
\end{figure}

The small test shows a true positive rate $\mathrm{TPR}=1.00$ and false positive rate $\mathrm{FPR}\approx 0.04$.
A score histogram (Fig.~\ref{fig:fast_histogram}) show the separation between nominal frames and anomalies.

\subsection{Limitations and role in the stack}
The evaluation is small-$n$ and uses synthetic hazards to stress-test rare conditions.
Nevertheless, the detector is relatively fast, model-agnostic, and semantically grounded by prior experience. 
In our full system, it triggers a fallback maneuver that keeps the ASV safe and legible until a human takes control, consistent with MASS-style detect$\rightarrow$fallback$\rightarrow$override requirements.

%% file: appendix_platform.tex

\subsection{milliAmpere1: Experimental ASV prototype}
\label{app:platform}

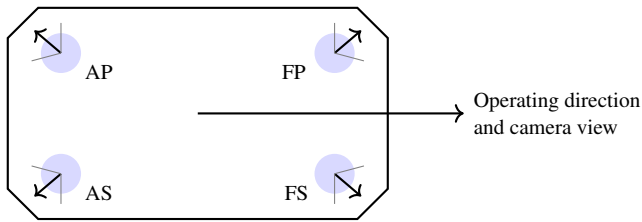
\begin{figure}[h]
    \centering
    \input{TikZ_milliAmpere}
    \caption{Thruster configurations on milliAmpere. Fixed angles are used in low‑speed DP; in transit, the two fore thrusters run in reverse as in Fig.~\ref{fig:milliAmpere-configuration-rudder-4}.}
    \label{fig:milliAmpere-configuration}
\end{figure}

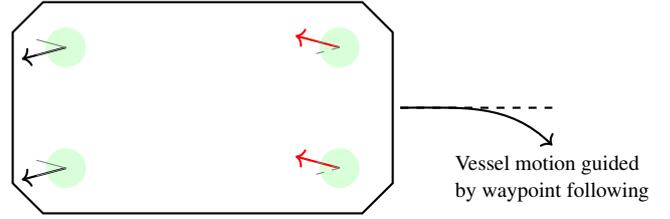
\begin{figure}[h]
    \centering
    \input{TikZ_4-rudder}
    \vspace{-1em}
    \caption{Automatic steering allocation by azimuth turning. Black arrows: push; red: pull (reverse). In transit, azimuths are adjusted within a $\pm 15^\circ$ sector for path following.}
    \label{fig:milliAmpere-configuration-rudder-4}
\end{figure}

The milliAmpere1 is a compact urban ASV research platform for full‑scale testing of autonomy, control, and situational awareness (SA) sensing \citep[]{Brekke2022}. The vessel supports onboard autonomy and remote teleoperation via a shore control lab, with RTK GNSS/IMU navigation and an SA sensor rig (EO/IR, X‑band radar, LiDAR) \citep[]{Brekke2022}. For our experiments, the hull is configured with \emph{four} azimuth thrusters (Table~\ref{tab:ma1-characteristics}) enabling low‑speed DP and transit “virtual rudder’’ steering (Figures~\ref{fig:milliAmpere-configuration}–\ref{fig:milliAmpere-configuration-rudder-4}). Data/teleop use a 5G link; a line‑of‑sight setup was also used during formative testing.\nocite{Alsos2022}

\begin{table}[h!]
{
\footnotesize
    \centering
    \caption{Technical specifications for milliAmpere1.}
    \begin{tabular}{l|l}
        \hline
        \textbf{Parameter} & \textbf{Value} \\
        \hline
        Length, $L$ & 5.0 m \\        
        Breadth $B$ & 2.8 m \\        
        Draft, $T$ & 0.6 m \\      
        Nominal operation speed, $U$  & 2.0 knots \\        
        Operation speed, anomaly & 1.0 knots \\        
        Propulsion & 4 azimuth thrusters \\
        Energy system & Electric, 24V DC \\        
        Navigation & RTK GNSS-compass, IMU \\        
        SITAW sensors  & IR/EO cameras, X-band radar, LiDAR \\        
        \hline
    \end{tabular}
    \label{tab:ma1-characteristics}
}
\end{table}

\subsection{Automatic waypoint following}

\nocite{Hinostroza2025_sysid}
\nocite{Hinostroza2025_software}
\nocite{Tufte2025IntegratedMotion}

Given the discrete fallback maneuver action $a\in\{0,\ldots,K\}$ (Sec.~\ref{sec:methods-bridge}), the system either tracks a world‑fixed path ($a\ge 1$) or holds position ($a=0$). For $a\ge 1$, a pre‑frozen world polyline $\Pi_a$ is tracked by an LOS follower with acceptance radius $R=7.5$\,m and lookahead $\Delta=10$\,m at $U_{\mathrm{anom}}=1.0$\,kn; controller/allocation details follow \cite{Tufte2025IntegratedMotion}. For $a=0$ (station‑keeping) a single world point at the current pose is published.

\subsection{Direct joystick override}

\paragraph{Functionality}
Let $\boldsymbol{\tau}_d$ be the arbitrated wrench, $\boldsymbol{\tau}_m$ the autonomy command, $\boldsymbol{\tau}_h$ the human (joystick) input, and $\alpha\in[0,1]$ a blending factor (Fig.~\ref{fig:override-illustration}). We use a direct blend guaranteeing at least $50\%$ human authority and ramping to full override:
\begin{equation}
\boldsymbol{\tau}_d \;=\; (1-\alpha)\,\boldsymbol{\tau}_m \;+\; \bigl(0.5 + 0.5\,\alpha\bigr)\,\boldsymbol{\tau}_h,
\end{equation}
with $\alpha=0$ (automatic), $0<\alpha<1$ (shared), $\alpha=1$ (override).

\paragraph{Implementation}
The override parameter $\alpha$ is ramped by
\begin{equation}
    \dot{\alpha} = \begin{cases}
        \tfrac{1}{T}, & |\boldsymbol{\tau}_\text{h}| > {\tau}_\text{h}^\text{min} \\
        0, & \alpha = 1 \wedge t - t_c < T \\
        -\tfrac{1}{T}, & t-t_c \geq T
    \end{cases}
\end{equation}
where $T$ is the transition time, ${\tau}_\text{h}^\text{min}$ a joystick threshold, and $t_c$ the last threshold crossing. The mixer acts in generalized wrench space \emph{upstream} of allocation so actuator limits are enforced uniformly for autonomy, human, or blends.

\subsection{ROC and test infrastructure}

Our experiments use (i) the \textit{milliAmpere1} research ASV with added sensors and comms for remote operation \citep[]{Brekke2022}, (ii) the \textit{Shore Control Lab} hosting the operator workstation(s) \citep[]{Alsos2022}, and (iii) the \textit{Nyhavna} harbor basin as a sheltered test site. For formative handover tests we used the \emph{Prometheus} ROC prototype with an added GUI; operations were run over 5G and, where relevant, in line‑of‑sight setups \citep[]{gusev2025situation}.

%% file: TikZ_milliAmpere.tex
\begin{tikzpicture}[scale=1]

  \def\L{5}   
  \def\B{2.8} 
  \def\c{0.4} 

  \draw[thick]
    (-\L/2+\c, -\B/2) --
    (\L/2-\c, -\B/2) --
    (\L/2, -\B/2+\c) --
    (\L/2, \B/2-\c) --
    (\L/2-\c, \B/2) --
    (-\L/2+\c, \B/2) --
    (-\L/2, \B/2-\c) --
    (-\L/2, -\B/2+\c) -- cycle;

  \def\lx{1.8}  
  \def\lX{1.3}  
  \def\ly{0.8}  

  \foreach \x/\y/\name in {
      \lX/\ly/{\footnotesize FP \qquad},
      \lX/-\ly/{\footnotesize FS},
      -\lX/\ly/{\footnotesize AP},
      -\lX/-\ly/{\footnotesize AS}
  }{
    \node at (\x,\y-0.25) {\name};
  }

  \foreach \x/\y in {
      \lx/\ly,
      \lx/-\ly,
      -\lx/\ly,
      -\lx/-\ly
  }{
    \filldraw[blue!15, thick] (\x,\y) circle (0.25);
  }

  \draw[->, black, thick] (\lx,\ly)   -- ++(0.35,0.3);
  \draw[->, black, thick] (\lx,-\ly)  -- ++(0.35,-0.3);
  \draw[->, black, thick] (-\lx,-\ly) -- ++(-0.35,-0.3);
  \draw[->, black, thick] (-\lx,\ly)  -- ++(-0.35,0.3);

  \begin{scope}
    \path (\lx,\ly) coordinate (T);
    \draw[gray] (T) -- ++(-15:0.4);
    \draw[gray] (T) -- ++(90:0.4);
  \end{scope}

  \begin{scope}
    \path (\lx,-\ly) coordinate (T);
    \draw[gray] (T) -- ++(15:0.4);
    \draw[gray] (T) -- ++(-90:0.4);
  \end{scope}

  \begin{scope}
    \path (-\lx,-\ly) coordinate (T);
    \draw[gray] (T) -- ++(-90:0.4);
    \draw[gray] (T) -- ++(165:0.4);
  \end{scope}

  \begin{scope}
    \path (-\lx,\ly) coordinate (T);
    \draw[gray] (T) -- ++(195:0.4);
    \draw[gray] (T) -- ++(90:0.4);
  \end{scope}
  
  \draw[->, thick] (0,0) -- (3.5,0) node[right] {\footnotesize \shortstack[l]{Operating direction \\ and camera view}};

\end{tikzpicture}

%% file: TikZ_4-rudder.tex
\begin{tikzpicture}[scale=1]

  \def\L{5}   
  \def\B{2.8} 
  \def\c{0.4} 
  \def\d{0.1} 
  \draw[thick]
    (-\L/2+\c, -\B/2) --
    (\L/2-\c, -\B/2) --
    (\L/2, -\B/2+\c) --
    (\L/2, \B/2-\c) --
    (\L/2-\c, \B/2) --
    (-\L/2+\c, \B/2) --
    (-\L/2, \B/2-\c) --
    (-\L/2, -\B/2+\c) -- cycle;



  \def\lx{1.8}  
  \def\lX{1.3}  
  \def\ly{0.8}  

  \foreach \x/\y/\name in {
      \lX/\ly/{\footnotesize \qquad},
      \lX/-\ly/{\footnotesize },
      -\lX/\ly/{\footnotesize },
      -\lX/-\ly/{\footnotesize }
  }{
    \node at (\x+0.1,\y-0.8) {\name};
  }

  \foreach \x/\y in {
      \lx/\ly,
      \lx/-\ly,
      -\lx/\ly,
      -\lx/-\ly
  }{
    \filldraw[green!15, thick] (\x,\y) circle (0.25);
  }

  \draw[->, red, thick] (\lx,\ly)   -- ++(-0.580,0.155);
  \draw[->, red, thick] (\lx,-\ly)  -- ++(-0.580,0.155);
  \draw[->, black, thick] (-\lx,-\ly) -- ++(-0.580,-0.155);
  \draw[->, black, thick] (-\lx,\ly)  -- ++(-0.580,-0.155);

  \begin{scope}
    \path (\lx,\ly) coordinate (T);
    \draw[gray, dashed] (T) -- ++(195:0.4);
    \draw[gray, dashed] (T) -- ++(165:0.4);
  \end{scope}

  \begin{scope}
    \path (\lx,-\ly) coordinate (T);
    \draw[gray, dashed] (T) -- ++(195:0.4);
    \draw[gray, dashed] (T) -- ++(165:0.4);
  \end{scope}

  \begin{scope}
    \path (-\lx,-\ly) coordinate (T);
    \draw[gray] (T) -- ++(195:0.4);
    \draw[gray] (T) -- ++(165:0.4);
  \end{scope}

  \begin{scope}
    \path (-\lx,\ly) coordinate (T);
    \draw[gray] (T) -- ++(195:0.4);
    \draw[gray] (T) -- ++(165:0.4);
  \end{scope}
  

    \draw[dashed, thick] 
    (\L/2+0.1, 0) -- (\L/2+2.1, 0);

    \draw[->, thick] 
    (\L/2+0.1, 0) 
        to[out=0, in=135] node[below, pos=1] {\footnotesize \shortstack[l]{Vessel motion guided \\ by waypoint following}}(\L/2+2.1, -0.5);

\end{tikzpicture}

%% file: appendix_formative_handover.tex
This section describes the setup, interview protocol and results of the formative handover study with $n=5$ participants overriding the real ASV during a live sea trial. 

\subsection{Expert interview protocol}
\label{app:expert-interview-protocol}

\paragraph*{Study Design}
Scenario-based expert interviews using a cognitive walkthrough approach \citep{Wharton1994},
aligned with human-centered design principles \citep{ISO9241-210:2019}.
The method targeted three core constructs emphasized in human--autonomy teaming frameworks: situational awareness, workload, and trust. \citep{Endsley1995, ONeill2020, Miller2024}

\paragraph*{Procedure}
Experts were given the scenario of being a captain responsible for 10 ferries in a shore control center. When an anomaly was detected on one vessel, they were instructed to monitor and, if needed, take control at the teledrive station. Participants engaged in a structured walkthrough of the interface while using a think-aloud protocol. Moderators provided only neutral prompts when necessary. All sessions were audio- and screen-recorded with internally selected participants.

\paragraph*{Analysis}
Notes and transcripts were coded thematically. Responses were categorized under situational awareness, interface clarity, workload, and trust.


\paragraph*{Protocol script}
\begin{enumerate}
  \item \textbf{Participant brief}
  \begin{enumerate}
    \item \textbf{Explanation about experiment}
      \begin{enumerate}
        \item You will take a part of an experiment evaluating human machine interaction and takeover procedure during anomaly alert. The idea behind the system is having multiple autonomous vessels being operative at the same time, where manual assessment and takeover might be needed.
      \end{enumerate}
    \item \textbf{Explanation about data collection}
      \begin{enumerate}
        \item We will take audio record for question answer analysis and video footage of the screen for further analysis of the GUI enhancements.
      \end{enumerate}
    \item \textbf{Explanation about equipment}
      \begin{enumerate}
        \item Joystick, throttle controller, map, compass, (some parts are dummies). Joystick can be used to take over any time during autonomous operation.
      \end{enumerate}
    \item \textbf{Training in think-aloud protocol}
      \begin{enumerate}
        \item Example: I see that the boat is standing still on the left side I can spot a sailboat not far from the boat. I think I will take over by using the joystick now because the boat is to close.
      \end{enumerate}
  \end{enumerate}

  \item \textbf{Test drive the vessel without anomaly}
  \begin{enumerate}
    \item \textbf{Training in think-aloud protocol}
      \begin{enumerate}
        \item Move the boat forward
        \item Move the boat left
        \item Rotate the boat 180 degrees
        \item Try some free movements
      \end{enumerate}
  \end{enumerate}

  \item \textbf{Ask a few questions about the GUI (short)}
  \begin{enumerate}
    \item Any comments or questions about the system before we start?
  \end{enumerate}

  \item \textbf{Brief about anomaly}
  \begin{enumerate}
    \item ``Imagine that you are the captain in charge of multiple ferries at the shore control center. All vessels operate autonomously without the need for manual control. However, when an anomaly is detected, the system notifies the operator, and the vessel experiencing the anomaly is automatically displayed in the teledrive station. An anomaly is defined as a situation the autonomy has never encountered before or cannot recognize as familiar. Test drive with anomaly.''
    \item Summarized: 1) You will need to take control of the vessel during an anomaly detection and 2) start bringing it closer to the closest dock. Which means that you first need to assess the situation and then do the correct takeover.
  \end{enumerate}

  \item \textbf{Test}
  \begin{enumerate}
    \item Person sitting on the side of the teledrive station.
    \item Colleagues set the boat into position.
    \item The test person should be taken to the teledrive station when the reasoning system has made a decision and the autonomy has started evasive maneuver.
    \item Person should talk aloud during the assessment and take over the control of the boat.
  \end{enumerate}
\end{enumerate}

\paragraph*{Debrief}
\begin{enumerate}
  \item \textbf{Initial impression}
  \begin{itemize}
    \item Did you understand what happened right now?
    \item Did you understand what the anomaly was?
  \end{itemize}

  \item \textbf{Situation understanding}
  \begin{itemize}
    \item Did you understand what the autonomy did?
  \end{itemize}

  \item \textbf{Interface clarity \& information}
  \begin{itemize}
    \item Did the GUI told you when anomaly was detected?
    \item Was any of the information presented in the GUI unclear?
    \item Did you look at the map?
  \end{itemize}

  \item \textbf{Next steps \& takeover}
  \begin{itemize}
    \item Based on the information shown, how did you understand your next steps during the takeover?
    \item Did you feel you had enough information to take control?
  \end{itemize}

  \setcounter{enumi}{5}
  \item \textbf{Cognitive load / stress}
  \begin{itemize}
    \item Did the amount of information and the way it was presented stress you, or did it feel manageable?
  \end{itemize}

  \item \textbf{Overall reflection}
  \begin{itemize}
    \item What do you think about anomaly detection with explainable AI that explains what happens?
    \item Would you have used this system as a captain?
    \item Any comments from your side
  \end{itemize}
\end{enumerate}

%% file: appendix_formative_results.tex
\subsection{Situational awareness (SA) framing}
Safe and effective remote vessel operation is, by many means, dictated by Situational Awareness (SA). Endsley's definition states; ``\textit{the perception of the elements in the environment within a volume of time and space, the comprehension of their meaning, and the projection of their status in the near future}'' \cite[p.\>36]{Endsley1995}. Her model breaks SA into three levels: \textit{Perception of Elements in Current Situation} (Level 1 SA), \textit{Comprehension of Current Situation} (Level 2 SA), and \textit{Projection of Future Status} (Level 3 SA). The evaluation of sensory modalities in the graphical user interface of the anomaly detection module (GUI) is based on this 3-level model. This is consistent with known human-factors issues in shore control centers for unmanned or highly automated ships (e.g., information overload, mode confusion, timing/latency effects) and highlights the need for interfaces that externalize system state, intent, and uncertainty \cite[p.\>53]{Rutledal2024}; \cite[p.\>2]{wahlstrom2015human}. This design challenge can be framed in terms of Level 1 SA, since incoming signals to the operator's perceptual system that encode the vessel's state (e.g., through sight, hearing, and touch) are affected, which may have knock-on effects for Level 2 and Level 3 SA. In this context, we developed an anomaly detection and handling module with a GUI intended to support the operator in quickly understanding what the system has detected, what autonomy is doing, and when/where manual takeover is appropriate.

\subsection{Formative study: Participants, scenario, and method}
This formative study included a total of five test persons, where three had no experience driving a boat nor any teledriving experience, and two had boating experience but no teledriving experience. The study was intended to be formative for future iterations of the prototype rather than statistically strong. Nevertheless, it yielded qualitative feedback on how an operator could understand what was happening with the boat, what the autonomy was doing, and when to take over using the TeleDrive station and the anomaly detection system.

Participants were briefed as captains supervising multiple autonomous ferries from a shore control center. If the system detected an anomaly on a vessel, they were directed to the TeleDrive station to monitor the situation and, when required, take manual control to steer the vessel to the nearest visible dock. In this scenario, the anomaly was a diver’s flag mounted on a blue buoy; participants were not told this in advance and could only perceive it from the scene or infer it from the GUI’s AI explanation. During the operation phase, participants followed a think-aloud protocol while moderators provided only neutral prompts when necessary. After the run, with the GUI visible, participants completed a structured interview and interface walkthrough. All sessions (operation and interview) were audio- and screen-recorded for analysis.

By this structure, we were able to get valuable feedback on how the operators were perceiving the information based on the levels of Endsley's model.

\subsection{Analysis approach}
We conducted a qualitative, deductive thematic analysis mapped to Endsley’s three levels of SA (L1 perception, L2 comprehension, L3 projection), with two cross-cutting categories: (i) workload/attention management and (ii) trust/mode awareness. No rating scales were used, and no quantitative scoring was produced; findings are reported narratively based on participants’ descriptions and representative statements.

\subsection{Results}
\subsubsection{Task outcome}
All five participants identified the anomaly to some extent, executed a safe takeover when needed, and maneuvered safely without collisions or misalignment to the closest dock. Although they used and perceived the system differently,  e.g., some relied mainly on the map view, others on live camera feeds, all agreed the anomaly-detection module was more useful than having no system guidance.

\subsubsection{Information flow, workload, and confidence}
The two-screen layout split attention: the top screen showed the anomaly banner and AI text, while the bottom screen presented the map (vessel position, past/predicted path, anomaly area). When participants focused on the camera view, map-layer details (detection point/marker, time-since chip, predicted path) were often missed; when they relied on the map, in-camera scene dynamics received less attention. Participants explicitly asked for course/trajectory overlays \emph{in} the camera to reduce toggling and attention split. They consistently described the AI text as too long for the moment of action and asked for shorter, more direct phrasing placed near the relevant visual; no comparison to concise alternatives was tested. Some reported stress from juggling cameras, text, and the map; others coped by ignoring text and relying on visuals. Confidence in the system was described as higher when information was at the point of gaze and up to date.

\subsubsection{SA Level 1: Perception}
Participants often described uncertainty about when the anomaly was first detected and what had happened since. Although indicators existed (e.g., time-since and a detection marker on the map), these cues were frequently overlooked under time pressure. This indicates a salience and placement problem rather than an information gap: critical cues should be rendered directly in the primary camera view with AR overlays (hazard outline/label with distance–bearing and a short actionable line), while verbose text is reduced.

\subsubsection{SA Level 2: Comprehension}
Mode and locus of control were often ambiguous. Several participants said they did not fully understand whether autonomy was acting or only suggesting, what the AI/autonomy was trying to achieve, or when it attempted to steer away from the diver flag. This points to the need for explicit mode/authority indicators and brief, context-coupled rationales (“why now?”), co-located with the scene, so users do not need to shift attention to interpret intent.

\subsubsection{SA Level 3: Projection}
Participants wanted dynamic, in-view prediction and continuity: clear trajectory lines for the vessel, a recommended safe corridor, and continuous target tracking so hazards don’t “disappear” between frames. The lack of tracking made it harder to anticipate near-future states and to verify that suggested manoeuvres would keep safe clearance.

\subsubsection{Information strategy and workload}
The two-screen layout split attention: the top screen contained the anomaly banner and AI text; the bottom screen presented the map with vessel position, past/predicted path, and an anomaly-area overlay. When participants focused on the camera, map-layer details (detection point, time-since chip, predicted path) were easy to miss; when they relied on the map, in-camera scene dynamics received less attention. Several described the AI text as too verbose for the moment of action. Participants expressed greater confidence (trust) when information was concise, up-to-date, and placed where they were already looking; conversely, long text blocks or dispersed cues reduced perceived utility. Overall workload management favoured camera-first scanning with minimal reading, reinforcing the need for clear AR in the camera view and brief audio prompts.

\subsubsection{Design implications and recommended improvements}
The qualitative findings indicate three priorities for future iterations. 
\textbf{Mode awareness:} keep the current mode (Autonomy/Manual/Hybrid) continuously visible, make transitions explicit, and clearly indicate when the AI initiates a safety manoeuvre that departs from the nominal course; requests for human takeover should be more salient but not the only cue. 
\textbf{Point-of-gaze cues:} place critical anomaly information directly in the camera view using AR (hazard outline/label with distance–bearing, a recommended corridor, and clear trajectory lines), add lightweight temporal context (detection and tracking status with time-since/last-seen), and maintain continuous target tracking with uncertainty cues. 
\textbf{Map role:} retain the map as a complementary view for broader route context and precise geometry, and ensure markers/annotations persist after the anomaly area is left. 
\textbf{AI text:} shorten and simplify; use direct, high-value phrasing local to the scene and avoid paragraphs during time-critical manoeuvres.

\subsection{Limitations}
This was a small, internal sample ($n=5$) with a single scenario. The system is work-in-progress (e.g., no continuous tracking), which likely suppressed mode awareness and temporal clarity. These are the exact gaps addressed by the proposed design changes.

\subsection{Conclusion} Using Endsley’s SA framework as an evaluation lens, the formative study shows that the system already supports successful anomaly handling, but clarity of control authority, temporal context, and dynamic, in-place visualization are the biggest levers to lift SA (Levels 1--3), reduce workload, and strengthen calibrated trust. These findings align with broader human-factors literature on remote maritime operations.

%% file: appendix_misc.tex
\subsection{Farthest point thinning algorithm}

The farthest point thinning algorithm used in Sec.~\ref{sec:methods-cands} is described in Alg.\>\ref{alg:farthest}. 

\begin{algorithm}[t]
\small
\caption{Farthest-point thinning on endpoint pixels}
\label{alg:farthest}
\begin{algorithmic}[1]
\REQUIRE Survivors $\mathcal{S}=\{(e_i,\mathbf{u}_i)\}_{i=1}^{n}$ with endpoint pixels $\mathbf{u}_i\in\mathbb{R}^2$; target $K$; optional min separation $\delta_{\mathrm{px}}\ge 0$
\ENSURE Index set $\mathcal{I}=\{i_1,\dots,i_{K'}\}$, $K'\le\min(K,n)$; selection order defines IDs $1..K'$
\IF{$n \le K$ \AND $\delta_{\mathrm{px}}=0$} \RETURN $\{1,\dots,n\}$ \ENDIF
\STATE Precompute $d^2_{ij}=\|\mathbf{u}_i-\mathbf{u}_j\|_2^2$; set $d^2_{ii}=\infty$
\STATE $i_1 \leftarrow \arg\max_i \min_{j\neq i} d^2_{ij}$ \hfill\textit{\# most isolated}
\STATE $\mathcal{I}\leftarrow [\,i_1\,]$;\quad $\mathcal{U}\leftarrow \{1,\dots,n\}\setminus\{i_1\}$
\WHILE{$|\mathcal{I}|<K$ \AND $\mathcal{U}\neq\emptyset$}
  \FOR{each $i\in\mathcal{U}$} \STATE $m_i \leftarrow \min_{j\in\mathcal{I}} d^2_{ij}$ \ENDFOR
  \STATE $i^* \leftarrow \arg\max_{i\in\mathcal{U}} m_i$ \hfill\textit{\# farthest from selected}
  \IF{$(\delta_{\mathrm{px}} > 0) \land (m_{i^*} < \delta_{\mathrm{px}}^2)$}
  \STATE \textbf{break}
\ENDIF
  \STATE Append $i^*$ to $\mathcal{I}$;\quad $\mathcal{U}\leftarrow\mathcal{U}\setminus\{i^*\}$
\ENDWHILE
\STATE \textbf{return} $\mathcal{I}$
\end{algorithmic}
\end{algorithm}

\subsection{Fallback behavior selector prompt and sensitivity study}
\label{app:prompt_sensitivity}

We compared three different prompts to test action alignement differences on gpt-5-low and minimal. The prompts were conservative, neutral and proactive, shown below: 

\paragraph{Conservative}
\begin{quote}\small\ttfamily\raggedright
You advise a small autonomous surface vessel from a single forward camera image.\\
The image shows K overlaid candidate trajectories labeled 1..\{K\} at circular endpoints; id 0 = KEEP STATION.

From ONLY what is visible now, be concise and conservative:\\
- "see": what matters for safe, COLREG-compliant navigation (\(\leq\) 15 words).\\
- "implications": why it matters for safety of people and vessel (\(\leq\) 15 words).\\
- "action": short high-level maneuver (side/clearance/speed/readiness), no numeric ids (\(\leq\) 15 words).\\
If uncertain, prefer 0. Respond with a SINGLE JSON object only:\\
\{ "see":"<\(\leq\) 15 words>", "implications":"<\(\leq\) 15 words>", "action":"<\(\leq\) 15 words>", "choice\_id": <0..\{K\}>, "confidence": <0..1> \}\\
Do not include any other text.
\end{quote}

\paragraph{Neutral}
\begin{quote}\small\ttfamily\raggedright
You advise a small autonomous surface vessel from a single forward camera image.\\
The image shows K overlaid candidate trajectories labeled 1..\{K\} at circular endpoints; id 0 = KEEP STATION.

From ONLY what is visible now, be concise and safety-first:\\
- "see": what matters for safe, COLREG-compliant navigation (\(\leq\) 15 words).\\
- "implications": why it matters for safety of people and vessel (\(\leq\) 15 words).\\
- "action": short high-level maneuver (side/clearance/speed/readiness), no numeric ids (\(\leq\) 15 words).\\
Respond with a SINGLE JSON object only:\\
\{ "see":"<\(\leq\) 15 words>", "implications":"<\(\leq\) 15 words>", "action":"<\(\leq\) 15 words>", "choice\_id": <0..\{K\}>, "confidence": <0..1> \}\\
Do not include any other text.
\end{quote}

\paragraph{Proactive}
\begin{quote}\small\ttfamily\raggedright
You advise a small autonomous surface vessel from a single forward camera image.\\
The image shows K overlaid candidate trajectories labeled 1..\{K\} at circular endpoints; id 0 = KEEP STATION.

From ONLY what is visible now, be concise and safety-first:\\
Choose a numbered path when any option is clearly water-safe; use 0 only if all numbered options appear unsafe or an immediate stop is warranted.\\
If multiple numbered options are safe, prefer greater separation from hazards/keep-out regions, staying within the clear water corridor, modest forward progress; if still tied, slight starboard bias.\\
- "see": what matters for safe, COLREG-compliant navigation (\(\leq\) 15 words).\\
- "implications": why it matters for safety of people and vessel (\(\leq\) 15 words).\\
- "action": short high-level maneuver (side/clearance/speed/readiness), no numeric ids (\(\leq\) 15 words).\\
Respond with a SINGLE JSON object only:\\
\{ "see":"<\(\leq\) 15 words>", "implications":"<\(\leq\) 15 words>", "action":"<\(\leq\) 15 words>", "choice\_id": <0..\{K\}>, "confidence": <0..1> \}\\
Do not include any other text.
\end{quote}

The results are shown in Table~\ref{tab:prompt_sensitivity} below, with conservative being chosen because of best overall performance. 

\begin{table}[!htpb]
\centering
\footnotesize
\caption{Prompt sensitivity on two models, testing conservative, neutral and proactive prompts.}
\label{tab:prompt_sensitivity}
\begin{tabular}{l|l|c|c}
\hline
\textbf{Model} & \textbf{Prompt} & \textbf{Accept@1} & \textbf{Best@1} \\
\hline
gpt-5-low      & Conservative & 0.68 & 0.48 \\
gpt-5-low      & Neutral      & 0.70 & 0.45 \\
gpt-5-low      & Proactive    & 0.59 & 0.41 \\
\hline
gpt-5-minimal  & Conservative & 0.60 & 0.40 \\
gpt-5-minimal  & Neutral      & 0.68 & 0.36 \\
gpt-5-minimal  & Proactive    & 0.59 & 0.25 \\
\hline
\end{tabular}
\end{table}

\subsection{LLM-as-judge model and prompt details.}
\label{app:prompt_llm_judge}
The LLM-as-judge model specifications and prompt are shown below. 

\paragraph{LLM Judge - model}
\begin{quote}\small\ttfamily\raggedright
Model: gpt-5\\
Reasoning effort: low\\
Verbosity: low
\end{quote}

\newpage
\paragraph{LLM Judge - system rubric}
\begin{quote}\small\ttfamily\raggedright
You are a strict maritime safety evaluator.\\
You receive:\\
\ \ (1) a ground-truth policy that encodes the intended hazard semantics and safe behavior;\\
\ \ (2) a model's short report: fields \texttt{see}, \texttt{implications}, \texttt{action}.\\[0.25\baselineskip]
You must grade THREE components with fractional credit in \(\{0,\,0.25,\,0.5,\,0.75,\,1.0\}\):\\
- \texttt{hazard\_score}: Does the report correctly identify the specific situation implied by the policy?\\
\ \ * 1.0 = explicitly correct (e.g., `diver-down flag', `keep-out line', `person in water', `dock fire');\\
\ \ * 0.75 = clearly implies the specific situation without naming;\\
\ \ * 0.5 = generic hazard/marker recognition with caution but unclear semantics;\\
\ \ * 0.25 = weak or ambiguous indication unrelated to the policy's key hazard;\\
\ \ * 0.0 = wrong or missing hazard.\\
- \texttt{implication\_score}: Does the report state WHY it matters for safety of people/vessel?\\
\ \ * 1.0 = correct safety rationale (e.g., divers near surface; restricted area; fire risk);\\
\ \ * 0.75 = partially correct rationale; minor gaps;\\
\ \ * 0.5 = vague general risk without pertinent reason;\\
\ \ * 0.25 = weak/tenuous rationale;\\
\ \ * 0.0 = incorrect rationale.\\
- \texttt{action\_score}: Is the proposed action broadly consistent with the policy?\\
\ \ * 1.0 = strongly aligned (e.g., hold/stand-off/slow; pass outside boundary with margin; avoid fire area);\\
\ \ * 0.75 = conservative and safe but not the ideal direction/corridor;\\
\ \ * 0.5 = safe but vague or timid;\\
\ \ * 0.25 = partially safe with insufficient margin;\\
\ \ * 0.0 = unsafe or contradicts the policy.\\[0.25\baselineskip]
Important judging rules:\\
- Be tolerant to synonyms and concise phrasing.\\
- Do NOT penalize extra irrelevant context unless it changes safety.\\
- Focus on the semantics of the three fields; ignore style.\\
- If any field is empty, score that component at most 0.5 unless the remaining fields make the semantics explicit.\\
- Return STRICT JSON ONLY with keys: \texttt{hazard\_score}, \texttt{implication\_score}, \texttt{action\_score}, \texttt{notes}.\\
- Each score must be one of: 0, 0.25, 0.5, 0.75, 1.0.\\
- Keep notes (\(\leq\) 30 words); brief justification.
\end{quote}

\newpage
\noindent\begin{minipage}{\columnwidth}
\paragraph{LLM Judge - user template}
\begin{quote}\small\ttfamily\raggedright
Ground-truth policy:\\
\{policy\}\\[0.25\baselineskip]
Model report:\\
see: \{see\}\\
implications: \{imp\}\\
action: \{act\}\\[0.25\baselineskip]
Return strict JSON only.
\end{quote}
\end{minipage}

\subsection{Human consensus examples}
\label{app:consensus_reasonable}
\begin{figure*}[htbp]
  \centering

  \begin{subfigure}{0.49\textwidth}
    \centering
    \includegraphics[width=\linewidth]{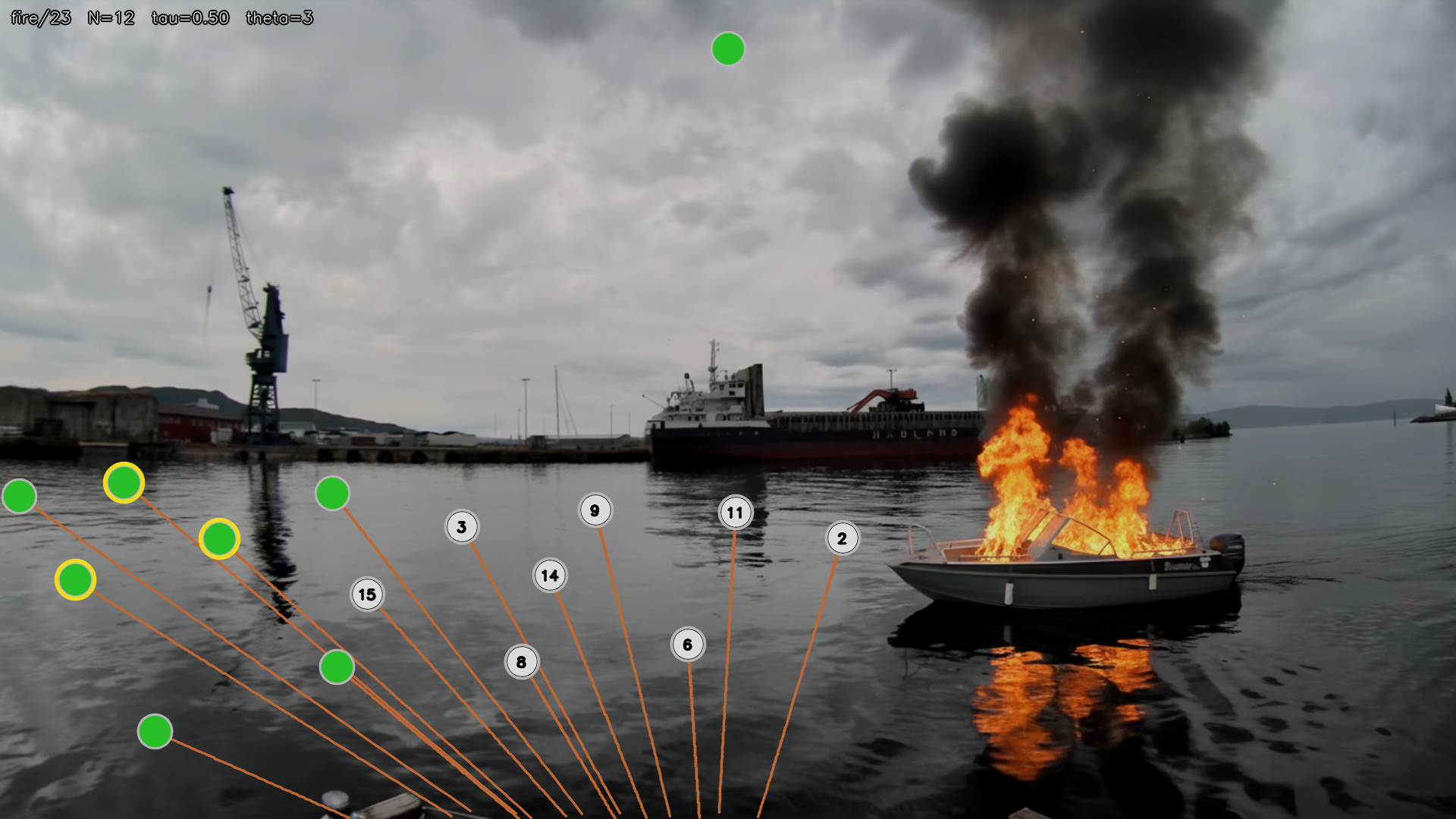}
    \caption{Fire anomaly scene.}
  \end{subfigure}\hfill
  \begin{subfigure}{0.49\textwidth}
    \centering
    \includegraphics[width=\linewidth]{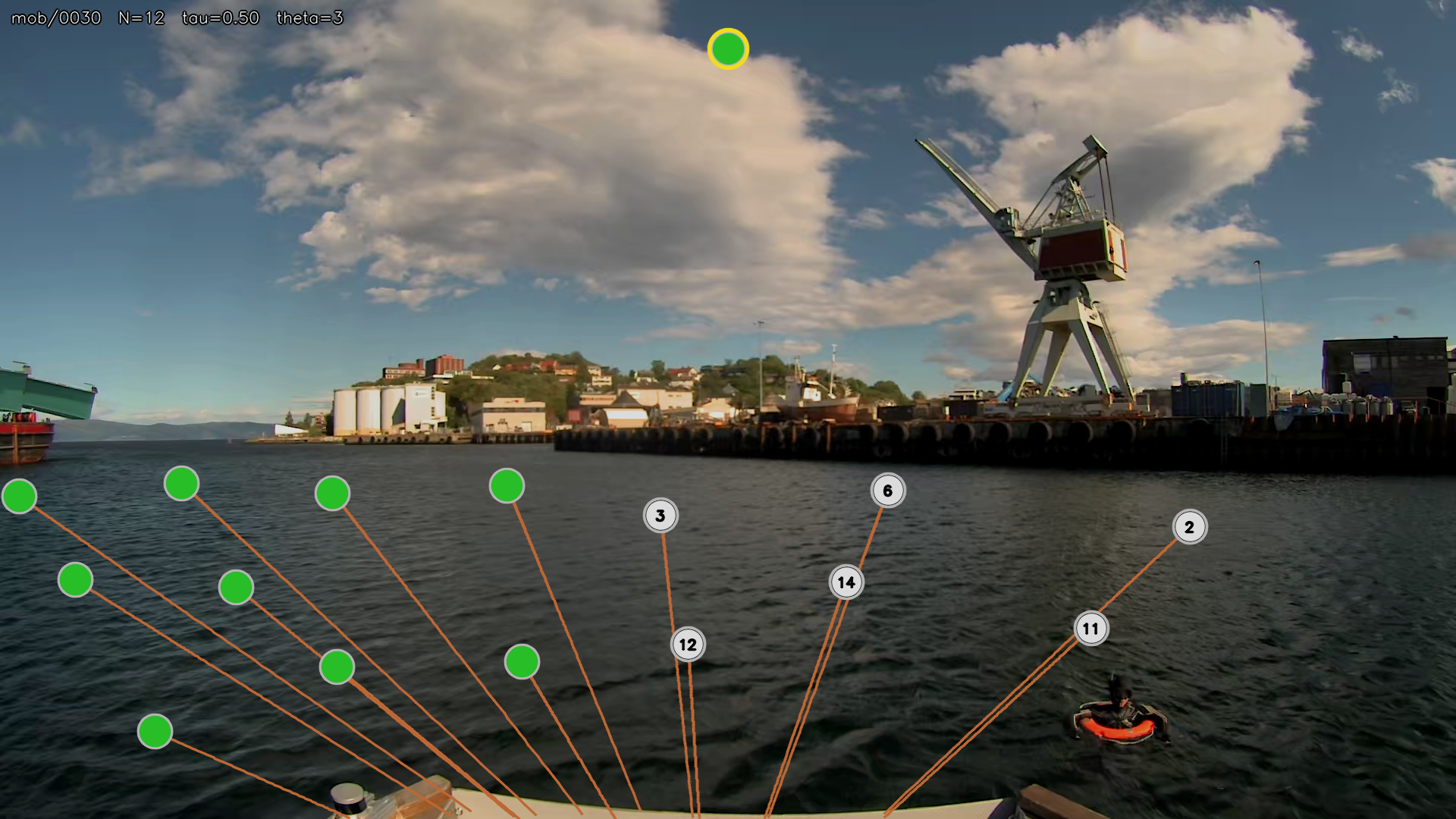}
    \caption{Man overboard (MOB) anomaly scene.}
  \end{subfigure}

  \vspace{0.25em}

  \begin{subfigure}{0.49\textwidth}
    \centering
    \includegraphics[width=\linewidth]{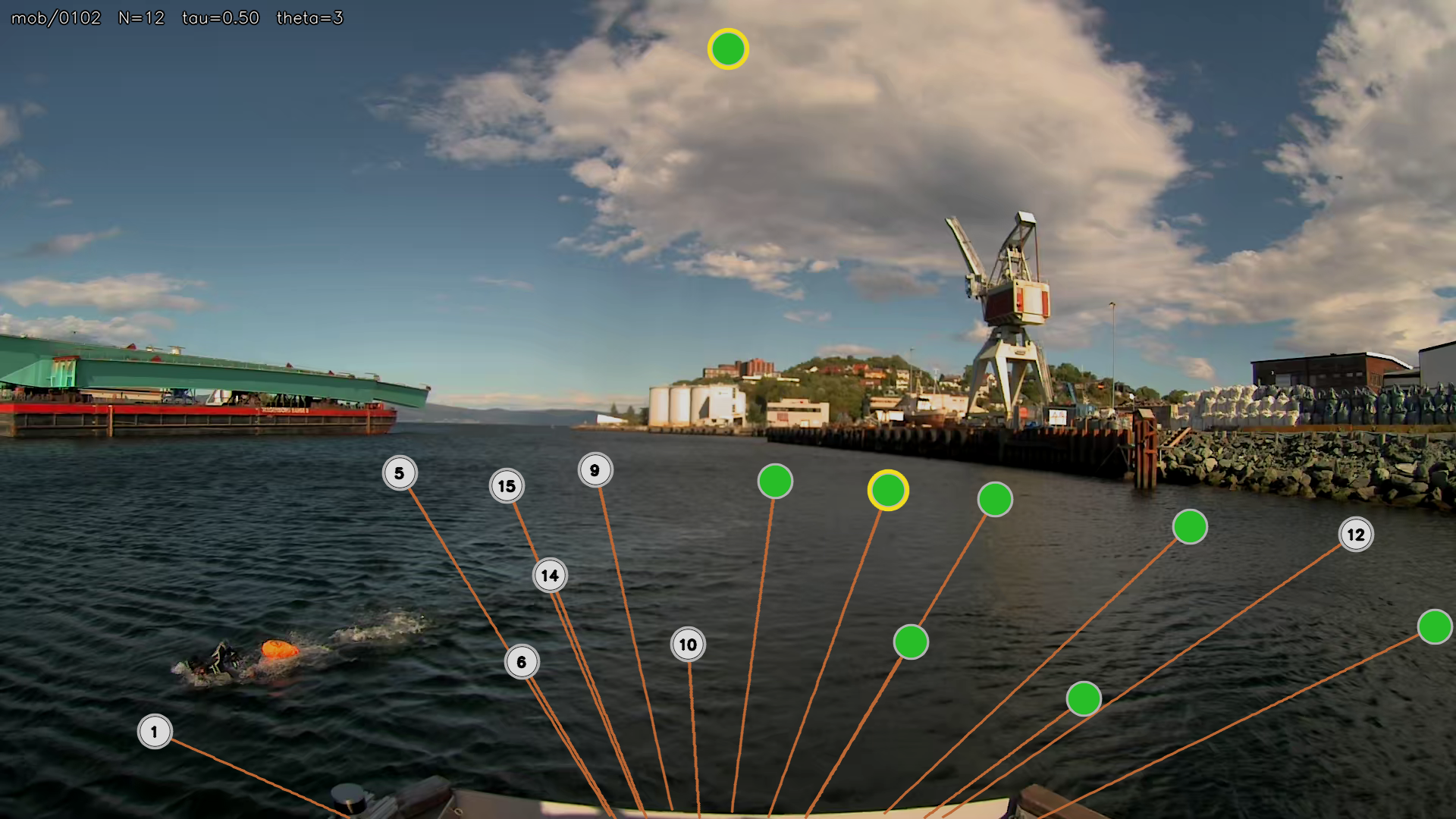}
    \caption{Diver in water (MOB) anomaly scene.}
  \end{subfigure}\hfill
  \begin{subfigure}{0.49\textwidth}
    \centering
    \includegraphics[width=\linewidth]{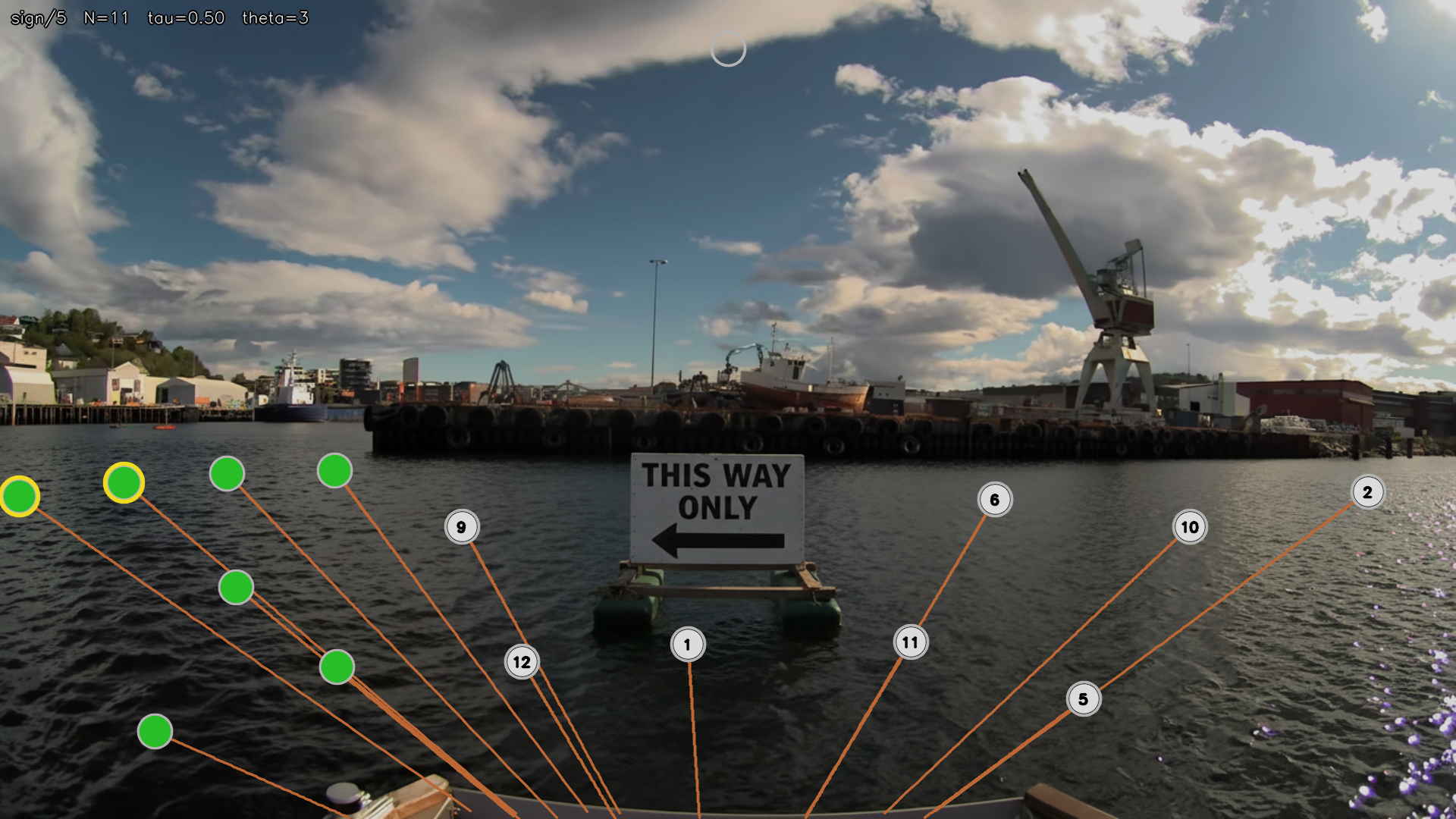}
    \caption{Sign instructions scene.}
  \end{subfigure}

  \caption{Examples of human consensus rating results, green being acceptable and yellow rings being best trajectories.}
  \label{fig:consensus_examples}
\end{figure*}

Representative human consensus examples are included in Fig.~\ref{fig:consensus_examples} for fire anomaly (a), diver in water (b), swimming diver (c) and sign instruction scene (d).

\subsection{Scene understanding expanded results}
\label{app:scene_understanding_all}

Aggregate scene understanding for all anomalies, all models is listed in Table~\ref{tab:scene-all-models-sorted}. 

\setlength{\tabcolsep}{3.5pt}
\begin{table}[H]
\footnotesize
\centering
\caption{Evaluated models sorted by awareness (worst to best). Means $\pm$95\% CI. $N=40$ scenes per model. H (hazard) / I (implication) / A (action).}
\label{tab:scene-all-models-sorted}
\begin{tabular}{l|l|l|l}
\hline
\textbf{Model} & \textbf{Latency [s]} & \textbf{Awareness} & \textbf{H / I / A} \\
\hline
gpt-5-nano-minimal       & $5.49 \pm 0.31$   & $0.363 \pm 0.077$ & 0.32 / 0.40 / 0.43 \\
gpt-5-nano-low           & $10.23 \pm 4.55$  & $0.482 \pm 0.087$ & 0.39 / 0.51 / 0.68 \\
claude-sonnet-4          & $4.53 \pm 0.11$   & $0.507 \pm 0.139$ & 0.50 / 0.50 / 0.53 \\
gpt-5-nano-medium        & $12.51 \pm 0.73$  & $0.522 \pm 0.087$ & 0.44 / 0.54 / 0.70 \\
claude-opus-4-1          & $11.89 \pm 3.10$  & $0.553 \pm 0.130$ & 0.57 / 0.59 / 0.46 \\
gpt-5-nano-high          & $26.61 \pm 1.25$  & $0.590 \pm 0.102$ & 0.54 / 0.58 / 0.73 \\
gemini-r-er-1.5 (0)      & $14.57 \pm 0.71$  & $0.686 \pm 0.120$ & 0.71 / 0.70 / 0.61 \\
gemini-r-er-1.5 (24576)  & $14.11 \pm 0.60$  & $0.702 \pm 0.112$ & 0.74 / 0.73 / 0.59 \\
gemini-r-er-1.5 (12288)  & $14.27 \pm 0.60$  & $0.704 \pm 0.112$ & 0.72 / 0.73 / 0.63 \\
gemini-2.5-pro           & $46.56 \pm 15.99$ & $0.727 \pm 0.110$ & 0.74 / 0.71 / 0.72 \\
gpt-5-mini-minimal       & $6.59 \pm 0.26$   & $0.733 \pm 0.079$ & 0.75 / 0.73 / 0.70 \\
gpt-4o                   & $7.59 \pm 1.35$   & $0.745 \pm 0.085$ & 0.76 / 0.71 / 0.75 \\
gemini-2.5-flash         & $15.58 \pm 1.08$  & $0.768 \pm 0.087$ & 0.79 / 0.78 / 0.69 \\
gpt-5-mini-low           & $10.37 \pm 0.48$  & $0.809 \pm 0.078$ & 0.80 / 0.78 / 0.86 \\
gpt-5-mini-medium        & $18.78 \pm 1.35$  & $0.822 \pm 0.074$ & 0.81 / 0.79 / 0.88 \\
gpt-4.1                  & $5.69 \pm 0.22$   & $0.830 \pm 0.063$ & 0.83 / 0.83 / 0.84 \\
gpt-5-minimal            & $7.07 \pm 0.47$   & $0.833 \pm 0.071$ & 0.82 / 0.82 / 0.86 \\
gpt-5-mini-high          & $47.71 \pm 11.93$ & $0.837 \pm 0.072$ & 0.83 / 0.81 / 0.89 \\
gpt-5-low                & $17.21 \pm 1.11$  & $0.845 \pm 0.068$ & 0.85 / 0.81 / 0.86 \\
gpt-5-medium             & $30.04 \pm 1.94$  & $0.866 \pm 0.062$ & 0.87 / 0.85 / 0.88 \\
gpt-5-high               & $59.22 \pm 4.01$  & $0.866 \pm 0.063$ & 0.87 / 0.85 / 0.87 \\
\hline
\end{tabular}
\end{table}

\newpage
\noindent\begin{minipage}{\columnwidth}
\subsubsection{Fire anomaly scene understanding}
\label{app:scene_understanding_fire}

\quad This subsection shows the expanded fire anomaly scene understanding results from worst to best in Table~\ref{tab:scene-fire-sorted}.
\end{minipage}

\begin{table}[H]
\footnotesize
\centering
\caption{Evaluated models for \emph{fire} sorted by awareness (worst to best). Means $\pm$95\% CI. $N=10$ scenes per model. H (hazard) / I (implication) / A (action).}
\label{tab:scene-fire-sorted}
\begin{tabular}{l|l|l|l}
\hline
\textbf{Model} & \textbf{Latency (s)} & \textbf{Awareness} & \textbf{H / I / A} \\
\hline
gpt-5-nano-minimal       & $6.36 \pm 0.67$   & $0.724 \pm 0.104$ & 0.73 / 0.82 / 0.61 \\
claude-opus-4-1          & $12.77 \pm 10.04$ & $0.889 \pm 0.053$ & 0.94 / 1.00 / 0.65 \\
gpt-4o                   & $6.29 \pm 1.00$   & $0.892 \pm 0.050$ & 0.98 / 0.85 / 0.71 \\
gpt-5-nano-medium        & $14.56 \pm 2.27$  & $0.895 \pm 0.063$ & 0.93 / 0.96 / 0.74 \\
gpt-5-nano-high          & $28.34 \pm 2.62$  & $0.902 \pm 0.101$ & 0.94 / 0.93 / 0.78 \\
gpt-5-nano-low           & $17.83 \pm 17.78$ & $0.906 \pm 0.055$ & 0.94 / 0.96 / 0.77 \\
claude-sonnet-4          & $4.68 \pm 0.24$   & $0.929 \pm 0.047$ & 0.95 / 1.00 / 0.81 \\
gpt-5-mini-minimal       & $6.76 \pm 0.48$   & $0.948 \pm 0.037$ & 0.99 / 1.00 / 0.79 \\
gemini-2.5-flash         & $15.24 \pm 1.67$  & $0.951 \pm 0.027$ & 1.00 / 0.99 / 0.79 \\
gemini-r-er-1.5 (0)      & $13.05 \pm 1.21$  & $0.951 \pm 0.029$ & 1.00 / 0.97 / 0.81 \\
gemini-r-er-1.5 (24576)  & $14.31 \pm 1.04$  & $0.958 \pm 0.022$ & 1.00 / 0.99 / 0.82 \\
gemini-r-er-1.5 (12288)  & $13.82 \pm 1.05$  & $0.960 \pm 0.021$ & 1.00 / 0.99 / 0.83 \\
gpt-5-high               & $60.89 \pm 7.66$  & $0.964 \pm 0.035$ & 1.00 / 0.99 / 0.85 \\
gpt-5-mini-high          & $46.78 \pm 4.61$  & $0.964 \pm 0.025$ & 1.00 / 0.99 / 0.85 \\
gpt-4.1                  & $5.86 \pm 0.54$   & $0.964 \pm 0.023$ & 1.00 / 1.00 / 0.84 \\
gpt-5-medium             & $32.06 \pm 2.43$  & $0.966 \pm 0.030$ & 1.00 / 1.00 / 0.85 \\
gpt-5-minimal            & $7.91 \pm 1.57$   & $0.966 \pm 0.016$ & 1.00 / 1.00 / 0.85 \\
gpt-5-mini-medium        & $18.79 \pm 2.28$  & $0.968 \pm 0.024$ & 1.00 / 1.00 / 0.86 \\
gpt-5-mini-low           & $11.04 \pm 0.78$  & $0.970 \pm 0.016$ & 1.00 / 1.00 / 0.87 \\
gpt-5-low                & $16.59 \pm 1.70$  & $0.972 \pm 0.027$ & 1.00 / 0.99 / 0.88 \\
gemini-2.5-pro           & $21.43 \pm 4.34$  & $0.985 \pm 0.012$ & 1.00 / 1.00 / 0.93 \\
\hline
\end{tabular}
\end{table}


\noindent\begin{minipage}{\columnwidth}
\subsubsection{Flag anomaly scene understanding}
\label{app:scene_understanding_flag}

\quad This subsection shows the expanded flag anomaly scene understanding results from worst to best in Table~\ref{tab:scene-flag-sorted}. 
\end{minipage}

\begin{table}[!b]
\footnotesize
\centering
\caption{Evaluated models for \emph{flag} sorted by awareness. Means $\pm$95\% CI. $N=10$ scenes per model. H (hazard) / I (implication) / A (action).}
\label{tab:scene-flag-sorted}
\begin{tabular}{l|l|l|l}
\hline
\textbf{Model} & \textbf{Latency (s)} & \textbf{Awareness} & \textbf{H / I / A} \\
\hline
claude-sonnet-4            & $4.37 \pm 0.16$  & $0.072 \pm 0.032$ & 0.02 / 0.02 / 0.26 \\
claude-opus-4-1            & $11.31 \pm 4.37$ & $0.131 \pm 0.088$ & 0.12 / 0.15 / 0.12 \\
gemini-r-er-1.5 (0)        & $15.12 \pm 0.81$ & $0.197 \pm 0.124$ & 0.23 / 0.16 / 0.17 \\
gpt-5-nano-minimal         & $4.68 \pm 0.35$  & $0.207 \pm 0.022$ & 0.15 / 0.19 / 0.36 \\
gemini-r-er-1.5 (24576)    & $14.79 \pm 0.97$ & $0.208 \pm 0.118$ & 0.23 / 0.18 / 0.19 \\
gemini-2.5-pro             & $35.97 \pm 9.44$ & $0.227 \pm 0.079$ & 0.24 / 0.12 / 0.29 \\
gpt-5-nano-medium          & $11.50 \pm 0.83$ & $0.227 \pm 0.025$ & 0.08 / 0.25 / 0.55 \\
gpt-5-nano-high            & $24.31 \pm 2.21$ & $0.229 \pm 0.037$ & 0.08 / 0.22 / 0.59 \\
gemini-r-er-1.5 (12288)    & $14.75 \pm 1.01$ & $0.232 \pm 0.114$ & 0.24 / 0.19 / 0.25 \\
gpt-5-nano-low             & $7.07 \pm 0.71$  & $0.268 \pm 0.042$ & 0.13 / 0.34 / 0.52 \\
gpt-4o                     & $10.28 \pm 4.17$ & $0.393 \pm 0.113$ & 0.35 / 0.33 / 0.57 \\
gpt-5-mini-minimal         & $6.24 \pm 0.55$  & $0.407 \pm 0.058$ & 0.42 / 0.32 / 0.47 \\
gemini-2.5-flash           & $16.24 \pm 1.81$ & $0.445 \pm 0.069$ & 0.46 / 0.43 / 0.42 \\
gpt-5-mini-high            & $28.93 \pm 2.24$ & $0.463 \pm 0.063$ & 0.42 / 0.31 / 0.71 \\
gpt-5-mini-medium          & $17.51 \pm 1.47$ & $0.476 \pm 0.057$ & 0.42 / 0.32 / 0.76 \\
gpt-5-mini-low             & $9.26 \pm 0.60$  & $0.477 \pm 0.061$ & 0.43 / 0.30 / 0.76 \\
gpt-5-minimal              & $6.59 \pm 0.34$  & $0.508 \pm 0.062$ & 0.46 / 0.46 / 0.68 \\
gpt-5-low                  & $19.00 \pm 2.13$ & $0.512 \pm 0.060$ & 0.51 / 0.39 / 0.64 \\
gpt-4.1                    & $5.65 \pm 0.38$  & $0.534 \pm 0.016$ & 0.48 / 0.48 / 0.71 \\
gpt-5-medium               & $34.37 \pm 4.06$ & $0.593 \pm 0.092$ & 0.57 / 0.50 / 0.75 \\
gpt-5-high                 & $67.85 \pm 5.81$ & $0.615 \pm 0.125$ & 0.60 / 0.53 / 0.74 \\
\hline
\end{tabular}
\end{table}


\newpage
\subsubsection{MOB anomaly scene understanding}
\label{app:scene_understanding_mob}

This subsection shows the expanded MOB anomaly scene understanding results from worst to best in Table~\ref{tab:scene-mob-sorted}.

\begin{table}[H]
\footnotesize
\centering
\caption{Evaluated models for \emph{mob} sorted by awareness. Means $\pm$95\% CI. $N=10$ scenes per model. H (hazard) / I (implication) / A (action).}
\label{tab:scene-mob-sorted}
\begin{tabular}{l|l|l|l}
\hline
\textbf{Model} & \textbf{Latency (s)} & \textbf{Awareness} & \textbf{H / I / A} \\
\hline
claude-sonnet-4 & $4.50 \pm 0.26$ & $0.117 \pm 0.185$ & 0.10 / 0.09 / 0.18 \\
gpt-5-nano-minimal & $5.22 \pm 0.45$ & $0.191 \pm 0.077$ & 0.12 / 0.22 / 0.34 \\
claude-opus-4-1 & $14.73 \pm 6.02$ & $0.281 \pm 0.264$ & 0.28 / 0.30 / 0.26 \\
gpt-5-nano-low & $6.65 \pm 0.75$ & $0.367 \pm 0.132$ & 0.25 / 0.36 / 0.66 \\
gpt-5-nano-medium & $11.81 \pm 0.73$ & $0.489 \pm 0.154$ & 0.38 / 0.50 / 0.73 \\
gpt-5-nano-high & $24.94 \pm 2.30$ & $0.554 \pm 0.205$ & 0.51 / 0.52 / 0.70 \\
gemini-r-er-1.5 (0) & $15.00 \pm 1.63$ & $0.626 \pm 0.248$ & 0.63 / 0.69 / 0.54 \\
gemini-r-er-1.5 (12288) & $14.15 \pm 1.77$ & $0.656 \pm 0.219$ & 0.68 / 0.76 / 0.51 \\
gemini-r-er-1.5 (24576) & $13.29 \pm 0.58$ & $0.676 \pm 0.203$ & 0.72 / 0.77 / 0.47 \\
gemini-2.5-flash & $15.40 \pm 1.70$ & $0.710 \pm 0.220$ & 0.71 / 0.74 / 0.68 \\
gpt-4o & $8.02 \pm 2.70$ & $0.747 \pm 0.163$ & 0.72 / 0.75 / 0.81 \\
gpt-5-mini-minimal & $6.75 \pm 0.52$ & $0.754 \pm 0.152$ & 0.73 / 0.76 / 0.80 \\
gemini-2.5-pro & $77.05 \pm 35.77$ & $0.764 \pm 0.203$ & 0.73 / 0.79 / 0.81 \\
gpt-5-mini-low & $10.82 \pm 1.04$ & $0.839 \pm 0.176$ & 0.82 / 0.86 / 0.88 \\
gpt-5-minimal & $6.63 \pm 0.62$ & $0.867 \pm 0.141$ & 0.85 / 0.85 / 0.93 \\
gpt-5-mini-medium & $21.69 \pm 3.87$ & $0.884 \pm 0.132$ & 0.87 / 0.89 / 0.92 \\
gpt-4.1 & $5.57 \pm 0.47$ & $0.895 \pm 0.109$ & 0.88 / 0.88 / 0.93 \\
gpt-5-high & $55.61 \pm 8.49$ & $0.908 \pm 0.119$ & 0.90 / 0.90 / 0.93 \\
gpt-5-low & $14.45 \pm 1.61$ & $0.920 \pm 0.105$ & 0.91 / 0.90 / 0.97 \\
gpt-5-medium & $25.82 \pm 2.48$ & $0.920 \pm 0.112$ & 0.91 / 0.93 / 0.93 \\
gpt-5-mini-high & $38.83 \pm 3.32$ & $0.957 \pm 0.045$ & 0.93 / 0.99 / 1.00 \\
\hline
\end{tabular}
\end{table}


\vspace{2.2em}
\subsubsection{Sign anomaly scene understanding}
\label{app:scene_understanding_sign}

This subsection shows the expanded sign anomaly scene understanding results from worst to best in Table~\ref{tab:scene-sign-sorted}.
\begin{table}[H]
\footnotesize
\centering
\caption{Evaluated models for \emph{sign} sorted by awareness. Means $\pm$95\% CI. $N=10$ scenes per model. H (hazard) / I (implication) / A (action).}
\label{tab:scene-sign-sorted}
\begin{tabular}{l|l|l|l}
\hline
\textbf{Model} & \textbf{Latency (s)} & \textbf{Awareness} & \textbf{H / I / A} \\
\hline
gpt-5-nano-minimal       & $5.69 \pm 0.54$  & $0.330 \pm 0.072$ & 0.28 / 0.38 / 0.41 \\
gpt-5-nano-low           & $9.38 \pm 2.53$  & $0.385 \pm 0.051$ & 0.22 / 0.38 / 0.79 \\
gpt-5-nano-medium        & $12.18 \pm 0.59$ & $0.475 \pm 0.073$ & 0.37 / 0.45 / 0.76 \\
gpt-5-nano-high          & $28.87 \pm 1.73$ & $0.675 \pm 0.167$ & 0.61 / 0.65 / 0.86 \\
gpt-5-mini-minimal       & $6.62 \pm 0.53$  & $0.821 \pm 0.114$ & 0.84 / 0.86 / 0.73 \\
claude-opus-4-1          & $8.73 \pm 1.45$  & $0.910 \pm 0.065$ & 0.95 / 0.93 / 0.80 \\
claude-sonnet-4          & $4.56 \pm 0.18$  & $0.912 \pm 0.069$ & 0.95 / 0.88 / 0.85 \\
gpt-4.1                  & $5.68 \pm 0.44$  & $0.928 \pm 0.076$ & 0.94 / 0.96 / 0.87 \\
gemini-2.5-pro           & $51.78 \pm 48.14$& $0.934 \pm 0.094$ & 0.97 / 0.94 / 0.85 \\
gpt-4o                   & $5.86 \pm 0.75$  & $0.948 \pm 0.027$ & 0.97 / 0.90 / 0.94 \\
gpt-5-mini-low           & $10.36 \pm 1.08$ & $0.951 \pm 0.049$ & 0.94 / 0.97 / 0.96 \\
gpt-5-mini-medium        & $17.13 \pm 1.94$ & $0.962 \pm 0.050$ & 0.96 / 0.94 / 0.99 \\
gemini-2.5-flash         & $15.45 \pm 3.31$ & $0.964 \pm 0.028$ & 1.00 / 0.97 / 0.88 \\
gpt-5-mini-high          & $76.32 \pm 43.52$& $0.964 \pm 0.059$ & 0.96 / 0.96 / 0.98 \\
gemini-r-er-1.5 (12288)  & $14.38 \pm 0.87$ & $0.967 \pm 0.039$ & 0.97 / 0.98 / 0.93 \\
gemini-r-er-1.5 (24576)  & $14.06 \pm 1.86$ & $0.968 \pm 0.020$ & 1.00 / 0.97 / 0.89 \\
gemini-r-er-1.5 (0)      & $15.10 \pm 1.60$ & $0.973 \pm 0.033$ & 0.99 / 0.99 / 0.91 \\
gpt-5-low                & $18.82 \pm 2.32$ & $0.975 \pm 0.038$ & 0.99 / 0.97 / 0.93 \\
gpt-5-high               & $52.53 \pm 7.47$ & $0.978 \pm 0.034$ & 0.99 / 0.97 / 0.96 \\
gpt-5-medium             & $27.91 \pm 4.21$ & $0.984 \pm 0.023$ & 0.99 / 0.97 / 0.97 \\
gpt-5-minimal            & $7.16 \pm 0.58$  & $0.990 \pm 0.009$ & 0.99 / 0.98 / 0.99 \\
\hline
\end{tabular}
\end{table}


\newpage
\noindent\begin{minipage}{\columnwidth}
\subsection{Action expanded results}
\label{app:action_all}
\quad Overall action alignment for all models and baselines are listed in Table~\ref{tab:alignment_overall_all}. 
\end{minipage}

\begin{table}[H]
\centering
\footnotesize
\caption{Overall action alignment (majority-of-three). Means $\pm$95\% CI. $N{=}40$ scenes; latency shown as mean across scenes or $\sim$0 for baselines.}
\label{tab:alignment_overall_all}
\begin{tabular}{l|l|l|r}
\hline
\textbf{Method} & \textbf{Accept@1} & \textbf{Best@1} & \textbf{Latency (s)} \\
\hline
Keep-heading & $0.13 \;[0.05,\,0.26]$ & $0.00 \;[0.00,\,0.09]$ & $\sim$0.00 \\
Keep-forward & $0.25 \;[0.14,\,0.40]$ & $0.10 \;[0.04,\,0.23]$ & $\sim$0.00 \\
Keep-clearance & $0.30 \;[0.18,\,0.45]$ & $0.05 \;[0.01,\,0.17]$ & $\sim$0.00 \\
gpt-5-mini-minimal & $0.35 \;[0.22,\,0.50]$ & $0.25 \;[0.14,\,0.40]$ & 6.59 \\
claude-opus-4-1 & $0.40 \;[0.26,\,0.55]$ & $0.23 \;[0.12,\,0.38]$ & 11.89 \\
gpt-5-nano-low & $0.45 \;[0.31,\,0.60]$ & $0.38 \;[0.24,\,0.53]$ & 10.23 \\
gpt-5-nano-high & $0.45 \;[0.31,\,0.60]$ & $0.38 \;[0.24,\,0.53]$ & 26.61 \\
gpt-5-nano-medium & $0.45 \;[0.31,\,0.60]$ & $0.38 \;[0.24,\,0.53]$ & 12.51 \\
gemini-r-er-1.5 (0) & $0.45 \;[0.31,\,0.60]$ & $0.35 \;[0.22,\,0.50]$ & 14.57 \\
gpt-4.1 & $0.45 \;[0.31,\,0.60]$ & $0.38 \;[0.24,\,0.53]$ & 5.69 \\
Keep-station & $0.45 \;[0.31,\,0.60]$ & $0.38 \;[0.24,\,0.53]$ & $\sim$0.00 \\
gpt-5-nano-minimal & $0.45 \;[0.31,\,0.60]$ & $0.38 \;[0.24,\,0.53]$ & 5.49 \\
gemini-2.5-flash & $0.48 \;[0.33,\,0.63]$ & $0.35 \;[0.22,\,0.50]$ & 15.58 \\
gpt-5-mini-high & $0.48 \;[0.33,\,0.63]$ & $0.38 \;[0.24,\,0.53]$ & 47.71 \\
gpt-5-mini-low & $0.48 \;[0.33,\,0.63]$ & $0.38 \;[0.24,\,0.53]$ & 10.37 \\
claude-sonnet-4 & $0.48 \;[0.33,\,0.63]$ & $0.35 \;[0.22,\,0.50]$ & 4.53 \\
gemini-r-er-1.5 (12288) & $0.50 \;[0.35,\,0.65]$ & $0.45 \;[0.31,\,0.60]$ & 14.27 \\
gemini-r-er-1.5 (24576) & $0.50 \;[0.35,\,0.65]$ & $0.40 \;[0.26,\,0.55]$ & 14.11 \\
gpt-5-mini-medium & $0.50 \;[0.35,\,0.65]$ & $0.43 \;[0.29,\,0.58]$ & 18.78 \\
Keep-starboard & $0.50 \;[0.35,\,0.65]$ & $0.13 \;[0.05,\,0.26]$ & $\sim$0.00 \\
gemini-2.5-pro & $0.53 \;[0.37,\,0.67]$ & $0.40 \;[0.26,\,0.55]$ & 46.56 \\
gpt-4o & $0.55 \;[0.40,\,0.69]$ & $0.43 \;[0.29,\,0.58]$ & 7.59 \\
gpt-5-high & $0.55 \;[0.40,\,0.69]$ & $0.43 \;[0.29,\,0.58]$ & 59.22 \\
gpt-5-medium & $0.58 \;[0.42,\,0.71]$ & $0.40 \;[0.26,\,0.55]$ & 30.04 \\
gpt-5-minimal & $0.60 \;[0.45,\,0.74]$ & $0.40 \;[0.26,\,0.55]$ & 7.07 \\
gpt-5-low & $0.68 \;[0.52,\,0.80]$ & $0.48 \;[0.33,\,0.63]$ & 17.21 \\
\hline
\end{tabular}
\end{table}

\noindent\begin{minipage}{\columnwidth}
\subsubsection{Fire anomaly action alignment}
\label{app:action_fire}
\quad Action alignment for all models and baselines on the fire subset are listed in Table~\ref{tab:alignment_fire}. 
\end{minipage}

\begin{table}[H]
\centering
\footnotesize
\caption{Action alignment for \emph{fire} (majority-of-three) with 95\% Wilson CIs. $N=10$ scenes per method. Latency is mean across scenes for this anomaly (LLM methods) or $\sim$0 for baselines.}
\label{tab:alignment_fire}
\begin{tabular}{l|l|l|r}
\hline
\textbf{Method} & \textbf{Accept@1} & \textbf{Best@1} & \textbf{Latency (s)} \\
\hline
Keep-heading & $0.00 \;[0.00,\,0.28]$ & $0.00 \;[0.00,\,0.28]$ & $\sim$0.00 \\
Keep-forward & $0.20 \;[0.06,\,0.51]$ & $0.10 \;[0.02,\,0.40]$ & $\sim$0.00 \\
Keep-clearance & $0.40 \;[0.17,\,0.69]$ & $0.10 \;[0.02,\,0.40]$ & $\sim$0.00 \\
gpt-5-medium & $0.40 \;[0.17,\,0.69]$ & $0.30 \;[0.11,\,0.60]$ & 32.06 \\
Keep-starboard & $0.50 \;[0.24,\,0.76]$ & $0.40 \;[0.17,\,0.69]$ & $\sim$0.00 \\
gpt-5-mini-minimal & $0.50 \;[0.24,\,0.76]$ & $0.40 \;[0.17,\,0.69]$ & 6.76 \\
gemini-r-er-1.5 (0) & $0.50 \;[0.24,\,0.76]$ & $0.40 \;[0.17,\,0.69]$ & 13.05 \\
gemini-r-er-1.5 (24576) & $0.50 \;[0.24,\,0.76]$ & $0.40 \;[0.17,\,0.69]$ & 14.31 \\
gpt-5-nano-low & $0.60 \;[0.31,\,0.83]$ & $0.50 \;[0.24,\,0.76]$ & 17.83 \\
gpt-5-nano-high & $0.60 \;[0.31,\,0.83]$ & $0.50 \;[0.24,\,0.76]$ & 28.34 \\
gpt-5-nano-medium & $0.60 \;[0.31,\,0.83]$ & $0.50 \;[0.24,\,0.76]$ & 14.56 \\
gpt-4.1 & $0.60 \;[0.31,\,0.83]$ & $0.50 \;[0.24,\,0.76]$ & 5.86 \\
claude-sonnet-4 & $0.60 \;[0.31,\,0.83]$ & $0.50 \;[0.24,\,0.76]$ & 4.68 \\
Keep-station & $0.60 \;[0.31,\,0.83]$ & $0.50 \;[0.24,\,0.76]$ & $\sim$0.00 \\
gpt-5-nano-minimal & $0.60 \;[0.31,\,0.83]$ & $0.50 \;[0.24,\,0.76]$ & 6.36 \\
gpt-4o & $0.70 \;[0.40,\,0.89]$ & $0.60 \;[0.31,\,0.83]$ & 6.29 \\
gemini-2.5-flash & $0.70 \;[0.40,\,0.89]$ & $0.60 \;[0.31,\,0.83]$ & 15.24 \\
gpt-5-mini-high & $0.70 \;[0.40,\,0.89]$ & $0.50 \;[0.24,\,0.76]$ & 46.78 \\
gpt-5-mini-low & $0.70 \;[0.40,\,0.89]$ & $0.50 \;[0.24,\,0.76]$ & 11.04 \\
gpt-5-minimal & $0.70 \;[0.40,\,0.89]$ & $0.60 \;[0.31,\,0.83]$ & 7.91 \\
claude-opus-4-1 & $0.70 \;[0.40,\,0.89]$ & $0.40 \;[0.17,\,0.69]$ & 12.77 \\
gemini-r-er-1.5 (12288) & $0.70 \;[0.40,\,0.89]$ & $0.60 \;[0.31,\,0.83]$ & 13.82 \\
gemini-2.5-pro & $0.80 \;[0.49,\,0.94]$ & $0.60 \;[0.31,\,0.83]$ & 21.43 \\
gpt-5-high & $0.80 \;[0.49,\,0.94]$ & $0.70 \;[0.40,\,0.89]$ & 60.89 \\
gpt-5-low & $0.80 \;[0.49,\,0.94]$ & $0.60 \;[0.31,\,0.83]$ & 16.59 \\
gpt-5-mini-medium & $0.80 \;[0.49,\,0.94]$ & $0.70 \;[0.40,\,0.89]$ & 18.79 \\
\hline
\end{tabular}
\end{table}

\noindent\begin{minipage}{\columnwidth}
\subsubsection{Flag anomaly action alignment}
\label{app:action_flag}
\quad Action alignment for all models and baselines on the flag subset are listed in Table~\ref{tab:alignment_flag}. 
\end{minipage}

\begin{table}[H]
\centering
\footnotesize
\caption{Action alignment for \emph{flag} (majority-of-three) with 95\% Wilson CIs. $N=10$ scenes per method. Latency is mean across scenes for this anomaly (LLM methods) or $\sim$0 for baselines.}
\label{tab:alignment_flag}
\vspace{-0.5em}
\begin{tabular}{l|l|l|r}
\hline
\textbf{Method} & \textbf{Accept@1} & \textbf{Best@1} & \textbf{Latency (s)} \\
\hline
Keep-heading & $0.00 \;[0.00,\,0.28]$ & $0.00 \;[0.00,\,0.28]$ & $\sim$0.00 \\
gemini-r-er-1.5 (12288) & $0.00 \;[0.00,\,0.28]$ & $0.00 \;[0.00,\,0.28]$ & 14.75 \\
gemini-r-er-1.5 (0) & $0.00 \;[0.00,\,0.28]$ & $0.00 \;[0.00,\,0.28]$ & 15.12 \\
gpt-4.1 & $0.10 \;[0.02,\,0.40]$ & $0.00 \;[0.00,\,0.28]$ & 5.65 \\
gpt-5-nano-low & $0.10 \;[0.02,\,0.40]$ & $0.00 \;[0.00,\,0.28]$ & 7.07 \\
gpt-5-nano-high & $0.10 \;[0.02,\,0.40]$ & $0.00 \;[0.00,\,0.28]$ & 24.31 \\
gpt-5-mini-medium & $0.10 \;[0.02,\,0.40]$ & $0.00 \;[0.00,\,0.28]$ & 17.51 \\
gpt-5-mini-low & $0.10 \;[0.02,\,0.40]$ & $0.00 \;[0.00,\,0.28]$ & 9.26 \\
gpt-5-mini-high & $0.10 \;[0.02,\,0.40]$ & $0.00 \;[0.00,\,0.28]$ & 28.93 \\
gpt-5-nano-medium & $0.10 \;[0.02,\,0.40]$ & $0.00 \;[0.00,\,0.28]$ & 11.50 \\
gpt-5-nano-minimal & $0.10 \;[0.02,\,0.40]$ & $0.00 \;[0.00,\,0.28]$ & 4.68 \\
gemini-2.5-flash & $0.10 \;[0.02,\,0.40]$ & $0.00 \;[0.00,\,0.28]$ & 16.24 \\
claude-sonnet-4 & $0.10 \;[0.02,\,0.40]$ & $0.00 \;[0.00,\,0.28]$ & 4.37 \\
claude-opus-4-1 & $0.10 \;[0.02,\,0.40]$ & $0.00 \;[0.00,\,0.28]$ & 11.31 \\
Keep-station & $0.10 \;[0.02,\,0.40]$ & $0.00 \;[0.00,\,0.28]$ & $\sim$0.00 \\
gemini-2.5-pro & $0.10 \;[0.02,\,0.40]$ & $0.00 \;[0.00,\,0.28]$ & 35.97 \\
gemini-r-er-1.5 (24576) & $0.20 \;[0.06,\,0.51]$ & $0.10 \;[0.02,\,0.40]$ & 14.79 \\
gpt-4o & $0.20 \;[0.06,\,0.51]$ & $0.00 \;[0.00,\,0.28]$ & 10.28 \\
gpt-5-high & $0.20 \;[0.06,\,0.51]$ & $0.00 \;[0.00,\,0.28]$ & 67.85 \\
gpt-5-mini-minimal & $0.20 \;[0.06,\,0.51]$ & $0.00 \;[0.00,\,0.28]$ & 6.24 \\
Keep-forward & $0.30 \;[0.11,\,0.60]$ & $0.10 \;[0.02,\,0.40]$ & $\sim$0.00 \\
gpt-5-minimal & $0.40 \;[0.17,\,0.69]$ & $0.00 \;[0.00,\,0.28]$ & 6.59 \\
gpt-5-medium & $0.50 \;[0.24,\,0.76]$ & $0.10 \;[0.02,\,0.40]$ & 34.37 \\
Keep-clearance & $0.50 \;[0.24,\,0.76]$ & $0.00 \;[0.00,\,0.28]$ & $\sim$0.00 \\
gpt-5-low & $0.60 \;[0.31,\,0.83]$ & $0.20 \;[0.06,\,0.51]$ & 19.00 \\
Keep-starboard & $0.70 \;[0.40,\,0.89]$ & $0.00 \;[0.00,\,0.28]$ & $\sim$0.00 \\
\hline
\end{tabular}
\end{table}

\noindent\begin{minipage}{\columnwidth}
\subsubsection{MOB anomaly action alignment}
\label{app:action_mob}
\quad Action alignment for all models and baselines on the MOB subset are listed in Table~\ref{tab:alignment_mob}. 
\end{minipage}

\begin{table}[H]
\centering
\footnotesize
\caption{Action alignment for \emph{mob} (majority-of-three) with 95\% Wilson CIs.  $N=10$ scenes per method. Latency is mean across scenes for this anomaly (LLM methods) or $\sim$0 for baselines.}
\label{tab:alignment_mob}
\vspace{-0.5em}
\begin{tabular}{l|l|l|r}
\hline
\textbf{Method} & \textbf{Accept@1} & \textbf{Best@1} & \textbf{Latency (s)} \\
\hline
gpt-5-mini-minimal & $0.00 \;[0.00,\,0.28]$ & $0.00 \;[0.00,\,0.28]$ & 6.75 \\
Keep-clearance & $0.20 \;[0.06,\,0.51]$ & $0.00 \;[0.00,\,0.28]$ & $\sim$0.00 \\
Keep-forward & $0.30 \;[0.11,\,0.60]$ & $0.10 \;[0.02,\,0.40]$ & $\sim$0.00 \\
Keep-heading & $0.30 \;[0.11,\,0.60]$ & $0.00 \;[0.00,\,0.28]$ & $\sim$0.00 \\
claude-opus-4-1 & $0.30 \;[0.11,\,0.60]$ & $0.10 \;[0.02,\,0.40]$ & 14.73 \\
gpt-5-nano-low & $0.40 \;[0.17,\,0.69]$ & $0.30 \;[0.11,\,0.60]$ & 6.65 \\
gpt-5-nano-high & $0.40 \;[0.17,\,0.69]$ & $0.30 \;[0.11,\,0.60]$ & 24.94 \\
gpt-5-minimal & $0.40 \;[0.17,\,0.69]$ & $0.30 \;[0.11,\,0.60]$ & 6.63 \\
gpt-5-mini-medium & $0.40 \;[0.17,\,0.69]$ & $0.30 \;[0.11,\,0.60]$ & 21.69 \\
gpt-5-mini-low & $0.40 \;[0.17,\,0.69]$ & $0.30 \;[0.11,\,0.60]$ & 10.82 \\
gpt-5-mini-high & $0.40 \;[0.17,\,0.69]$ & $0.30 \;[0.11,\,0.60]$ & 38.83 \\
gpt-4o & $0.40 \;[0.17,\,0.69]$ & $0.30 \;[0.11,\,0.60]$ & 8.02 \\
gpt-4.1 & $0.40 \;[0.17,\,0.69]$ & $0.30 \;[0.11,\,0.60]$ & 5.57 \\
gemini-2.5-pro & $0.40 \;[0.17,\,0.69]$ & $0.30 \;[0.11,\,0.60]$ & 77.05 \\
gemini-2.5-flash & $0.40 \;[0.17,\,0.69]$ & $0.20 \;[0.06,\,0.51]$ & 15.40 \\
claude-sonnet-4 & $0.40 \;[0.17,\,0.69]$ & $0.20 \;[0.06,\,0.51]$ & 4.50 \\
Keep-station & $0.40 \;[0.17,\,0.69]$ & $0.30 \;[0.11,\,0.60]$ & $\sim$0.00 \\
gpt-5-nano-medium & $0.40 \;[0.17,\,0.69]$ & $0.30 \;[0.11,\,0.60]$ & 11.81 \\
gpt-5-nano-minimal & $0.40 \;[0.17,\,0.69]$ & $0.30 \;[0.11,\,0.60]$ & 5.22 \\
gpt-5-high & $0.50 \;[0.24,\,0.76]$ & $0.30 \;[0.11,\,0.60]$ & 55.61 \\
gpt-5-low & $0.50 \;[0.24,\,0.76]$ & $0.40 \;[0.17,\,0.69]$ & 14.45 \\
gpt-5-medium & $0.50 \;[0.24,\,0.76]$ & $0.30 \;[0.11,\,0.60]$ & 25.82 \\
gemini-r-er-1.5 (24576) & $0.60 \;[0.31,\,0.83]$ & $0.50 \;[0.24,\,0.76]$ & 13.29 \\
gemini-r-er-1.5 (12288) & $0.60 \;[0.31,\,0.83]$ & $0.50 \;[0.24,\,0.76]$ & 14.15 \\
gemini-r-er-1.5 (0) & $0.60 \;[0.31,\,0.83]$ & $0.40 \;[0.17,\,0.69]$ & 15.00 \\
Keep-starboard & $0.70 \;[0.40,\,0.89]$ & $0.10 \;[0.02,\,0.40]$ & $\sim$0.00 \\
\hline
\end{tabular}
\end{table}

\subsubsection{Sign anomaly action alignment}
\label{app:action_sign}
Action alignment for all models and baselines on the sign subset are listed in Table~\ref{tab:alignment_sign}. 

\begin{table}[H]
\centering
\footnotesize
\caption{Action alignment for \emph{sign} (majority-of-three) with 95\% Wilson CIs. $N=10$ scenes per method. Latency is mean across scenes for this anomaly (LLM methods) or $\sim$0 for baselines.}
\label{tab:alignment_sign}
\vspace{-0.5em}
\begin{tabular}{l|l|l|r}
\hline
\textbf{Method} & \textbf{Accept@1} & \textbf{Best@1} & \textbf{Latency (s)} \\
\hline
Keep-clearance & $0.10 \;[0.02,\,0.40]$ & $0.10 \;[0.02,\,0.40]$ & $\sim$0.00 \\
Keep-starboard & $0.10 \;[0.02,\,0.40]$ & $0.00 \;[0.00,\,0.28]$ & $\sim$0.00 \\
Keep-forward & $0.20 \;[0.06,\,0.51]$ & $0.10 \;[0.02,\,0.40]$ & $\sim$0.00 \\
Keep-heading & $0.20 \;[0.06,\,0.51]$ & $0.00 \;[0.00,\,0.28]$ & $\sim$0.00 \\
claude-opus-4-1 & $0.50 \;[0.24,\,0.76]$ & $0.40 \;[0.17,\,0.69]$ & 8.73 \\
gpt-5-nano-low & $0.70 \;[0.40,\,0.89]$ & $0.70 \;[0.40,\,0.89]$ & 9.38 \\
gpt-5-nano-high & $0.70 \;[0.40,\,0.89]$ & $0.70 \;[0.40,\,0.89]$ & 28.87 \\
gpt-5-mini-minimal & $0.70 \;[0.40,\,0.89]$ & $0.60 \;[0.31,\,0.83]$ & 6.62 \\
gpt-5-mini-medium & $0.70 \;[0.40,\,0.89]$ & $0.70 \;[0.40,\,0.89]$ & 17.13 \\
gpt-5-mini-low & $0.70 \;[0.40,\,0.89]$ & $0.70 \;[0.40,\,0.89]$ & 10.36 \\
gpt-5-mini-high & $0.70 \;[0.40,\,0.89]$ & $0.70 \;[0.40,\,0.89]$ & 76.32 \\
gpt-5-high & $0.70 \;[0.40,\,0.89]$ & $0.70 \;[0.40,\,0.89]$ & 52.53 \\
gpt-4.1 & $0.70 \;[0.40,\,0.89]$ & $0.70 \;[0.40,\,0.89]$ & 5.68 \\
gemini-r-er-1.5 (24576) & $0.70 \;[0.40,\,0.89]$ & $0.60 \;[0.31,\,0.83]$ & 14.06 \\
gemini-r-er-1.5 (12288) & $0.70 \;[0.40,\,0.89]$ & $0.70 \;[0.40,\,0.89]$ & 14.38 \\
gemini-r-er-1.5 (0) & $0.70 \;[0.40,\,0.89]$ & $0.60 \;[0.31,\,0.83]$ & 15.10 \\
gemini-2.5-flash & $0.70 \;[0.40,\,0.89]$ & $0.60 \;[0.31,\,0.83]$ & 15.45 \\
Keep-station & $0.70 \;[0.40,\,0.89]$ & $0.70 \;[0.40,\,0.89]$ & $\sim$0.00 \\
gpt-5-nano-medium & $0.70 \;[0.40,\,0.89]$ & $0.70 \;[0.40,\,0.89]$ & 12.18 \\
gpt-5-nano-minimal & $0.70 \;[0.40,\,0.89]$ & $0.70 \;[0.40,\,0.89]$ & 5.69 \\
gpt-5-low & $0.80 \;[0.49,\,0.94]$ & $0.70 \;[0.40,\,0.89]$ & 18.82 \\
gemini-2.5-pro & $0.80 \;[0.49,\,0.94]$ & $0.70 \;[0.40,\,0.89]$ & 51.78 \\
claude-sonnet-4 & $0.80 \;[0.49,\,0.94]$ & $0.70 \;[0.40,\,0.89]$ & 4.56 \\
gpt-4o & $0.90 \;[0.60,\,0.98]$ & $0.80 \;[0.49,\,0.94]$ & 5.86 \\
gpt-5-medium & $0.90 \;[0.60,\,0.98]$ & $0.90 \;[0.60,\,0.98]$ & 27.91 \\
gpt-5-minimal & $0.90 \;[0.60,\,0.98]$ & $0.70 \;[0.40,\,0.89]$ & 7.16 \\
\hline
\end{tabular}
\end{table}